\documentclass{article} 
\usepackage{iclr2025_conference,times}

\usepackage{amsmath,amsfonts,bm}

\def\eqref#1{equation~\ref{#1}}

\def\1{\bm{1}}

\DeclareMathAlphabet{\mathsfit}{\encodingdefault}{\sfdefault}{m}{sl}
\SetMathAlphabet{\mathsfit}{bold}{\encodingdefault}{\sfdefault}{bx}{n}

\usepackage{hyperref}
\usepackage{url}

\author{Fu-Chieh Chang \\
MediaTek Research, Taipei, Taiwan \\
Graduate Institute of Communication Engineering, National Taiwan University, Taipei, Taiwan \\
\texttt{d09942015@ntu.edu.tw}
\AND
You-Chen Lin  \\
Graduate Institute of Communication Engineering, National Taiwan University, Taipei, Taiwan \\
\texttt{r13921A30@ntu.edu.tw}
\AND
Pei-Yuan Wu  \\
Graduate Institute of Communication Engineering, National Taiwan University, Taipei, Taiwan \\
\texttt{peiyuanwu@ntu.edu.tw}
}

\usepackage{microtype}
\usepackage{graphicx}
\usepackage{subfigure}
\usepackage{booktabs} 
\usepackage{fmtcount}

\usepackage{hyperref}
\usepackage{subcaption}
\usepackage{amsmath}
\usepackage{graphicx}
\usepackage{amsfonts}
\usepackage{tikz-cd}
\usepackage{mathrsfs}
\usepackage{caption}
\usepackage{subcaption}
\usepackage{extpfeil}
\usepackage{algorithm}
\usepackage{algorithmic}
\usepackage{CJKutf8}
\usepackage{xcolor}
\usepackage{pgfplots}
\pgfplotsset{compat=1.17}

\definecolor{mycolor1}{rgb}{0.5, 0, 0}    
\definecolor{mycolor2}{rgb}{1, 0.5, 0} 
\definecolor{mycolor3}{rgb}{0.8, 0.8, 0}   
\definecolor{mycolor4}{rgb}{0, 0.4, 0.2}   
\definecolor{mycolor5}{rgb}{0, 0.8, 0.4}   
\definecolor{mycolor6}{rgb}{0, 0.8, 0.8}   

\definecolor{dc0}{rgb}{1, 0, 0}   
\definecolor{dc1}{rgb}{0.5, 0, 0}   
\definecolor{dc2}{rgb}{0.5, 0.5, 0}   
\definecolor{dc3}{rgb}{0, 1, 0}   
\definecolor{dc4}{rgb}{0, 0.5, 0}   
\definecolor{dc5}{rgb}{0, 0.5, 0.5}   
\definecolor{dc6}{rgb}{0, 0, 1}   
\definecolor{dc7}{rgb}{0, 0, 0.5}   
\definecolor{dc8}{rgb}{0.5, 0, 0.5}   
\definecolor{dc9}{rgb}{0.5, 0.5, 0.5}   
\definecolor{dc10}{rgb}{0.8, 0, 0}   
\definecolor{dc11}{rgb}{0.6, 0.2, 0}   
\definecolor{dc12}{rgb}{0.2, 0.6, 0}   
\definecolor{dc13}{rgb}{0, 0.8, 0}   
\definecolor{dc14}{rgb}{0, 0.6, 0.2}   
\definecolor{dc15}{rgb}{0, 0.2, 0.6}   
\definecolor{dc16}{rgb}{0, 0, 0.8}   
\definecolor{dc17}{rgb}{0.2, 0, 0.6}   
\definecolor{dc18}{rgb}{0.2, 0, 0.6}   
\definecolor{dc19}{rgb}{0.2, 0.2, 0.2}   

\definecolor{pc11}{rgb}{1, 0.7, 0.7}   
\definecolor{pc12}{rgb}{1, 0.4, 0.4}   
\definecolor{pc13}{rgb}{0.4, 0, 0}   
\definecolor{pc21}{rgb}{0.1, 1, 0.1}   
\definecolor{pc22}{rgb}{0, 0.8, 0}   
\definecolor{pc23}{rgb}{0, 0.2, 0}   
\definecolor{pc31}{rgb}{0.7, 0.7, 1}   
\definecolor{pc32}{rgb}{0.4, 0.4, 1}   
\definecolor{pc33}{rgb}{0, 0, 0.4}   

\usepackage{amsmath}
\usepackage{amssymb}
\usepackage{mathtools}
\usepackage{amsthm}

\usepackage[capitalize,noabbrev]{cleveref}

\usepackage{amsmath}
\usepackage{cases}
\usepackage{graphicx}
\usepackage{hyperref}

\usepackage{tikz-cd}
\tikzcdset{scale cd/.style={every label/.append style={scale=#1},
    cells={nodes={scale=#1}}}}
\usepackage{caption}
\usepackage{subcaption}
\usepackage{extpfeil}
\usepackage{algorithmic}
\usepackage{comment}
\usepackage{pgfplots}
\usepackage{listings}
\usepackage{natbib}

\newcommand{\zfill}[2]{%
  \numberstringnum{#1}%
  \ifnum\value{#1}<10 0\fi%
  \ifnum\value{#1}<100 0\fi%
  \ifnum\value{#1}<1000 0\fi%
  \the#1%
}

\usepackage{CJKutf8}
\title{Unraveling Arithmetic in Large Language Models: The Role of Algebraic Structures}

\usepackage{amsthm}

\theoremstyle{plain}
\newtheorem{theorem}{Theorem}[section]

\theoremstyle{definition}
\newtheorem{definition}[theorem]{Definition}

\theoremstyle{remark}
\newtheorem{remark}[theorem]{Remark}

\iclrfinalcopy 

\begin{document}

\maketitle

\begin{abstract}
Large language models (LLMs) have demonstrated remarkable mathematical capabilities, largely driven by chain-of-thought (CoT) prompting, which decomposes complex reasoning into step-by-step solutions. This approach has enabled significant advancements, as evidenced by performance on benchmarks like GSM8K and MATH. However, the mechanisms underlying LLMs' ability to perform arithmetic in a single step of CoT remain poorly understood. Existing studies debate whether LLMs encode numerical values or rely on symbolic reasoning, while others explore attention and multi-layered processing in arithmetic tasks. In this work, we propose that LLMs learn arithmetic by capturing algebraic structures, such as commutativity and identity properties. Since these structures are observable through input-output relationships, they can generalize to unseen data. We empirically demonstrate that LLMs can learn algebraic structures using a custom dataset of arithmetic problems, as well as providing theoretical evidence showing that, under specific configurations of weights and biases, the transformer-based LLMs can generate embeddings that remain invariant to both permutations of input tokens and the presence of identity elements. Our findings indicate that leveraging algebraic structures can enhance the LLMs' arithmetic capabilities, offering insights into improving their arithmetic performance.  
\end{abstract}

\section{Introduction}
\label{sec:problem_settings}

With the advancement of large-language models (LLMs), their mathematical capabilities have become a crucial factor in their success. This progress is largely attributed to chain-of-thought (CoT) prompting~\cite{wei2022chain}, which enables LLMs to move beyond pattern matching and tackle complex reasoning problems through step-by-step guidance. The mathematical prowess of LLMs is demonstrated by several commercial models~\cite{chatgpt2024,claude3} that have achieved notable success on benchmarks such as GSM8K~\cite{cobbe2021training} and MATH~\cite{hendrycks2021measuring}.

Despite their achievements in solving arithmetic problems, the underlying mechanisms through which LLMs learn arithmetic operations from training data remain unclear. Although CoT explains how breaking down tasks into smaller steps facilitates mathematical problem solving, how LLMs process arithmetic tokens and execute arithmetic operations within individual steps of CoT are not yet well understood. Some studies~\cite{Fangwei240103735,levy2024language} suggest that LLMs encode numerical values, while others~\cite{deng2024language} propose that LLMs learn symbolic relationships between numbers rather than directly encoding their values. Furthermore, studies such as \cite{gorceix2024learningmathematicalruleslarge} suggest that LLMs can learn mathematical rules. However, no consensus has been reached on how or why LLMs are capable of arithmetic reasoning.
A detailed literature review is shown in Sec.\ref{sec:related_works}.

In this work, we provide a novel point of view on how LLMs acquire arithmetic abilities, suggesting that such skills stem from the learning of algebraic structures such as commutativity and identity. Since LLMs only observe input and output tokens rather than explicit numerical values, direct token-to-number mapping is challenging. Instead, LLMs infer algebraic structures by examining the relationship between inputs and outputs. Previous work~\cite{karjol2023neural,karjol2024unified,yang2023latent} has demonstrated that machine learning methods can learn symmetric structures from training data; however, these studies have not yet connected such findings to arithmetic learning in LLMs. In this work, we present both empirical and theoretical support showing that LLMs can learn algebraic structures from training data, generalizing these structures to unseen inputs. Our main contributions are as follows.
\begin{itemize}
    \item \textbf{Empirical Evidence:} By constructing a dataset of arithmetic problems and splitting it into training and testing sets, we demonstrate that LLMs can learn and generalize the \emph{commutativity} and \emph{identity} properties to unseen inputs.
    \item \textbf{Theoretical Construction:} We provide a constructive proof illustrating how transformer-based models, with specific weight and bias configurations, can preserve hidden-state invariance under token permutations and the insertion of identity elements.
\end{itemize}
Overall, these findings suggest that LLMs can internalize algebraic structures, providing a foundation for designing strategies to further enhance their arithmetic capabilities.

\section{Methodology}
\subsection{Problem Settings}
We demonstrate that LLMs can learn algebraic structures from training samples and generalize to previously unseen instances. 
To keep our focus clear, we concentrate on arithmetic problems represented by numeric and operator symbols, rather than natural language or real-world contexts. 
Moreover, our study is set within a finite Abelian group (see Sec.\ref{label:abelian_group}). 
A well-known example of a finite Abelian group is $\mathbb{Z}_n$, the set of integers modulo $n$ under addition modulo $n$. 
For example, in $\mathbb{Z}_5$, the elements are $\{0,1,2,3,4\}$, and the group operation is defined as $a+b \bmod 5$. The identity element of this group is 0 , and every element $a \in \mathbb{Z}_n$ has an inverse $b \in \mathbb{Z}_n$ such that $a+b \equiv 0 \bmod 5$. 
In this work, we analyze the group operation properties of \emph{commutativity} and \emph{identity} in $\mathbb{Z}_n$.

\begin{figure*}[hbt!]
    \begin{center}
    \resizebox{.98\textwidth}{!}{
    \begin{tabular}{|l |l |l |l|}
\hline
\multicolumn{2}{|c|}{\scriptsize{Training, for Operator ``$+$"}} & \multicolumn{2}{|c|}{\scriptsize{Testing, for Operator ``$+$"}} \\ 
\hline   \multicolumn{1}{|c|}{\scriptsize{Commutativity}} &   \multicolumn{1}{|c|}{\scriptsize{Identity}} &   \multicolumn{1}{|c|}{\scriptsize{Commutativity}} & \multicolumn{1}{|c|}{\scriptsize{Identity}} \\ 
\hline
\tiny{\color{red}{$z_{3}+z_{4}+z_{5}+z_{5}+z_{5}+z_{6}=z_{0}$}} 
    &
    \tiny{\color{blue}{$z_{0}+z_{4}+z_{3}+z_{5}+z_{3}+z_{1}=z_{2}$}}
&
\tiny{\color{orange}{$z_{2}+z_{3}+z_{3}+z_{5}+z_{5}+z_{6}=z_{3}$}}
&
\tiny{\color{cyan}{$z_{0}+z_{5}+z_{2}+z_{5}+z_{2}+z_{3}=z_{3}$}}
\\   
\tiny{\color{red}{$z_{4}+z_{5}+z_{3}+z_{5}+z_{5}+z_{6}=z_{0}$}} 
&
    \tiny{\color{blue}{$z_{4}+z_{0}+z_{3}+z_{5}+z_{3}+z_{1}=z_{2}$}}
&
\tiny{\color{orange}{$z_{3}+z_{3}+z_{5}+z_{5}+z_{6}+z_{2}=z_{3}$}}
&
\tiny{\color{cyan}{$z_{5}+z_{0}+z_{2}+z_{5}+z_{2}+z_{3}=z_{3}$}}
\\   
\tiny{\color{red}{$z_{3}+z_{5}+z_{5}+z_{6}+z_{4}+z_{5}=z_{0}$}} 
&
    \tiny{\color{blue}{$z_{4}+z_{3}+z_{0}+z_{5}+z_{3}+z_{1}=z_{2}$}}
&
\tiny{\color{orange}{$z_{5}+z_{6}+z_{5}+z_{3}+z_{2}+z_{3}=z_{3}$}}
&
\tiny{\color{cyan}{$z_{5}+z_{2}+z_{0}+z_{5}+z_{2}+z_{3}=z_{3}$}}
\\   
\tiny{\color{red}{$z_{4}+z_{5}+z_{5}+z_{6}+z_{3}+z_{5}=z_{0}$}} 
&
    \tiny{\color{blue}{$z_{4}+z_{3}+z_{5}+z_{0}+z_{3}+z_{1}=z_{2}$}}
&
\tiny{\color{orange}{$z_{3}+z_{2}+z_{5}+z_{3}+z_{6}+z_{5}=z_{3}$}}
& 
\tiny{\color{cyan}{$z_{5}+z_{2}+z_{5}+z_{0}+z_{2}+z_{3}=z_{3}$}}
\\   
\tiny{\color{red}{$z_{6}+z_{5}+z_{5}+z_{3}+z_{5}+z_{4}=z_{0}$}} 
& 
\tiny{\color{blue}{$z_{4}+z_{3}+z_{5}+z_{3}+z_{0}+z_{1}=z_{2}$}}
&
\tiny{\color{orange}{$z_{5}+z_{3}+z_{5}+z_{3}+z_{2}+z_{6}=z_{3}$}}
&
\tiny{\color{cyan}{$z_{5}+z_{2}+z_{5}+z_{2}+z_{0}+z_{3}=z_{3}$}}
\\   
\tiny{\color{red}{$z_{3}+z_{4}+z_{5}+z_{5}+z_{6}+z_{5}=z_{0}$}} 
&
\tiny{\color{blue}{$z_{4}+z_{3}+z_{5}+z_{3}+z_{1}+z_{0}=z_{2}$}}

&
\tiny{\color{orange}{$z_{6}+z_{3}+z_{5}+z_{3}+z_{2}+z_{5}=z_{3}$}}
& 
\tiny{\color{cyan}{$z_{5}+z_{2}+z_{5}+z_{2}+z_{3}+z_{0}=z_{3}$}}
\\   
\tiny{\color{red}{$z_{3}+z_{6}+z_{4}+z_{5}+z_{5}+z_{5}=z_{0}$}} 
&
\tiny{\color{blue}{$z_{4}+z_{3}+z_{5}+z_{3}+z_{1}=z_{2}$}}
&
& 
\\   
\tiny{\color{orange}{$z_{6}+z_{3}+z_{5}+z_{3}+z_{5}+z_{2}=z_{3}$}}
&
\tiny{\color{cyan}{$z_{5}+z_{2}+z_{5}+z_{2}+z_{3}=z_{3}$}}
&
& 
\\
\hline
\hline
\multicolumn{2}{|c|}{\scriptsize{Training, for Operator $\oplus$}} & \multicolumn{2}{|c|}{\scriptsize{Testing, for Operator $\oplus$}} \\ 
\hline   \multicolumn{1}{|c|}{\scriptsize{Commutativity}} &   \multicolumn{1}{|c|}{\scriptsize{Identity}} &   \multicolumn{1}{|c|}{\scriptsize{Commutativity}} & \multicolumn{1}{|c|}{\scriptsize{Identity}} \\ 
\hline
\tiny{\color{red}{$z_{3}\oplus z_{4}\oplus z_{5}\oplus z_{5}\oplus z_{5}\oplus z_{6}=r_{2}$}} 
    &
    \tiny{\color{blue}{$z_{0}\oplus z_{4}\oplus z_{3}\oplus z_{5}\oplus z_{3}\oplus z_{1}=r_{1}$}}
&
\tiny{\color{orange}{$z_{2}\oplus z_{3}\oplus z_{3}\oplus z_{5}\oplus z_{5}\oplus z_{6}=r_{4}$}}
&
\tiny{\color{cyan}{$z_{0}\oplus z_{5}\oplus z_{2}\oplus z_{5}\oplus z_{2}\oplus z_{3}=r_{5}$}}
\\   
\tiny{\color{red}{$z_{4}\oplus z_{5}\oplus z_{3}\oplus z_{5}\oplus z_{5}\oplus z_{6}=r_{2}$}} 
&
    \tiny{\color{blue}{$z_{4}\oplus z_{0}\oplus z_{3}\oplus z_{5}\oplus z_{3}\oplus z_{1}=r_{1}$}}
&
\tiny{\color{orange}{$z_{3}\oplus z_{3}\oplus z_{5}\oplus z_{5}\oplus z_{6}\oplus z_{2}=r_{4}$}}
&
\tiny{\color{cyan}{$z_{5}\oplus z_{0}\oplus z_{2}\oplus z_{5}\oplus z_{2}\oplus z_{3}=r_{5}$}}
\\   
\tiny{\color{red}{$z_{3}\oplus z_{5}\oplus z_{5}\oplus z_{6}\oplus z_{4}\oplus z_{5}=r_{2}$}} 
&
    \tiny{\color{blue}{$z_{4}\oplus z_{3}\oplus z_{0}\oplus z_{5}\oplus z_{3}\oplus z_{1}=r_{1}$}}
&
\tiny{\color{orange}{$z_{5}\oplus z_{6}\oplus z_{5}\oplus z_{3}\oplus z_{2}\oplus z_{3}=r_{4}$}}
&
\tiny{\color{cyan}{$z_{5}\oplus z_{2}\oplus z_{0}\oplus z_{5}\oplus z_{2}\oplus z_{3}=r_{5}$}}
\\   
\tiny{\color{red}{$z_{4}\oplus z_{5}\oplus z_{5}\oplus z_{6}\oplus z_{3}\oplus z_{5}=r_{2}$}} 
&
    \tiny{\color{blue}{$z_{4}\oplus z_{3}\oplus z_{5}\oplus z_{0}\oplus z_{3}\oplus z_{1}=r_{1}$}}
&
\tiny{\color{orange}{$z_{3}\oplus z_{2}\oplus z_{5}\oplus z_{3}\oplus z_{6}\oplus z_{5}=r_{4}$}}
& 
\tiny{\color{cyan}{$z_{5}\oplus z_{2}\oplus z_{5}\oplus z_{0}\oplus z_{2}\oplus z_{3}=r_{5}$}}
\\   
\tiny{\color{red}{$z_{6}\oplus z_{5}\oplus z_{5}\oplus z_{3}\oplus z_{5}\oplus z_{4}=r_{2}$}} 
& 
\tiny{\color{blue}{$z_{4}\oplus z_{3}\oplus z_{5}\oplus z_{3}\oplus z_{0}\oplus z_{1}=r_{1}$}}
&
\tiny{\color{orange}{$z_{5}\oplus z_{3}\oplus z_{5}\oplus z_{3}\oplus z_{2}\oplus z_{6}=r_{4}$}}
&
\tiny{\color{cyan}{$z_{5}\oplus z_{2}\oplus z_{5}\oplus z_{2}\oplus z_{0}\oplus z_{3}=r_{5}$}}
\\   
\tiny{\color{red}{$z_{3}\oplus z_{4}\oplus z_{5}\oplus z_{5}\oplus z_{6}\oplus z_{5}=r_{2}$}} 
&
\tiny{\color{blue}{$z_{4}\oplus z_{3}\oplus z_{5}\oplus z_{3}\oplus z_{1}\oplus z_{0}=r_{1}$}}

&
\tiny{\color{orange}{$z_{6}\oplus z_{3}\oplus z_{5}\oplus z_{3}\oplus z_{2}\oplus z_{5}=r_{4}$}}
& 
\tiny{\color{cyan}{$z_{5}\oplus z_{2}\oplus z_{5}\oplus z_{2}\oplus z_{3}\oplus z_{0}=r_{5}$}}
\\   
\tiny{\color{red}{$z_{3}\oplus z_{6}\oplus z_{4}\oplus z_{5}\oplus z_{5}\oplus z_{5}=r_{2}$}} 
&
\tiny{\color{blue}{$z_{4}\oplus z_{3}\oplus z_{5}\oplus z_{3}\oplus z_{1}=r_{1}$}}
&
& 
\\   
\tiny{\color{orange}{$z_{6}\oplus z_{3}\oplus z_{5}\oplus z_{3}\oplus z_{5}\oplus z_{2}=r_{4}$}}
&
\tiny{\color{cyan}{$z_{5}\oplus z_{2}\oplus z_{5}\oplus z_{2}\oplus z_{3}=r_{5}$}}
&
& 
\\
\hline
\hline
\multicolumn{2}{|c|}{\scriptsize{Training, for Operator $\ominus$}} & \multicolumn{2}{|c|}{\scriptsize{Testing, for Operator $\ominus$}} \\ 
\hline  
\tiny{\color{red}{$z_{3}\ominus z_{4}\ominus z_{5}\ominus z_{5}\ominus z_{5}\ominus z_{6}=2$}} 
    &
    \tiny{\color{blue}{$z_{0}\ominus z_{4}\ominus z_{3}\ominus z_{5}\ominus z_{3}\ominus z_{1}=3$}}
&
\tiny{\color{orange}{$z_{2}\ominus z_{3}\ominus z_{3}\ominus z_{5}\ominus z_{5}\ominus z_{6}=2$}}
&
\tiny{\color{cyan}{$z_{0}\ominus z_{5}\ominus z_{2}\ominus z_{5}\ominus z_{2}\ominus z_{3}=2$}}
\\   
\tiny{\color{red}{$z_{4}\ominus z_{5}\ominus z_{3}\ominus z_{5}\ominus z_{5}\ominus z_{6}=2$}} 
&
    \tiny{\color{blue}{$z_{4}\ominus z_{0}\ominus z_{3}\ominus z_{5}\ominus z_{3}\ominus z_{1}=3$}}
&
\tiny{\color{orange}{$z_{3}\ominus z_{3}\ominus z_{5}\ominus z_{5}\ominus z_{6}\ominus z_{2}=3$}}
&
\tiny{\color{cyan}{$z_{5}\ominus z_{0}\ominus z_{2}\ominus z_{5}\ominus z_{2}\ominus z_{3}=2$}}
\\   
\tiny{\color{red}{$z_{3}\ominus z_{5}\ominus z_{5}\ominus z_{6}\ominus z_{4}\ominus z_{5}=2$}} 
&
    \tiny{\color{blue}{$z_{4}\ominus z_{3}\ominus z_{0}\ominus z_{5}\ominus z_{3}\ominus z_{1}=4$}}

&
\tiny{\color{orange}{$z_{5}\ominus z_{6}\ominus z_{5}\ominus z_{3}\ominus z_{2}\ominus z_{3}=3$}}
&
\tiny{\color{cyan}{$z_{5}\ominus z_{2}\ominus z_{0}\ominus z_{5}\ominus z_{2}\ominus z_{3}=3$}}
\\   
\tiny{\color{red}{$z_{4}\ominus z_{5}\ominus z_{5}\ominus z_{6}\ominus z_{3}\ominus z_{5}=2$}} 
&
    \tiny{\color{blue}{$z_{4}\ominus z_{3}\ominus z_{5}\ominus z_{0}\ominus z_{3}\ominus z_{1}=3$}}
&
\tiny{\color{orange}{$z_{3}\ominus z_{2}\ominus z_{5}\ominus z_{3}\ominus z_{6}\ominus z_{5}=3$}}
& 
\tiny{\color{cyan}{$z_{5}\ominus z_{2}\ominus z_{5}\ominus z_{0}\ominus z_{2}\ominus z_{3}=2$}}
\\   
\tiny{\color{red}{$z_{6}\ominus z_{5}\ominus z_{5}\ominus z_{3}\ominus z_{5}\ominus z_{4}=4$}} 
& 
    \tiny{\color{blue}{$z_{4}\ominus z_{3}\ominus z_{5}\ominus z_{3}\ominus z_{0}\ominus z_{1}=3$}}
&
\tiny{\color{orange}{$z_{5}\ominus z_{3}\ominus z_{5}\ominus z_{3}\ominus z_{2}\ominus z_{6}=3$}}
&
\tiny{\color{cyan}{$z_{5}\ominus z_{2}\ominus z_{5}\ominus z_{2}\ominus z_{0}\ominus z_{3}=3$}}
\\   
\tiny{\color{red}{$z_{3}\ominus z_{4}\ominus z_{5}\ominus z_{5}\ominus z_{6}\ominus z_{5}=2$}} 
&
    \tiny{\color{blue}{$z_{4}\ominus z_{3}\ominus z_{5}\ominus z_{3}\ominus z_{1}\ominus z_{0}=4$}}

&
\tiny{\color{orange}{$z_{6}\ominus z_{3}\ominus z_{5}\ominus z_{3}\ominus z_{2}\ominus z_{5}=3$}}
& 
\tiny{\color{cyan}{$z_{5}\ominus z_{2}\ominus z_{5}\ominus z_{2}\ominus z_{3}\ominus z_{0}=3$}}
\\   
\tiny{\color{red}{$z_{3}\ominus z_{6}\ominus z_{4}\ominus z_{5}\ominus z_{5}\ominus z_{5}=3$}} 
&
    \tiny{\color{blue}{$z_{4}\ominus z_{3}\ominus z_{5}\ominus z_{3}\ominus z_{1}=3$}}
&
& 
\\   
\tiny{\color{orange}{$z_{6}\ominus z_{3}\ominus z_{5}\ominus z_{3}\ominus z_{5}\ominus z_{2}=3$}}
&
\tiny{\color{cyan}{$z_{5}\ominus z_{2}\ominus z_{5}\ominus z_{2}\ominus z_{3}=2$}}
&
& 
\\
\hline
\end{tabular}

    }
    \end{center}
    \caption{Illustration of dataset for operator ``$+$", $\oplus$ and $\ominus$. Notice that the same set of tokens is maintained across all operators to ensure that certain token combinations appear exclusively either in the training set or the testing set, as required.}
    \label{fig:dataset}
\end{figure*}
\subsection{Dataset for Commutativity and Identity}\label{sec:dataset}
In this work, we show that LLMs can acquire the concepts of commutativity and identity purely from the dataset we provide, rather than relying on any preexisting, pre-trained knowledge. To validate this, we construct a dataset of addition problems in $\mathbb{Z}_n$. Each element in $\mathbb{Z}_n$ is denoted as $z_i$, where $z_0 = 0$, $z_1 = 1$, $\ldots$, $z_{n-1} = n-1$. The dataset consists  of addition problems with $M$ terms, formally expressed as:
\begin{equation}\label{eq:addition_problem}
z_{i_1} + z_{i_2} + \cdots + z_{i_M}=z_{\bigl(i_1 + i_2 + \cdots + i_M\bigr) \bmod n},
\end{equation}
where $0 \leq i_1, i_2, \ldots, i_M < n$. Here, $\bmod$ denotes the modulo operator. For each problem, ``$z_{i_1}+z_{i_2}+\cdots+z_{i_M} =$ '' serves as input tokens, while the label is ``$z_{\bigl(i_1 + i_2 + \cdots + i_M\bigr) \bmod n}$''. Thus, the LLM must predict the correct element ``$z_{\bigl(i_1 + i_2 + \cdots + i_M\bigr) \bmod n}$'' given the inputs. In addition, we focus on the scenario that  the model must directly predict the label from the input without performing any intermediate CoT reasoning.  Fig.~\ref{fig:dataset} and Appendix~\ref{sec:example_dataset} provide examples of both the training and testing sets for $\mathbb{Z}_7$. In the subsequent sections, we explain how the datasets are constructed so as to test whether LLMs genuinely learn and generalize the underlying algebraic properties of commutativity and identity.

\paragraph{Commutativity:}  
To demonstrate that LLMs can learn the commutative property, we begin by selecting a sequence \(\bigl(z_{i_1}, z_{i_2}, \dots, z_{i_M}\bigr)\) from \(\mathbb{Z}_n\), where \(z_{i_1} \le z_{i_2} \le \dots \le z_{i_M}\) and \(z_{i_1} > 0\), and generate sequences in the form of Eq.~\eqref{eq:addition_problem}.
Next, we sample several permutations of \(\bigl(z_{i_1}, z_{i_2}, \dots, z_{i_M}\bigr)\), denoting each permutation as \(\bigl(z_{j_1}, z_{j_2}, \dots, z_{j_M}\bigr)\), and generate sequences as in Eq.~\eqref{eq:addition_problem} accordingly. 
For any given sequence \(\bigl(z_{i_1}, z_{i_2}, \dots, z_{i_M}\bigr)\), to allow the model to recognize the element \(z_{\bigl(i_1 + i_2 + \cdots + i_M\bigr)\bmod n}\) in the first place, at least one permutation of each sequence in testsing set appears in the training set.
However, if more than one permutations are included in the training set, all other permutations of this same sequence are excluded from the training set and placed in the testing set. 
This ensures that the model must generalize the commutativity property, i.e. it must infer correct outputs for unseen permutations in the testing set, based on the only one permutation learned during training. 
As shown in the upper row of Fig.~\ref{fig:dataset}, every permutation of \((z_3, z_4, z_5, z_5, z_5, z_6)\) is included in the training set (highlighted in red). Meanwhile, the testing set contains a different permutation sequences of Eq.~\eqref{eq:addition_problem}, \((z_2, z_3, z_3, z_5, z_5, z_6)\), highlighted in orange. However, one permutation of that sequence—\((z_6, z_3, z_5, z_3, z_5, z_2)\)—does appear in the training set, ensuring that the model is exposed to at least one variant of the same element sum.

\paragraph{Identity:}
In \(\mathbb{Z}_n\), the identity element for addition is \(z_0\). To verify whether LLMs learn this identity property, we first pick \(M-1\) variables, \(\bigl(z_{i_1}, \dots, z_{i_{M-1}}\bigr)\) from $\mathbb{Z}_n$, and insert \(z_0\) among them in all possible positions. Substituting each arrangement into Eq.~\eqref{eq:addition_problem} yields \(M\) distinct equations:
\[
\begin{aligned}
    z_0 + z_{i_1} + z_{i_2} + \cdots + z_{i_{M-1}} &= z_{\bigl(i_1 + i_2 + \cdots + i_{M-1}\bigr) \bmod n}, \\
    z_{i_1} + z_0 + z_{i_2} + \cdots + z_{i_{M-1}} &= z_{\bigl(i_1 + i_2 + \cdots + i_{M-1}\bigr) \bmod n}, \\
    &\;\;\vdots \\
    z_{i_1} + z_{i_2} + \cdots + z_{i_{M-1}} + z_0 &= z_{\bigl(i_1 + i_2 + \cdots + i_{M-1}\bigr) \bmod n}.
\end{aligned}
\]
To ensure that the model does not merely exploit permutation invariance, we assign all possible insertions of \(z_0\) in the sequence \(\bigl(z_{i_1}, \dots, z_{i_{M-1}}\bigr)\) exclusively to either the training set or the testing set. Moreover, because the value of \(z_{\bigl(i_1 + i_2 + \cdots + i_{M-1}\bigr)\bmod n}\) should be established using the equation without the identity element, the following \emph{base equation} without \(z_0\) must be included in the training set:
\[
z_{i_1} + z_{i_2} + \cdots + z_{i_{M-1}}
= z_{\bigl(i_1 + i_2 + \cdots + i_{M-1}\bigr)\bmod n}.
\]
In the first row of Fig.~\ref{fig:dataset}, we illustrate how identity elements appear in both the training and testing sets. The training set includes the \emph{base equation}, \(z_4 + z_3 + z_5 + z_3 + z_1 = z_2\), along with variants where \(z_0\) is inserted into every possible position (highlighted in blue). In contrast, the equation \(z_0 + z_5 + z_2 + z_5 + z_2 + z_3 = z_3\) and its variants with \(z_0\) inserted in different positions are placed in the testing set (highlighted in cyan). However, the base equation \(z_5 + z_2 + z_5 + z_2 + z_3 = z_3\) appears in the training set, ensuring that the model can infer \(z_0 + z_5 + z_2 + z_5 + z_2 + z_3 = z_3\) when recognizing \(z_0\) as the identity element.
\subsection{Dataset to Exclude Numerical Calculation}\label{sec:additional_dataset_no_numerical}
Training on the dataset introduced in Sec.~\ref{sec:dataset} leaves the possibility that LLMs might exploit numerical relationships inherent in the indices. Specifically, an LLM could potentially identify the input token indices and rely on computing the modulo sum of these indices to derive results, bypassing the application of commutativity and identity principles. 
For example, in Fig.~\ref{fig:dataset}, the dataset for commutativity contains
$ z_{6}+z_{3}+z_{5}+z_{3}+z_{5}+z_{2}=z_{3}$ in training set and $z_{2}+z_{3}+z_{3}+z_{5}+z_{5}+z_{6}=z_{3}$ in testing set.
When answering the test equation, we anticipate that LLMs are able to leverage commutativity to infer \(z_{3}\) from the training equation. Nonetheless, we cannot entirely rule out the possibility that an LLM could deduce the numerical value of the index \((2,3,3,5,5,6)\) and use the sum \((2+3+3+5+5+6)\bmod 7 = 3\). To prevent this, we introduce a new operator \(\oplus\) which takes the same inputs as “\(+\)” but produces outputs unrelated to the inputs in a numerical sense.
Specifically, given any sequence \(\bigl(z_{i_1}, z_{i_2}, \dots, z_{i_M}\bigr)\) in \(\mathbb{Z}_n\) where \(z_{i_1} \le z_{i_2} \le \dots \le z_{i_M}\) and \(z_{i_1} > 0\), the result of applying \(\oplus\) to this sequence (or any permutation of it) is a randomly selected element \(r_i\) from the set \(\{r_0, r_1, \dots, r_{n-1}\}\), whose elements lie outside \(\mathbb{Z}_n\).
Hence, $\oplus$ is invariant under input permutations. Namely,\begin{align*}
z_{i_1} \oplus z_{i_2} \oplus \dots \oplus z_{i_M} = r_i, \quad
z_{i_2} \oplus z_{i_1} \oplus \dots \oplus z_{i_M} = r_i, \quad
\cdots, \quad
z_{i_M} \oplus z_{i_{M-1}} \oplus \dots \oplus z_{i_1} = r_i.
\end{align*}
Besides, \(\oplus\) is invariant under insertion of the identity element $z_0$ at any position such that
$$ 
\begin{aligned}
z_{0} \oplus z_{i_1}  \oplus  z_{i_2} \oplus \dots \oplus z_{i_M} = r_i,  \quad
\cdots, \quad
z_{i_1} \oplus  z_{i_2} \oplus \dots\oplus z_{i_M}  \oplus z_{0}  = r_i.
\end{aligned}
$$
Hence, for LLMs to perform well on this dataset, they cannot rely on the numerical relationship of the index; rather, they must learn the commutative property and identity property of \(\oplus\).
We include a corresponding dataset involving \(\oplus\) in both training and testing. As shown in the second row of Fig.~\ref{fig:dataset}, we replace “\(+\)” with \(\oplus\) and adjust the resulting values accordingly. 
By comparing how ``\(+\)'' and \(\oplus\) perform on commutativity and identity tasks under the same training and test inputs in \(\mathbb{Z}_n\), we can discern whether the performance of ``\(+\)'' benefits from numerical computation or purely from  algebraic structures.

\subsection{Dataset to Avoid Trivial Solutions}\label{sec:additional_dataset}
In this work, our goal is for LLMs to learn the commutative and identity properties specifically for the ``\(+\)" and $\oplus$ operator. If the model were trained only on the dataset described in the previous sections, it could adopt a trivial solution in which commutativity and identity trivially apply to all tokens $z_i$ regardless of the existence of the operator ``$+$" or $\oplus$. For instance, by setting all position embeddings to zero (see Sec.~\ref{sec:llm_explanation}). To prevent this, we include an additional dataset that features operators that lack commutativity and identity properties. Specifically, we introduce three new operators, \(\ominus\), \(\triangleleft\), and \(\triangleright\). Details of these operators are provided as follows.

\paragraph{Operator  $\ominus$ : Counts of Encountering $z_0$ along the Cyclic Group $Z_{n}$.}
It is straightforward to show that \(\mathbb{Z}_n\) is a cyclic group, as illustrated below:
\[
\begin{tikzcd}
	{z_5} & {z_4} & {z_3} & {z_2} \\
	\cdots & {z_{n-1}} & {z_0} & {z_1}
	\arrow[dashed, from=1-1, to=2-1]
	\arrow[from=1-2, to=1-1]
	\arrow[from=1-3, to=1-2]
	\arrow[from=1-4, to=1-3]
	\arrow[dashed, from=2-1, to=2-2]
	\arrow[from=2-2, to=2-3]
	\arrow[from=2-3, to=2-4]
	\arrow[from=2-4, to=1-4]
\end{tikzcd}
\]
We define \(\ominus : \mathbb{Z}_n \times \mathbb{Z}_n \to \mathbb{N}\) as an operator that counts the number of times \(z_0\) is encountered. Concretely, for any \(z_i, z_j \in \mathbb{Z}_n\), \(z_i \ominus z_j\) equals the number of occurrences of \(z_0\) when traveling from \(z_i\) to \(z_j\) around the cyclic group.
\begin{align*}
z_i\to z_{(i+1)\bmod n} &\to z_{(i+2)\bmod n} \to \dots\to z_{(j-1)\bmod n}\to z_j.
\end{align*}
 For example, in \(\mathbb{Z}_5\), \(z_3 \ominus z_1 = 1\) because traveling from \(z_3 \to z_4 \to z_0 \to z_1\) encounters \(z_0\) once. Similarly, \(z_2 \ominus z_5 = 0\) because \(z_2 \to z_3 \to z_4 \to z_5\) does not pass through \(z_0\). We set \(z_i \ominus z_i = 1\) because one must traverse all elements of \(\mathbb{Z}_n\) to return to the same element. We also define \(z_i \ominus z_j \ominus z_k\) as the total number of \(z_0\) encounters when traveling from \(z_i\) to \(z_j\), then continuing to \(z_k\), such that
$$
z_i\ominus z_j \ominus z_k = (z_i\ominus z_j) + (z_j \ominus z_k).
$$
For instance, in \(\mathbb{Z}_5\), \(z_4 \ominus z_2 \ominus z_1 = 2\).
Besides, it is straightforward to verify that \(\ominus\) does not satisfy commutativity, nor is \(z_0\) an identity element, 
because
$z_i \ominus z_j\neq z_j \ominus z_i
 ~(\text{for all } i \neq j \text{ with } i,j>0)$, and
$z_i \ominus z_j\neq z_i \ominus z_0 \ominus z_j ~(\text{for all } i < j \text{ with } i,j>0)$.
We add a dataset involving the \(\ominus\) operator to the training set and testing set. As shown in Fig.~\ref{fig:dataset}, we replace ``\(+\)'' with \(\ominus\) and update the resulting values accordingly. Note that the output of \(\ominus\) is a natural number in \(\mathbb{N}\) rather than an element of \(\mathbb{Z}_n\).
\paragraph{Operators $\triangleleft$ and $\triangleright$ : Left-Hand Side and Right-Hand Side Elements.} 
Alongside the \(\ominus\) operator, we introduce two additional operators, \(\triangleleft\) and \(\triangleright\), which also lack commutativity and  identity elements. These operators simply return the left-hand-side or right-hand-side argument, respectively:
\[
z_i \triangleright z_j=z_j
\quad\text{and}\quad
z_i \triangleleft z_j=z_i.
\]
It is straightforward to see that these operators do not satisfy commutativity and that \(z_0\) is not an identity element for either. However, they satisfy  associative, so expressions such as \(z_i \triangleright z_j \triangleright z_k\) produce a unique result.
We add a dataset that includes these operators in both the training and testing sets. 
An example of the whole dataset, which includes all operators ``$+$", $\oplus$, $\ominus$, $\triangleleft$ and $\triangleright$ is shown in Sec.~\ref{sec:example_dataset}. 
In the following section, we discuss how LLMs can learn the underlying algebraic structures from these datasets.
\begin{figure*}[hbt!]
    \begin{center}
    \resizebox{\textwidth}{!}{
    \begin{tikzcd}[column sep=tiny,row sep=tiny]
	{z_{i_1}} & {+} & \cdots & {+} & {z_{i_M}} & {=} \\
	{e_{i_1,1}} & {e_{+,2}} && {e_{+,2M-2}} & {e_{i_M,2M-1}} & {e_{=,2M}} \\
	\\
	{s^{(\ell-1)}_{1}} & {s^{(\ell-1)}_{2}} && {s^{(\ell-1)}_{2M-2}} & {s^{(\ell-1)}_{2M-1}} & {s^{(\ell-1)}_{2M}} \\
	\\
	\\
	\\
	{\quad\quad q_1,k_1,v_1\quad\quad} & {\quad\quad q_2,k_2,v_2\quad\quad} && {q_{2M-2},k_{2M-2},v_{2M-2}} & {q_{2M-1},k_{2M-1},v_{2M-1}} & {q_{2M},k_{2M},v_{2M}} \\
	&&&&& {s^{(\ell)}_{2M}=\sum_{i=1}^{2M}\sigma(\frac{q_{2M}^{\top}k_i}{\sum_{j=1}^{2M}q_{2M}^{\top}k_j})v_i} \\
	&&&&& {z_{\left(i_1+i_2+\cdots+i_M\right)\bmod n}}
	\arrow[from=1-1, to=2-1]
	\arrow[from=1-2, to=2-2]
	\arrow[from=1-4, to=2-4]
	\arrow[from=1-5, to=2-5]
	\arrow[from=1-6, to=2-6]
	\arrow[dashed, from=2-1, to=4-1]
	\arrow[dashed, from=2-2, to=4-2]
	\arrow[dashed, from=2-4, to=4-4]
	\arrow[dashed, from=2-5, to=4-5]
	\arrow[dashed, from=2-6, to=4-6]
	\arrow["{{{W_q,b_q,W_k,b_k,W_v,b_v}}}"{description}, from=4-1, to=8-1]
	\arrow["{{{W_q,b_q,W_k,b_k,W_v,b_v}}}"{description}, from=4-2, to=8-2]
	\arrow["{{{W_q,b_q,W_k,b_k,W_v,b_v}}}"{description}, from=4-4, to=8-4]
	\arrow["{{{W_q,b_q,W_k,b_k,W_v,b_v}}}"{description}, from=4-5, to=8-5]
	\arrow["{{{W_q,b_q,W_k,b_k,W_v,b_v}}}"{description}, from=4-6, to=8-6]
	\arrow[bend right=15, from=8-1, to=9-6]
	\arrow[bend right=15, from=8-2, to=9-6]
	\arrow[bend right=15, from=8-4, to=9-6]
	\arrow[bend right=15, from=8-5, to=9-6]
	\arrow[from=8-6, to=9-6]
	\arrow[dashed, from=9-6, to=10-6]
\end{tikzcd}
    }
    \end{center}
    \caption{Illustration of the symbols defined for the hidden states of tokens and the variables for the attnetion layers}
    \label{fig:transformer_illustration}
\end{figure*}

\subsection{How Do LLMs Learn Algebraic Structure?}\label{sec:llm_explanation}
In this section, we provide theoretical evidence that LLMs composed of attention layers can be constructed to compute addition under commutativity and identity. We present an example using \(\mathbb{Z}_n\) with \(M\) input elements as an illustrative example (see Fig.~\ref{fig:transformer_illustration}). The input sequence intersperses elements \(z_i\in \mathbb{Z}_n\) with the symbols ``\(+\)'' and ``\(=\)'', resulting in a total length of \(2M\). We denote \(e_i\), where $i \in \{0,...,n-1\}$ as the embedding for \(z_i \in \mathbb{Z}_n\), and let \(e_+\) and \(e_=\) represent the embeddings for the ``\(+\)" and  ``\(=\)'' tokens, respectively. If \(z_i\) appears in position \(m\) where $1\leq m \leq 2M$, its embedding is \(e_{i,m}\). Analogously, if a ``\(+\)'' or ``\(=\)'' symbol appears at position \(m\), its embedding is \(e_{+,m}\) or \(e_{=,m}\). These embeddings consist of  word embedding \(w_i\) and position embedding \(p_m\), as 
$e_{i,m} = [w_i, p_m]^\top ~\text{for  $i \in \{0,1,\ldots,n-1\} \cup \{+,=\}.$}$
We assume that all the vectors \(w_i\) and \(p_m\) are mutually orthogonal. In the following section, we illustrate how a language model can enforce commutativity for the ``\(+\)'' operator by providing a proof through construction. Specifically, we assign explicit values to the model’s weights and biases, assuming that these parameters can be learned from training. In the $\ell$-th attention layer, we denote $s_{m}^{(\ell-1)}$ as the hidden state at position $m$ from the previous attention layer. Let \(W_q, W_k, W_v\) and \(b_q, b_k, b_v\) denote the attention weights and biases that produce the query, key, and value, respectively, and let \(q_m, k_m,\) and \(v_m\) denote the query, key, and value vectors at position \(m\) in the $\ell$-th attention layer, respectively. Furthermore, we define \(s^{(\ell)}_{2M}\) as the hidden state in position \(2M\) after \(\ell\) attention layers. Namely,
$
s^{(\ell)}_{2M}=\sum_{i=1}^{2M}\sigma(\frac{q_{2M}^{\top}k_i}{\sum_{j=1}^{2M}q_{2M}^{\top}k_j})v_i.
$
The following theorem demonstrates that LLM can learn hidden states to achieve commutativity.
\begin{theorem}[\textbf{Commutativity--Invariant to the Input Permutations}]\label{theorem:commutative}
Given the LLMs' settings mentioned in Sec.\ref{sec:llm_explanation}, there exists a special assignment of the weights and biases $W_q,W_k,W_v$ and $b_q,b_k,b_v$ and specific assignment of embeddings $e_{i,m}$, for $i \in \{0,1,\ldots,n-1\} \cup \{+,=\}$, such that $s^{(\ell)}_{2M}$ could be invariant to the permutation of input elements $z_{i_1},z_{i_2},\cdots, z_{i_M} \in \mathbb{Z}_n$. However, this invariance holds only when the input contains commutative operators.
\end{theorem}
\proof The proof can be found in Sec.~\ref{sec:proof_of_theorem:commutativity}. \qedhere

With the invariance of the hidden states \(s^{(\ell)}_{2M}\), the subsequent layers of the transformer could serve as a classifier, mapping \(s^{(\ell)}_{2M}\) to the token \(z_{(i_1 + i_2 + \cdots + i_M)\bmod n}\), and hence endow the addition operation with commutativity. In addition to commutativity, we present a theorem demonstrating how an LLM can produce hidden states that remain essentially unchanged under the insertion of identity elements.

\begin{theorem}[\textbf{Identity--Invariant to the Insertion of Identity Tokens}]\label{theorem:identity}
Under the LLM settings in Sec.~\ref{sec:llm_explanation}, let \(s^{(\ell)}_{2M'}\) where $M'=M+1$ denote the hidden state after inserting an identity token \(z_0\) and an operator's token into the input sequence. There exists a specific assignment of weights and biases \(W_q, W_k, W_v\) and \(b_q, b_k, b_v\), together with particular embeddings \(e_{i,m}\) for $i \in \{0,1,\ldots,n-1\} \cup \{+,=\}$, such that \(s^{(\ell)}_{2M'}\) is equal to \(s^{(\ell)}_{2M}\). However, this property is valid only when the input includes operators for which \(z_0\) serves as the identity element.
\end{theorem}
\proof The proof can be found in Sec.~\ref{sec:proof_of_theorem:identity}. \qedhere

With this theorem, a classifier can interpret
$s_{2M'}^{(\ell)}$
as
$s_{2M}^{(\ell)}$,
which is already learned from \emph{base equation} in the training set. This ensures that the appending \(z_0\) does not alter the outcome, thus reflecting the identity property. 

\begin{remark}[Non-uniqueness of Weights and Bias Assignments]
Note that the weights, biases, and embeddings described in the proof of these theorems represent only one possible configuration to achieve commutativity and identity; many others could also be valid. However, it is critical that these properties be triggered specifically by operators with the properties of commutativity and ideneity, rather than by the operand tokens themselves. 
\end{remark}
Here we discuss a solution in which commutativity and identity arise without the existence of operators' embeddings.
\begin{remark}[Trivial Solution of Embeddings]
Language models may converge to a trivial embedding solution to achieve commutativity and identity. For instance, one might assign all non-identity tokens \(\bigl\{z_1, z_2, \ldots, z_{n-1}, +, =\bigr\}\) zero-valued position embeddings:
$
e_{i,m} = [w_i, 0]^\top ~ \text{for } i \in \{1,2,\ldots,n-1\} \cup \{+,=\},
$
and give the identity token \(z_0\) zero-valued word and position embeddings:
$
e_{0,m} = [\,0,\,0\,]^\top.
$
While this setup indeed satisfies both commutativity and identity, these properties are no longer tied to the  operator itself. Consequently, the model fails to produce correct outputs for operators lacking commutativity and identity—such as \(\ominus\), \(\triangleleft\), and \(\triangleright\)—when using these same trivial embeddings.
\end{remark}

\begin{figure*}[hbt!]
    \begin{center}
    \begin{subfigure}
        \centering
    \begin{tabular}{c}
        \centering
        \input{figures/convergence_3000_short}
    \end{tabular}
    \end{subfigure}
    \hfill
    \begin{subfigure}
        \centering
    \begin{tabular}{c c }
        \begin{tikzpicture}
\begin{axis}[
    width=4cm,
    height=3.2cm,
    xmode=log,
    xlabel={$K$}, 
   ylabel={accuracy},
    legend pos=north west,
    grid=major,
    grid style={dashed,gray!30},
    xmin=100, xmax=10000,
    ymin=0.3, ymax=1.05,
    title={},
    title style={font=\scriptsize},
    label style={font=\scriptsize},
    tick label style={font=\tiny},
    legend style={font=\tiny},
        xlabel style={
        at={(current axis.south east)}, 
        anchor=north east,              
        yshift=0pt,                   
        xshift=20pt                      
    },
]
\addplot[pc11, thick, dashed] table[row sep=\\] {
  x y \\ 
  100 1.0 \\  
  200 1.0 \\  
  300 1.0 \\  
  600 1.0 \\  
  1000 1.0 \\  
  2000 1.0 \\  
  3000 1.0 \\  
  6000 1.0 \\  
  10000 1.0 \\  
}; 
\addplot[pc12, thick, dashed] table[row sep=\\] {
  x y \\ 
  100 0.378 \\  
  200 0.378 \\  
  300 0.356 \\  
  600 0.4225 \\  
  1000 0.4785 \\  
  2000 0.698 \\  
  3000 0.894 \\  
  6000 0.9855 \\  
  10000 1.0 \\  
}; 
\addplot[pc13, thick, dashed] table[row sep=\\] {
  x y \\ 
  100 0.4615 \\  
  200 0.423 \\  
  300 0.6505000000000001 \\  
  600 0.95 \\  
  1000 1.0 \\  
  2000 0.9995 \\  
  3000 1.0 \\  
  6000 1.0 \\  
  10000 1.0 \\  
}; 
\addplot[pc21, thick, densely dotted] table[row sep=\\] {
  x y \\ 
  100 1.0 \\  
  200 1.0 \\  
  300 1.0 \\  
  600 1.0 \\  
  1000 1.0 \\  
  2000 1.0 \\  
  3000 1.0 \\  
  6000 1.0 \\  
  10000 1.0 \\  
}; 
\addplot[pc22, thick, densely dotted] table[row sep=\\] {
  x y \\ 
  100 0.526 \\  
  200 0.47050000000000003 \\  
  300 0.484 \\  
  600 0.5885 \\  
  1000 0.7035 \\  
  2000 0.7755000000000001 \\  
  3000 0.9305000000000001 \\  
  6000 0.9915 \\  
  10000 1.0 \\  
}; 
\addplot[pc23, thick, densely dotted] table[row sep=\\] {
  x y \\ 
  100 0.5215000000000001 \\  
  200 0.434 \\  
  300 0.597 \\  
  600 0.9635 \\  
  1000 0.9995 \\  
  2000 1.0 \\  
  3000 1.0 \\  
  6000 1.0 \\  
  10000 1.0 \\  
}; 
\addplot[pc31, very thick, loosely dotted] table[row sep=\\] {
  x y \\ 
  100 1.0 \\  
  200 1.0 \\  
  300 1.0 \\  
  600 1.0 \\  
  1000 1.0 \\  
  2000 1.0 \\  
  3000 1.0 \\  
  6000 1.0 \\  
  10000 1.0 \\  
}; 
\addplot[pc32, very thick, loosely dotted] table[row sep=\\] {
  x y \\ 
  100 0.5405 \\  
  200 0.6255 \\  
  300 0.8492500000000001 \\  
  600 0.991 \\  
  1000 1.0 \\  
  2000 1.0 \\  
  3000 1.0 \\  
  6000 1.0 \\  
  10000 1.0 \\  
}; 
\addplot[pc33,  very thick, loosely dotted] table[row sep=\\] {
  x y \\ 
  100 0.91675 \\  
  200 1.0 \\  
  300 1.0 \\  
  600 1.0 \\  
  1000 1.0 \\  
  2000 1.0 \\  
  3000 1.0 \\  
  6000 1.0 \\  
  10000 1.0 \\  
}; 
\end{axis} 

\end{tikzpicture}  &
         \begin{tikzpicture}
\begin{axis}[
    width=4cm,
    height=3.2cm,
    xmode=log,
    xlabel={$K$}, 
   ylabel={accuracy},
    legend pos=north west,
    grid=major,
    grid style={dashed,gray!30},
    xmin=100, xmax=20000,
    ymin=0.3, ymax=1.05,
    title={},
    title style={font=\scriptsize},
    label style={font=\scriptsize},
    tick label style={font=\tiny},
    legend style={font=\tiny},
        xlabel style={
        at={(current axis.south east)}, 
        anchor=north east,              
        yshift=0pt,                   
        xshift=20pt                      
    },
]
\addplot[pc11, thick, dashed] table[row sep=\\] {
  x y \\ 
  100 1.0 \\  
  200 1.0 \\  
  300 1.0 \\  
  600 1.0 \\  
  1000 1.0 \\  
  2000 1.0 \\  
  3000 1.0 \\  
  6000 1.0 \\  
  10000 1.0 \\  
  20000 1.0 \\  
  30000 1.0 \\  
}; 
\addplot[pc12, thick, dashed] table[row sep=\\] {
  x y \\ 
  100 0.425 \\  
  200 0.4735 \\  
  300 0.4935 \\  
  600 0.5385 \\  
  1000 0.597 \\  
  2000 0.7375 \\  
  3000 0.6984999999999999 \\  
  6000 0.8065 \\  
  10000 0.9835 \\  
  20000 0.989 \\  
  30000 0.9995 \\  
}; 
\addplot[pc13, thick, dashed] table[row sep=\\] {
  x y \\ 
  100 0.692 \\  
  200 0.6745000000000001 \\  
  300 0.8534999999999999 \\  
  600 0.966 \\  
  1000 0.998 \\  
  2000 1.0 \\  
  3000 1.0 \\  
  6000 1.0 \\  
  10000 1.0 \\  
  20000 1.0 \\  
  30000 1.0 \\  
}; 
\addplot[pc21, thick, densely dotted] table[row sep=\\] {
  x y \\ 
  100 1.0 \\  
  200 1.0 \\  
  300 1.0 \\  
  600 1.0 \\  
  1000 1.0 \\  
  2000 1.0 \\  
  3000 1.0 \\  
  6000 1.0 \\  
  10000 1.0 \\  
  20000 1.0 \\  
  30000 1.0 \\  
}; 
\addplot[pc22, thick, densely dotted] table[row sep=\\] {
  x y \\ 
  100 0.4665 \\  
  200 0.506 \\  
  300 0.498 \\  
  600 0.5700000000000001 \\  
  1000 0.58 \\  
  2000 0.8145 \\  
  3000 0.8525 \\  
  6000 0.893 \\  
  10000 0.992 \\  
  20000 0.9924999999999999 \\  
  30000 0.9944999999999999 \\  
}; 
\addplot[pc23, thick, densely dotted] table[row sep=\\] {
  x y \\ 
  100 0.773 \\  
  200 0.626 \\  
  300 0.8274999999999999 \\  
  600 0.9884999999999999 \\  
  1000 0.9964999999999999 \\  
  2000 1.0 \\  
  3000 1.0 \\  
  6000 1.0 \\  
  10000 1.0 \\  
  20000 1.0 \\  
  30000 1.0 \\  
}; 
\addplot[pc31, very thick, loosely dotted] table[row sep=\\] {
  x y \\ 
  100 1.0 \\  
  200 1.0 \\  
  300 1.0 \\  
  600 1.0 \\  
  1000 1.0 \\  
  2000 1.0 \\  
  3000 1.0 \\  
  6000 1.0 \\  
  10000 1.0 \\  
  20000 1.0 \\  
  30000 1.0 \\  
}; 
\addplot[pc32, very thick, loosely dotted] table[row sep=\\] {
  x y \\ 
  100 0.47275 \\  
  200 0.5245 \\  
  300 0.615 \\  
  600 0.9215 \\  
  1000 0.98925 \\  
  2000 1.0 \\  
  3000 1.0 \\  
  6000 1.0 \\  
  10000 1.0 \\  
  20000 1.0 \\  
  30000 1.0 \\  
}; 
\addplot[pc33,  very thick, loosely dotted] table[row sep=\\] {
  x y \\ 
  100 0.916875 \\  
  200 1.0 \\  
  300 1.0 \\  
  600 1.0 \\  
  1000 1.0 \\  
  2000 1.0 \\  
  3000 1.0 \\  
  6000 1.0 \\  
  10000 1.0 \\  
  20000 1.0 \\  
  30000 1.0 \\  
}; 
\end{axis} 

\end{tikzpicture}  
                   \begin{tikzpicture}
\begin{axis}[
    width=4cm,
    height=3.2cm,
    xmode=log,
    xlabel={$K$}, 
   ylabel={accuracy},
    legend pos=north west,
    grid=major,
    grid style={dashed,gray!30},
    xmin=100, xmax=20000,
    ymin=0.3, ymax=1.05,
    title={},
    title style={font=\scriptsize},
    label style={font=\scriptsize},
    tick label style={font=\tiny},
    legend style={font=\tiny},
        xlabel style={
        at={(current axis.south east)}, 
        anchor=north east,              
        yshift=0pt,                   
        xshift=20pt                      
    },
]
\addplot[pc11, thick, dashed] table[row sep=\\] {
  x y \\ 
  100 1.0 \\  
  200 1.0 \\  
  300 1.0 \\  
  600 1.0 \\  
  1000 1.0 \\  
  2000 1.0 \\  
  3000 1.0 \\  
  6000 1.0 \\  
  10000 1.0 \\  
  30000 1.0 \\  
}; 
\addplot[pc12, thick, dashed] table[row sep=\\] {
  x y \\ 
  100 0.473 \\  
  200 0.473 \\  
  300 0.4505 \\  
  600 0.5525 \\  
  1000 0.5754999999999999 \\  
  2000 0.6345000000000001 \\  
  3000 0.701 \\  
  6000 0.931 \\  
  10000 0.989 \\  
  30000 1.0 \\  
}; 
\addplot[pc13, thick, dashed] table[row sep=\\] {
  x y \\ 
  100 0.9390000000000001 \\  
  200 0.8915 \\  
  300 0.898 \\  
  600 0.993 \\  
  1000 0.988 \\  
  2000 1.0 \\  
  3000 1.0 \\  
  6000 1.0 \\  
  10000 1.0 \\  
  30000 1.0 \\  
}; 
\addplot[pc21, thick, densely dotted] table[row sep=\\] {
  x y \\ 
  100 1.0 \\  
  200 1.0 \\  
  300 1.0 \\  
  600 1.0 \\  
  1000 1.0 \\  
  2000 1.0 \\  
  3000 1.0 \\  
  6000 1.0 \\  
  10000 1.0 \\  
  30000 1.0 \\  
}; 
\addplot[pc22, thick, densely dotted] table[row sep=\\] {
  x y \\ 
  100 0.5075000000000001 \\  
  200 0.4635 \\  
  300 0.497 \\  
  600 0.5985 \\  
  1000 0.588 \\  
  2000 0.727 \\  
  3000 0.788 \\  
  6000 0.976 \\  
  10000 0.9924999999999999 \\  
  30000 1.0 \\  
}; 
\addplot[pc23, thick, densely dotted] table[row sep=\\] {
  x y \\ 
  100 0.8325 \\  
  200 0.9105000000000001 \\  
  300 0.906 \\  
  600 0.991 \\  
  1000 0.9964999999999999 \\  
  2000 1.0 \\  
  3000 1.0 \\  
  6000 1.0 \\  
  10000 1.0 \\  
  30000 1.0 \\  
}; 
\addplot[pc31, very thick, loosely dotted] table[row sep=\\] {
  x y \\ 
  100 1.0 \\  
  200 1.0 \\  
  300 1.0 \\  
  600 1.0 \\  
  1000 1.0 \\  
  2000 1.0 \\  
  3000 1.0 \\  
  6000 1.0 \\  
  10000 1.0 \\  
  30000 1.0 \\  
}; 
\addplot[pc32, very thick, loosely dotted] table[row sep=\\] {
  x y \\ 
  100 0.47824999999999995 \\  
  200 0.5115000000000001 \\  
  300 0.6112500000000001 \\  
  600 0.7915000000000001 \\  
  1000 0.97975 \\  
  2000 1.0 \\  
  3000 1.0 \\  
  6000 1.0 \\  
  10000 1.0 \\  
  30000 1.0 \\  
}; 
\addplot[pc33,  very thick, loosely dotted] table[row sep=\\] {
  x y \\ 
  100 0.91675 \\  
  200 1.0 \\  
  300 1.0 \\  
  600 1.0 \\  
  1000 1.0 \\  
  2000 1.0 \\  
  3000 1.0 \\  
  6000 1.0 \\  
  10000 1.0 \\  
  30000 1.0 \\  
}; 
\end{axis} 

\end{tikzpicture} 

    \end{tabular}
    \end{subfigure}
    \hfill
    \end{center}
    \caption{
    Plots of training and testing accuracy. The first row is the training dynamics for $\mathbb{Z}_7$ given the  scale of training set $K=3000$. The second row are the accuracies for $\mathbb{Z}_{7}$ (left), $\mathbb{Z}_{11}$ (middle), $\mathbb{Z}_{13}$ (right) with varying $K$ of training set.
}
    \label{fig:exp_accuracy}
\end{figure*}

\section{Experiments}

\paragraph{ Settings:}
We conduct our experiments using the datasets described in Sec.~\ref{sec:dataset}, Sec.~\ref{sec:additional_dataset_no_numerical} and \ref{sec:additional_dataset}, which encompass addition problems that test for commutativity and identity of operator ``$+$" and \(\oplus\), as well as operations involving \(\ominus\), \(\triangleleft\), and \(\triangleright\). We set $n=7,11$ and $13$ for $\mathbb{Z}_n$ and the number of input elements $M=6$. For the language model, we choose GPT-2 \cite{radford2019language} but reinitialize its weights before training to strip away any pre-existing knowledge, ensuring that the model acquires its understanding of algebraic structures solely from our data. In addition, we customize the tokenizer so that each element is represented as a single token (e.g., \(z_{10}\) becomes the token \(\texttt{[z10]}\) rather than several character-based tokens). For reproducibility, we have made our experimental code publicly available\footnote{\url{https://github.com/d09942015ntu/unraveling_llm_algebra}}.

\paragraph{Dataset Construction:}
We construct both training and testing sets by first choosing a scale \(K\), which determines the number of examples. Each training set or testing set with scale \(K\)  contains $10K$ instances, including
\begin{itemize}
    \item \(4K\) instances: Operators with commutativity and identity, including \(+\)'s commutativity, \(+\)'s identity, \(\oplus\)'s commutativity, and \(\oplus\)'s identity. Each of them contains $K$ instances.
    \item \(6K\) instances: Operator without commutativity and identity, including \(\ominus\),  \(\triangleleft\), and \(\triangleright\). Each of them contains $2K$ instances.
\end{itemize} 
An example illustrating both training and testing with \(K = 50\)  appears in Sec.~\ref{sec:example_dataset}. 
Throughout subsequent experiments, we fix the \(K=1000\) for the testing set, while \(K\) ranges from \(100\) to \(30{,}000\) for training set. 

\subsection{Results}
\subsubsection{Training Dynamics}  
We investigate how the training progresses for the case \(\mathbb{Z}_{7}\) when \(K=3000\) for the training set. 
The upper row of Fig.~\ref{fig:exp_accuracy} tracks the evolution of training and test accuracy over the course of training.
We observe that the model ultimately achieves 100\% accuracy in the training set, indicating that it has memorized all training instances.
However, for the commutativity property of ``\(+\)'' or  ``\(\oplus\)'' operators, it does not achieve high accuracy in testing set.
A plausible explanation is that the scale of the training set is still insufficient.
In the next experiment, we examine the testing accuracy for multiple training scales to investigate this further.

\subsubsection{Varying the Training Set's Scale.}
We vary the size of the training set from \(K=100\) to \(K=30{,}000\) and measure testing accuracy once both the training and testing accuracy have plateaued. The results, depicted in the second row of Fig.~\ref{fig:exp_accuracy}, yield the following observations.
\paragraph{All Tasks Achieve Over 99\% Testing Accuracy:}  
We find that \(\triangleleft\) and \(\triangleright\) are the most easily learned, each achieving 99\% testing accuracy with relatively few training samples. Next are \(\ominus\) and the identity properties of ``\(+\)'' and \(\oplus\), which converge to 100\% at around \(K=1000\). The most challenging part is learning the commutative properties of ``\(+\)" and $\oplus$, which requires \(K\) between \(10{,}000\) and \(20{,}000\) to reach over 99\% testing acuracy. Despite these differences, all tasks ultimately achieve over 99\% accuracy, suggesting that commutativity and identity learning is indeed operator-driven rather than a trivial result of embeddings.

\paragraph{Generalization of Commutative Operations.}
Despite the number of training instances required to achieve high accuracy appears large, it is still much smaller than the full combinatorial space of expressions like \(z_{i_1} + z_{i_2} + \dots + z_{i_6}\), which, for instance, includes \((7-1)^6 = 46{,}656\) possibilities in \(\mathbb{Z}_7\) and \((13-1)^6 = 2{,}985{,}984\) in \(\mathbb{Z}_{13}\). Thus, LLMs can actually learn and generalize commutativity for both ``\(+\)'' and \(\oplus\) without enumerating all possible permutations.

\paragraph{No Reliance on Numerical Computation}
We observe that ``\(+\)'' does not exceed \(\oplus\) in performance, indicating that LLMs learn commutativity and identity rather than relying on a direct numerical calculation. It also provides evidence that LLMs could not  acquire computation skills for numerical values if the numerical values are not explicitly specified in the input.

\begin{figure}[hbt!]
    \begin{tabular}[h]{c c c }
        \resizebox{.26\textwidth}{!}{
        \begin{tikzpicture}[scale=0.3] \foreach \y [count=\n] in 
{{0000,-011,-009,-009,-011,-013,-024,-035,-054,-060,-079,-104,0000},
{0000,-009,-022,-031,-042,-045,-060,-057,-085,-120,-155,-207,-010},
{0000,-015,-048,-075,-111,-128,-170,-192,-219,-235,-297,-370,-032},
{0000,0002,-018,-041,-075,-112,-148,-209,-275,-326,-397,-570,-052},
{0000,0020,-019,-064,-091,-124,-192,-265,-339,-392,-478,-632,-068}} {
          \foreach \x [count=\m] in \y {
               \ifnum \x < 0
                    \node[fill=yellow!\x!purple, minimum width=1.5mm, text=white] at (\m*0.8, -\n*0.8) {};
                \else
                     \node[fill=lime!\x!green, minimum width=1.5mm, text=white] at (\m*0.8, -\n*0.8) {};
                \fi
                  \ifnum \n < 2
                    \node[minimum size=4mm] at (\m*0.8, 0) {\tiny \m};
                \fi
      }
    }
  \foreach \a [count=\i] in {100,300,1000,3000,10000} {
    \node[minimum size=4mm] at (-0.5, -\i*0.8) {\tiny \a};

  }
\end{tikzpicture}
        }
        & \resizebox{.26\textwidth}{!}{
        \begin{tikzpicture}[scale=0.3] \foreach \y [count=\n] in 
{{0000,-016,-014,-010,0005,0023,0039,0061,0077,0064,0011,-070,-007},
{0000,-005,0000,0004,0015,0024,0044,0065,0063,0023,-069,-175,-014},
{0000,-010,-006,-002,-006,0000,0006,0014,0013,-023,-099,-221,-034},
{0000,0001,0001,-001,0001,0000,0002,-031,-059,-118,-189,-286,-044},
{0000,-001,-018,-029,-028,-026,-021,-027,-058,-123,-251,-414,-069}} {
          \foreach \x [count=\m] in \y {
               \ifnum \x < 0
                    \node[fill=yellow!\x!purple, minimum width=1.5mm, text=white] at (\m*0.8, -\n*0.8) {};
                \else
                     \node[fill=lime!\x!green, minimum width=1.5mm, text=white] at (\m*0.8, -\n*0.8) {};
                \fi
                  \ifnum \n < 2
                    \node[minimum size=4mm] at (\m*0.8, 0) {\tiny \m};
                \fi
      }
    }
  \foreach \a [count=\i] in {100,300,1000,3000,10000} {
    \node[minimum size=4mm] at (-0.5, -\i*0.8) {\tiny \a};

  }
\end{tikzpicture}
        }
        &\resizebox{.26\textwidth}{!}{
        \begin{tikzpicture}[scale=0.3] \foreach \y [count=\n] in 
{{0000,-019,-021,-019,0000,0019,0031,0053,0060,0039,-009,-073,-001},
{0000,-020,-015,-011,0000,0011,0031,0058,0037,0013,-056,-163,-010},
{0000,-027,-044,-052,-068,-063,-071,-059,-081,-129,-206,-291,-026},
{0000,-019,-033,-065,-075,-087,-112,-123,-165,-231,-343,-437,-045},
{0000,0000,-050,-073,-075,-077,-083,-099,-160,-238,-360,-520,-071}} {
          \foreach \x [count=\m] in \y {
               \ifnum \x < 0
                    \node[fill=yellow!\x!purple, minimum width=1.5mm, text=white] at (\m*0.8, -\n*0.8) {};
                \else
                     \node[fill=lime!\x!green, minimum width=1.5mm, text=white] at (\m*0.8, -\n*0.8) {};
                \fi
                  \ifnum \n < 2
                    \node[minimum size=4mm] at (\m*0.8, 0) {\tiny \m};
                \fi
      }
    }
  \foreach \a [count=\i] in {100,300,1000,3000,10000} {
    \node[minimum size=4mm] at (-0.5, -\i*0.8) {\tiny \a};

  }
\end{tikzpicture}
        }
        \\
        \resizebox{.26\textwidth}{!}{
        \begin{tikzpicture}[scale=0.3] \foreach \y [count=\n] in 
{{0000,-012,-003,0003,0004,0012,0014,0017,0015,0020,0017,0020,0007},
{0000,-020,-025,-027,-032,-032,-036,-026,-026,-035,-057,-078,0013},
{0000,-037,-073,-103,-124,-138,-154,-170,-178,-185,-215,-248,-009},
{0000,-010,-025,-040,-050,-066,-096,-126,-185,-216,-249,-311,-002},
{0000,0005,-091,-204,-219,-250,-287,-331,-361,-364,-372,-371,-022}} {
          \foreach \x [count=\m] in \y {
               \ifnum \x < 0
                    \node[fill=yellow!\x!purple, minimum width=1.5mm, text=white] at (\m*0.8, -\n*0.8) {};
                \else
                     \node[fill=lime!\x!green, minimum width=1.5mm, text=white] at (\m*0.8, -\n*0.8) {};
                \fi
                  \ifnum \n < 2
                    \node[minimum size=4mm] at (\m*0.8, 0) {\tiny \m};
                \fi
      }
    }
  \foreach \a [count=\i] in {100,300,1000,3000,10000} {
    \node[minimum size=4mm] at (-0.5, -\i*0.8) {\tiny \a};

  }
\end{tikzpicture}
        }
        
        &\resizebox{.26\textwidth}{!}{
        \begin{tikzpicture}[scale=0.3] \foreach \y [count=\n] in 
{{0000,-016,-014,-010,0005,0023,0039,0061,0077,0064,0011,-070,-007},
{0000,-005,0000,0004,0015,0024,0044,0065,0063,0023,-069,-175,-014},
{0000,-010,-006,-002,-006,0000,0006,0014,0013,-023,-099,-221,-034},
{0000,0001,0001,-001,0001,0000,0002,-031,-059,-118,-189,-286,-044},
{0000,-001,-018,-029,-028,-026,-021,-027,-058,-123,-251,-414,-069}} {
          \foreach \x [count=\m] in \y {
               \ifnum \x < 0
                    \node[fill=yellow!\x!purple, minimum width=1.5mm, text=white] at (\m*0.8, -\n*0.8) {};
                \else
                     \node[fill=lime!\x!green, minimum width=1.5mm, text=white] at (\m*0.8, -\n*0.8) {};
                \fi
                  \ifnum \n < 2
                    \node[minimum size=4mm] at (\m*0.8, 0) {\tiny \m};
                \fi
      }
    }
  \foreach \a [count=\i] in {100,300,1000,3000,10000} {
    \node[minimum size=4mm] at (-0.5, -\i*0.8) {\tiny \a};

  }
\end{tikzpicture}
        }
        & \resizebox{.26\textwidth}{!}{
        \begin{tikzpicture}[scale=0.3] \foreach \y [count=\n] in 
{{0000,-024,-009,0006,0029,0051,0063,0078,0092,0105,0121,0161,0050},
{0000,-040,-030,-024,-016,-008,0004,0022,0015,0016,0001,-016,0036},
{0000,-046,-054,-063,-059,-062,-069,-069,-077,-087,-121,-154,0003},
{0000,-041,-047,-069,-073,-080,-097,-093,-099,-114,-137,-151,0003},
{0000,-041,-114,-164,-177,-182,-187,-197,-213,-223,-256,-285,-023}} {
          \foreach \x [count=\m] in \y {
               \ifnum \x < 0
                    \node[fill=yellow!\x!purple, minimum width=1.5mm, text=white] at (\m*0.8, -\n*0.8) {};
                \else
                     \node[fill=lime!\x!green, minimum width=1.5mm, text=white] at (\m*0.8, -\n*0.8) {};
                \fi
                  \ifnum \n < 2
                    \node[minimum size=4mm] at (\m*0.8, 0) {\tiny \m};
                \fi
      }
    }
  \foreach \a [count=\i] in {100,300,1000,3000,10000} {
    \node[minimum size=4mm] at (-0.5, -\i*0.8) {\tiny \a};

  }
\end{tikzpicture}
        }
        \vspace{3px}
        \\
        \multicolumn{2}{c}{
        \resizebox{.40\textwidth}{!}{
         \begin{tikzpicture}[scale=0.4] \foreach \y [count=\n] in 
{{200,100,50,0,-1,-50,-100,-200,-400}} {
      \foreach \x [count=\m] in \y {
        \ifnum \x < 0
            \node[fill=yellow!\x!purple, minimum width=8mm, text=white] at (\m*2, -\n) {\scriptsize \textbf{\texttt{\x}}};
        \else
             \node[fill=lime!\x!green, minimum width=8mm, text=black] at (\m*2, -\n) {\scriptsize \textbf{\texttt{\x}}};
        \fi
      }
    }
\end{tikzpicture}

         } 
         }
    \end{tabular}
    \hfill
    \centering
    \caption{ 
    Visualization of $S^{\ell}_{\text{com}}$ and $S^{\ell}_{\text{ide}}$ where $1\leq \ell \leq 13$. The upper row displays the values of $\operatorname{S}^{\ell}_{\text{com}}(+,\ominus)$, $\operatorname{S}^{\ell}_{\text{com}}(+,\triangleleft)$, and $\operatorname{S}^{\ell}_{\text{com}}(+,\triangleright)$ and the lower row displays the values of $\operatorname{S}^{\ell}_{\text{ide}}(+,\ominus)$, $\operatorname{S}^{\ell}_{\text{ide}}(+,\triangleleft)$ and $\operatorname{S}^{\ell}_{\text{ide}}(+,\triangleright)$. The numbers in the left axis represent $K \in \{100,300,\cdots,10000\}$. For clarity, non-negative values are highlighted in green and yellow.}
    \label{fig:vis_com_operator}
\end{figure}

\subsection{Visualization of Hidden States}

\paragraph{Commutative:}
As pointed out in Sec.~\ref{sec:llm_explanation}, when the operator preserves commutativity,  the hidden states remain invariant under permuting the inputs. In practice, however, these states need not be identical; it suffices
that hidden states from different permutations be recognized as the same category. Consequently, given a set of input tokens, we expect slight variation (that is, a small standard deviation) in the hidden states across different permutations.
For example, assume that the inputs are \(z_{i_1}, z_{i_2}, \cdots, z_{i_m}\) and the output is \(y\). Let \(\odot\) denote an operator. Then, considering all permutations, we have
\[
\begin{aligned}
y_1=z_{i_1} \odot z_{i_2} \odot\cdots \odot z_{i_m},\quad
y_2=z_{i_2} \odot z_{i_1} \odot \cdots \odot z_{i_3},\quad
\cdots, \quad
y_m=z_{i_m} \odot\cdots\odot z_{i_2} \odot z_{i_1}.
\end{aligned}
\]
If we define \(s_i \in \mathbb{R}^{D}\) as the hidden state of each \(y_i\) where $D$ is the size of hidden states, then for a commutative operator \(\odot\), we would have \(y_1 = y_2 = \cdots = y_m\). Consequently, \(s_1, s_2, \dots, s_m\) should also be similar. The sum of their element-wise standard-deviation is defined by
$
\operatorname{S}_{\text{std}}(s_1,\ldots,s_m;\odot) 
= \sum_{k=1}^{D} \operatorname{std}\bigl(\{s_1\}_k,\dots,\{s_m\}_k\bigr),
$
where $\operatorname{std}(x_1,\cdots,x_m)$ denotes the standard deviation among $x_1,\cdots,x_m$,  \(\{s_1\}_k\) denotes the \(k\)-th element of \(s_1\), and \(\odot\) indicates that the hidden states are produced by the operator \(\odot\). For a non-commutative operator $\odot'$, these hidden states would differ more substantially, resulting in a higher value of $\operatorname{S}_{\text{std}}(s_1,\ldots,s_m;\odot') $.
The left column of Fig.~\ref{fig:vis_com_operator} shows the differences in $S_{\text{std}}$ between the “\(+\)” operator and various non-commutative operators, denoted as:
\begin{align*}
&\operatorname{S}_{\text{com}}^{(\ell)}(+,\odot')=
\operatorname{S}_{\text{std}}(s^{(\ell)}_1,\ldots,s^{(\ell)}_m; +) 
\quad - \operatorname{S}_{\text{std}}(s^{(\ell)}_1,\ldots,s^{(\ell)}_m; \odot'),~\text{where $\odot' \in \{\ominus,\triangleleft, \triangleright\}$}.
\end{align*}
Here, \(\ell\) denotes the layer index (GPT-2 has 13 layers), and each column of the heat map in Fig.~\ref{fig:vis_com_operator} corresponds to one of these layers. As the scale of the training set increases and the accuracy of the model in commutative operations improves, $\operatorname{S}_{\text{std}}$ for the commutative operator “\(+\)” becomes noticeably smaller compared to that of the non-commutative operators and consequently $\operatorname{S}_{\text{com}}(+,\odot')$ become more negative. 

\paragraph{Identity:} 
We consider  non-identity tokens \(z_1, z_2, \cdots, z_m\) and an identity token \(z_0\) to show the hidden states when the operator \(\odot\) remains invariant in the presence of an identity element. Concretely, we compare the outputs
\[
\begin{aligned}
\bar{y} = z_1 \odot z_2 \odot \cdots \odot z_m,
~
y_1 = z_0 \odot z_1 \odot z_2 \odot \cdots \odot z_m,~
 \cdots, ~
y_m = z_1 \odot z_2 \odot \cdots \odot  z_m \odot z_0.
\end{aligned}
\]
where $\bar{y}$ is the result without $z_0$. We denote \(\bar{s}\) and \(s_1, s_2, \dots, s_m\) as the hidden states of \(\bar{y}\) and \(y_1, y_2, \dots, y_m\), respectively. If $y_i$ for $i\in\{1,...,m\}$ and $\bar{y}$ the same under the insertion of identity elements.
Then, the distance between $\bar{s}$ and any of $s_i$ for $i\in\{1,...,m\}$ should be small. The sum of their distances is defined by
$
\operatorname{S}_{\text{dist}}(\bar{s},s_1,\ldots,s_m;\odot) 
= \sum_{k=1}^{D} \sum_{i=1}^{m}\left|\{s_i\}_k-\{\bar{s}\}_k\right|,
$
where \(\{s_i\}_k\) denotes the \(k\)-th element of \(s_i\), and \(\odot\) indicates that the hidden states are produced by the operator \(\odot\). For an operator without an identity element, denoted as $\odot'$, the distance between these two hidden states would be substantially larger, resulting in a larger value of $\operatorname{S}_{\text{dist}}(\bar{s},s_1,\ldots,s_m;\odot') $.
The right column of Fig.~\ref{fig:vis_com_operator} shows the differences in $S_{\text{dist}}$ between the “\(+\)” operator and various operators without identity elements, denoted as:
\begin{align*}
&\operatorname{S}_{\text{ide}}^{(\ell)}(+,\odot')=
\operatorname{S}_{\text{dist}}(\bar{s}^{(\ell)},s^{(\ell)}_1,\ldots,s^{(\ell)}_m; +) \quad - \operatorname{S}_{\text{dist}}(\bar{s}^{(\ell)}, s^{(\ell)}_1,\ldots,s^{(\ell)}_m; \odot'),~\text{where $\odot' \in \{\ominus,\triangleleft, \triangleright\}$}.
\end{align*}
As more training data is used and the model becomes better at identity-invariant operations, the value of \(\operatorname{S}_{\text{dist}}\) for the “\(+\)” operator decreases significantly compared to operators without identity elements, leading to more negative values of \(\operatorname{S}_{\text{ide}}(+,\odot')\).

\section{Limitations}
In this work, we assume our problem scope is limited to a finite Abelian group $\mathbb{Z}_5$, focusing exclusively on commutativity and identity. Other properties such as inverse and associativity, remain to be verified. Furthermore, real-world mathematical problems often involve real numbers and diverse forms of descriptions including natural language. Despite these limitations, we believe that our research takes the first step toward unraveling the mystery of the mathematical capabilities of LLMs.
On the other hand, we tested only a relatively small LLM, GPT-2. Nevertheless, we hypothesize that larger models, with greater expressive power, are also capable of capturing algebraic structures within training data.

\section{Conclusion}
\label{sec:conclusion}
We have demonstrated that LLMs can learn and internalize fundamental algebraic properties, especially commutativity and identity, purely from training data. Our strategy involved constructing a dataset of finite Abelian group expressions, ensuring that both commutative and identity instances appear in training and are held out for testing. Using a reinitialized GPT-2, we observed successful generalization to unseen tasks. 
Furthermore, We also provided a constructive proof showing how transformer-based models preserve invariance under permutations and identity insertion. Hidden-state visualizations revealed that operators preserving commutativity and identity produced more uniform internal representations compared to those that did not.
Although our experiments centered on finite Abelian groups and basic algebraic properties, these results indicate the potential for LLMs to acquire and generalize more intricate algebraic structures directly from data. Extensions to larger systems, real numbers, advanced group properties, and more natural language settings remain promising directions for future research.

\section*{Acknowledgment}
This work was supported in part by the Asian Office of Aerospace Research \& Development (AOARD) under Grant NTU-112HT911020,
National Science and Technology Council of Taiwan under Grant NSTC-112-2221-E-002-204- and NSTC-113-2221-E-002-208-,
Ministry of Education (MOE) of Taiwan under Grant NTU-113L891406, and
Ministry of Environment under Grant NTU-113BT911001.

\bibliography{sample}
\bibliographystyle{iclr2025_conference}

\newpage
\onecolumn
\appendix
\section{Appendix}

\subsection{Related Works}
\label{sec:related_works}
\paragraph{Theory of Chain-of-thought Reasoning in LLMs:}
Chain-of-Thought (CoT) techniques \cite{wei2022chain} empower large language models (LLMs) to tackle complex mathematical reasoning tasks by breaking solutions into sequential steps, making them essential for solving mathematical problems. Recent studies shed light on CoT's theoretical underpinnings. For example, \cite{prystawski2024think} models CoT with Bayesian networks, where questions, answers, and reasoning steps form interconnected nodes, demonstrating that structured reasoning improves LLM performance. \cite{xiao2024theory} introduces the concept of length generalization, showing that LLMs can extrapolate from simple examples to address more complex problems. Expanding the PAC learning framework, \cite{malach2023auto} shows that auto-regressive learners can effectively learn linear threshold circuits when CoT steps are provided. Additionally, \cite{feng2024towards} proves that CoT enables transformers to handle dynamic programming problems, even with polynomially increasing complexity.  Although these studies establish a theoretical basis for CoT, which decomposes complex mathematical problems into manageable steps, they rarely address how LLMs solve mathematical problems within a single step of CoT reasoning. 

\paragraph{Enhancing mathematical reasoning in LLMs:} Several recent works have developed different fine-tuning strategies to improve LLMs' mathematical reasoning. First, \cite{guo-etal-2024-exploring} mainly focuses on improving the “reversal curse” by introducing a reverse training task, thereby enhancing logical consistency. Similarly, \cite{zhou2024dual} enhances the CoT ability by introducing two auxiliary tasks, including Intermediate Reasoning State Prediction and Instruction Reconstruction task, which model mathematical reasoning from both forward and reverse direction. Moreover, \cite{liu2023improving} provides three fine-tuning methods to improve the LLMs' performance on mathematical problems. By utilizing the supervision signal of the evaluation tasks, these methods effectively improve the model performance in generating solutions for math problems. Meanwhile, \cite{yin2024scaffolding} proposes Scaffolding Learning, which first allows the model to master arithmetic operations and then fine-tunes it efficiently on the more general task of solving word math problems. Futhermore, \cite{lyu2024adapting} provides a two-component fine-tuning method, consisting of World Knowledge Distillation (WKD) and Tool Usage Adaptation (TUA). By leveraging these two components, the model surpasses state-of-the-art models such as GPT-4o in mathematical problem-solving. Finally, \cite{tang2024mathscale} proposes a method called MathScale, which is used to construct the fine-tuning dataset MathScaleQA to enhance mathematical reasoning capabilities. Additionally, MWPBENCH is introduced as a benchmark to systematically evaluate performance. Although mathematical reasoning in these works has been enhanced, the underlying principles behind the reasoning process remain unknown. This gap suggests the need for further exploration in how LLMs solve such problems. To better understand these mechanism, we should start with the fundamental aspects, such as arithmetic, to uncover their underlying principles. 


\paragraph{Improving Arithmetic ability in LLMs:}
The LLMs have demonstrated their power in natural language process tasks. However, they still exhibit limitations when it comes to performing arithmetic calculations. Recent studies have explored the application based on fine-tuning techniques to enhance the arithmetic capabilities of LLMs. For example,
\cite{liu2025arithmeticgpt} propose ArithemticGPT, which enhances advanced arithmetic calculation, such as exponentiation, logarithms, and trigonometric functions. Similarly, \cite{liu2023goat} propose supervised fine-tuning, mainly focuses on large-number arithmetic problem, particularly improving addition and developing decomposition strategies for multiplication and division. \cite{nye2021show} applied scratchpads fine-tuning, enabling the model to generalize to unseen 9-digit addition. In a different approach, \cite{zhang2024interpreting} examines the inner component responsible for arithmetic calculations and uses the precise fine-tuning to enhance the attention head values and MLPs within the associated components. While \cite{lai2024executing} fine-tunes LLMs to imitate Turing machine behavior, enabling step-by-step arithmetic calculations and enhancing their computational capability. Beyond fine-tuning, alternative methods have been proposed, \cite{shen2024revorder} apply RevOrder, a technique that reverses the arithmetic output order, to fine-tune LLMs, leading to a significant reduction in calculation errors. \cite{schwartz2024numerologic} incorporates digit length information as a prefix, enabling the model to better understand numerical magnitude, thereby improving its arithmetic performance. Despite these improvements, these studies primarily aim to enhance arithmetic capabilities, rather than understanding the fundamental principle of how LLMs could acquire arithematic ability.
Consequently, a deeper investigation into how LLMs internalize and generalize arithmetic concepts is still needed.

\paragraph{How Arithmetic Abilities Arise in LLMs:}
The mechanisms behind LLMs' arithmetic abilities remain debated. Some studies suggest that LLMs encode numerical values internally. \cite{Fangwei240103735} demonstrates this by using linear probes on addition problems, showing that number values are encoded across layers and can be extracted. On the other hand, \cite{levy2024language} finds that LLM errors are distributed across digits rather than numeric values, revealing that numbers are represented with per-digit circular structures in base 10. In addition, other works argue that LLMs rely on symbolic reasoning. \cite{deng2024language} shows LLMs learn simple patterns at the edges of a sequence of numbers faster than in the middle of the numbers of a sequence, indicating an easy-to-hard learning approach and symbolic arithmetic processing. \cite{hanna2024does} explores GPT-2 small's mechanism for predicting valid end years in date-related tasks, identifying a circuit responsible for "greater-than" comparisons that generalize across contexts. Additionally, \cite{stolfo2023mechanisticinterpretationarithmeticreasoning} demonstrates that LLMs transmit query-relevant information through attention mechanisms and process results with MLP modules, integrating them into the residual stream. Despite these insights, there is no consensus on whether LLMs primarily encode numerical values or rely on symbolic reasoning, highlighting the need for further research to clarify their mathematical processing mechanisms.

\paragraph{Machine Learning for Symmetric Discovery:}
The ability to discover symmetries enables machine learning models to uncover algebraic structures from training data. \cite{karjol2023neural} demonstrate that sub-groups can be identified through a neural network with a specially designed architecture, supported by a general theorem. They validate their approach with numerical experiments on tasks such as image-digit sum and symmetric polynomial regression. Similarly, \cite{karjol2024unified} present a unified framework for discovering symmetries across various subgroups, including locally symmetric, dihedral, and cyclic subgroups. Their architecture combines linear, matrix-valued, and non-linear functions to systematically capture invariance.  
\cite{yang2023latent} introduce Latent LieGAN (LaLiGAN), a generative model that maps data to a latent space where nonlinear symmetries become linear. LaLiGAN simultaneously learns the mapping and the latent space symmetries, theoretically proving its ability to express nonlinear symmetries under specific group action conditions.   
However, these works do not explore the connection between symmetry learning and large language models' (LLMs) arithmetic capabilities. While \cite{imani2024exploring} reveals that LLMs struggle with fundamental group properties and exhibit vulnerabilities in arithmetic reasoning, it does not investigate whether LLMs are possible to learn algebraic structures from training data. In contrast, our work demonstrates that LLMs can learn algebraic structures from training instances and generalize to solve unseen arithmetic problems.

\paragraph{LLMs can Learn Mathematical Rules:}
The work most closely related to ours is \cite{gorceix2024learningmathematicalruleslarge}, where the authors propose that LLMs can learn mathematical rules, such as distributivity or equation simplification. Although distributivity is also a type of algebraic structure, our work still has significant difference from them. First, our research demonstrates that LLMs can learn algebraic structures by training from scratch, showing that these rules are learned solely from the arithmetic equations we provide. We also rule out the possibility that LLMs learn these roles without any numerical computation. This differs from their approach, which relies on pre-trained models and cannot rule out the possibility that these rules were acquired from external materials or numerical computations. Additionally, our work further provides theoretical evidences of how transformers learn algebraic structures and providing an analysis of the hidden states of transformers, aspects that are not addressed in their study.


\subsection{Background Knowledge}
\begin{definition}\label{label:abelian_group}
A finite Abelian group is a set $\mathcal{G}$ with a binary operation $\circ$ that satisfies the following properties:
\begin{itemize}
    \item \textbf{Closure}: For any $a, b \in \mathcal{G}$, the results of $a \circ b$ are in $\mathcal{G}$.
    \item \textbf{Associativity}: For any $a, b, c \in \mathcal{G}$, $(a \circ b) \circ c = a \circ (b \circ c)$.
    \item \textbf{Commutativity}: For any $a, b \in \mathcal{G}$, $a \circ b = b \circ a$.
    \item \textbf{Identity}: There exists an element $e \in \mathcal{G}$ such that for any $a \in \mathcal{G}$, $a \circ e = e \circ a = a$.
    \item \textbf{Inverse}: For each $a \in \mathcal{G}$, there exists an element $a^{-1} \in \mathcal{G}$ such that $a \circ a^{-1} = a^{-1} \circ a = e$.
\end{itemize}
\end{definition}

\subsection{Proof of Theorems}\label{sec:proof_of_theorems}
\subsubsection{Proof of Theorem~\ref{theorem:commutative}}\label{sec:proof_of_theorem:commutativity}
\proof
Here, we prove this theorem by constructing a concrete example. In this example, the layer index $\ell=1$, and hence the hidden state of the previous layer $s_{m}^{(\ell-1)}$ at position $m$ is the input embedding $e_{i,m}$.  We choose \(W_v\) to be the identity matrix, so the value vector at position $m$ is simply:
\[
v_m = W_v   e_{i,m} = e_{i,m}.
\]
To enforce commutativity for ``\(+\)'', we assign the embedding of ``\(+\)'' at position \(m\) as
\[
e_{+,m} = \bigl[w_{+}, -\infty\bigr]^\top,
\]
where \(-\infty\) represents the smallest floating-point value. Additionally, we set both \(W_q\) and \(W_k\) to zero matrices and make the biases \(b_q\) and \(b_k\) all ones. We assume that the context window size of attention is $L$, where $L\gg 2M$ and the remaining positions $2M < m \leq L$ are padded with zero embeddings 
$$
e_{m} = \left[0,0\right]^{\top}\quad\text{ for all $2M<m\leq L$}.
$$ 
Under these settings, the attention weights in the first layer become uniform across positions, i.e.,
\[
\sigma\!\Bigl(\frac{q_{m}^{\top} k_i}{\sum_{j=1}^{L} q_{m}^{\top} k_j}\Bigr)
= \frac{1}{L}
\quad\text{for all } 1 \le i,m \le L.
\]
Thus, the hidden state of the ``\(=\)'' token at position \(2M\) after the first attention layer, denoted \(s^{(1)}_{2M}\), is
\[
\begin{aligned}
s^{(1)}_{2M}
= \sum_{i=1}^{2M} 
\sigma\!\Bigl(\frac{q_{2M}^{\top} k_i}{\sum_{j=1}^{2M} q_{2M}^{\top} k_j}\Bigr)  
v_i 
= \sum_{m=1}^{2M} \frac{1}{L}   e_{i,m} 
= \Bigl[\frac{1}{L}\sum_{m=1}^{2M} w_{i_m},  -\infty\Bigr]^\top.
\end{aligned}
\]
Since 
\(\bigl[\frac{1}{L}\sum_{m=1}^{2M} w_{i_m},  -\infty\bigr]^\top\)
is invariant to the position embeddings of \(e_{i_1}, \dots, e_{i_M}\) relative to \(e_{+}\), this hidden state does not depend on the order of the input tokens \(z_{i_1}, \dots, z_{i_M}\). 
\qedhere
\subsubsection{Proof of Theorem~\ref{theorem:identity}}\label{sec:proof_of_theorem:identity}
\proof
Building on the setup from the previous section, let each non-identity token \(z_i\) (\(i \neq 0\)) at position \(m\) have the embedding 
\[
e_{i,m} = [w_i, 0, p_m]^\top
\quad
\text{for}
\quad
i \in \{1,2,\ldots,n-1\},
\]
where \(w_i\) and \(p_m\) are mutually orthogonal vectors. We then define the embedding of the identity token \(z_0\) at position \(m\) as
\[
e_{0,m} = [ 0,  w_0,  p_{m}]^\top,
\]
and the embedding of the addition operator ``\(+\)" at position \(m\) as
\[
e_{+,m} = [ 0, -\infty, -\infty]^\top.
\]
When we append \(z_0\) with an extra ``\(+\)" operator to the sequence of input tokens, the position of ``$=$" become $2M'=2M+2$, and the hidden state after the attention layer at position $2M'$ is
\[
\begin{aligned}
s^{(1)}_{2M'}=\frac{\sum_{m=1}^{2M} e_{i,m}+e_{+,2M+1}+e_{0,2M+2}}{L} =  
\Bigl[
\frac{1}{L} 
\sum_{m=1}^{2M} w_{i_m}, 
-\infty, 
-\infty
\Bigr]^\top,
\end{aligned}
\]
and the hidden states before inserting $z_0$ is
\[
s^{(1)}_{2M} = \frac{\sum_{m=1}^{2M} e_{i,m}}{L} =
\Bigl[\frac{1}{L}
\sum_{m=1}^{2M} w_{i_m}, 
-\infty, 
-\infty
\Bigr]^\top=s_{2M}^{(1)}.
\]
\qedhere

\subsection{Example of Dataset for $\mathbb{Z}_7$ with $K=50$} 
\label{sec:example_dataset}

\paragraph{\color{black}{Training, for Operator $+$'s Commutativity}}\mbox{}\\
{\footnotesize \color{dc7}{$z_{2}+z_{2}+z_{4}+z_{3}+z_{6}+z_{4}=z_{0}$} \quad
\color{dc7}{$z_{4}+z_{3}+z_{6}+z_{4}+z_{2}+z_{2}=z_{0}$} \quad
\color{dc7}{$z_{3}+z_{6}+z_{4}+z_{4}+z_{2}+z_{2}=z_{0}$} \quad
\color{dc7}{$z_{6}+z_{2}+z_{3}+z_{2}+z_{4}+z_{4}=z_{0}$} \quad
\color{dc7}{$z_{2}+z_{3}+z_{4}+z_{2}+z_{6}+z_{4}=z_{0}$} \quad
\color{dc7}{$z_{2}+z_{6}+z_{4}+z_{4}+z_{3}+z_{2}=z_{0}$} \quad
\color{dc7}{$z_{6}+z_{2}+z_{4}+z_{2}+z_{4}+z_{3}=z_{0}$} \quad
\color{dc7}{$z_{4}+z_{2}+z_{6}+z_{2}+z_{3}+z_{4}=z_{0}$} \quad
\color{dc7}{$z_{6}+z_{2}+z_{2}+z_{3}+z_{4}+z_{4}=z_{0}$} \quad
\color{dc7}{$z_{6}+z_{4}+z_{2}+z_{2}+z_{3}+z_{4}=z_{0}$} \quad
\color{dc8}{$z_{6}+z_{6}+z_{4}+z_{4}+z_{2}+z_{3}=z_{4}$} \quad
\color{dc8}{$z_{6}+z_{4}+z_{2}+z_{6}+z_{4}+z_{3}=z_{4}$} \quad
\color{dc8}{$z_{4}+z_{6}+z_{2}+z_{3}+z_{6}+z_{4}=z_{4}$} \quad
\color{dc8}{$z_{4}+z_{2}+z_{6}+z_{4}+z_{6}+z_{3}=z_{4}$} \quad
\color{dc8}{$z_{6}+z_{4}+z_{6}+z_{3}+z_{2}+z_{4}=z_{4}$} \quad
\color{dc8}{$z_{4}+z_{4}+z_{6}+z_{2}+z_{3}+z_{6}=z_{4}$} \quad
\color{dc8}{$z_{6}+z_{3}+z_{4}+z_{6}+z_{4}+z_{2}=z_{4}$} \quad
\color{dc8}{$z_{3}+z_{6}+z_{2}+z_{4}+z_{4}+z_{6}=z_{4}$} \quad
\color{dc8}{$z_{6}+z_{3}+z_{4}+z_{4}+z_{6}+z_{2}=z_{4}$} \quad
\color{dc8}{$z_{6}+z_{2}+z_{4}+z_{4}+z_{3}+z_{6}=z_{4}$} \quad
\color{dc9}{$z_{5}+z_{5}+z_{6}+z_{2}+z_{3}+z_{3}=z_{3}$} \quad
\color{dc9}{$z_{5}+z_{2}+z_{3}+z_{5}+z_{6}+z_{3}=z_{3}$} \quad
\color{dc9}{$z_{6}+z_{2}+z_{3}+z_{5}+z_{3}+z_{5}=z_{3}$} \quad
\color{dc9}{$z_{5}+z_{2}+z_{3}+z_{3}+z_{5}+z_{6}=z_{3}$} \quad
\color{dc9}{$z_{3}+z_{5}+z_{5}+z_{2}+z_{3}+z_{6}=z_{3}$} \quad
\color{dc9}{$z_{5}+z_{6}+z_{3}+z_{2}+z_{3}+z_{5}=z_{3}$} \quad
\color{dc9}{$z_{2}+z_{5}+z_{3}+z_{3}+z_{6}+z_{5}=z_{3}$} \quad
\color{dc9}{$z_{6}+z_{2}+z_{3}+z_{3}+z_{5}+z_{5}=z_{3}$} \quad
\color{dc9}{$z_{3}+z_{5}+z_{3}+z_{6}+z_{5}+z_{2}=z_{3}$} \quad
\color{dc9}{$z_{6}+z_{3}+z_{2}+z_{3}+z_{5}+z_{5}=z_{3}$} \quad
\color{dc10}{$z_{5}+z_{5}+z_{5}+z_{3}+z_{5}+z_{2}=z_{4}$} \quad
\color{dc10}{$z_{5}+z_{5}+z_{2}+z_{5}+z_{3}+z_{5}=z_{4}$} \quad
\color{dc10}{$z_{3}+z_{5}+z_{5}+z_{2}+z_{5}+z_{5}=z_{4}$} \quad
\color{dc10}{$z_{5}+z_{2}+z_{5}+z_{5}+z_{3}+z_{5}=z_{4}$} \quad
\color{dc10}{$z_{5}+z_{3}+z_{2}+z_{5}+z_{5}+z_{5}=z_{4}$} \quad
\color{dc10}{$z_{3}+z_{5}+z_{5}+z_{5}+z_{2}+z_{5}=z_{4}$} \quad
\color{dc10}{$z_{5}+z_{5}+z_{3}+z_{5}+z_{5}+z_{2}=z_{4}$} \quad
\color{dc10}{$z_{5}+z_{5}+z_{5}+z_{3}+z_{2}+z_{5}=z_{4}$} \quad
\color{dc10}{$z_{5}+z_{2}+z_{3}+z_{5}+z_{5}+z_{5}=z_{4}$} \quad
\color{dc10}{$z_{2}+z_{5}+z_{3}+z_{5}+z_{5}+z_{5}=z_{4}$} \quad
\color{dc11}{$z_{6}+z_{6}+z_{2}+z_{5}+z_{5}+z_{3}=z_{6}$} \quad
\color{dc11}{$z_{6}+z_{6}+z_{5}+z_{2}+z_{5}+z_{3}=z_{6}$} \quad
\color{dc11}{$z_{5}+z_{6}+z_{3}+z_{2}+z_{5}+z_{6}=z_{6}$} \quad
\color{dc11}{$z_{5}+z_{3}+z_{2}+z_{6}+z_{6}+z_{5}=z_{6}$} \quad
\color{dc1}{$z_{6}+z_{5}+z_{4}+z_{1}+z_{2}+z_{4}=z_{1}$} \quad
\color{dc2}{$z_{6}+z_{2}+z_{3}+z_{5}+z_{6}+z_{1}=z_{2}$} \quad
\color{dc3}{$z_{4}+z_{6}+z_{2}+z_{2}+z_{4}+z_{5}=z_{2}$} \quad
\color{dc4}{$z_{2}+z_{3}+z_{6}+z_{1}+z_{5}+z_{5}=z_{1}$} \quad
\color{dc5}{$z_{5}+z_{4}+z_{5}+z_{5}+z_{4}+z_{2}=z_{4}$} \quad
\color{dc6}{$z_{4}+z_{1}+z_{1}+z_{4}+z_{5}+z_{1}=z_{2}$} \quad
}

\paragraph{\color{black}{Testing, for Operator $+$'s Commutativity}}\mbox{}\\
{\footnotesize \color{dc1}{$z_{4}+z_{4}+z_{5}+z_{6}+z_{2}+z_{1}=z_{1}$} \quad
\color{dc1}{$z_{4}+z_{2}+z_{6}+z_{5}+z_{4}+z_{1}=z_{1}$} \quad
\color{dc1}{$z_{4}+z_{4}+z_{2}+z_{1}+z_{6}+z_{5}=z_{1}$} \quad
\color{dc1}{$z_{6}+z_{2}+z_{1}+z_{5}+z_{4}+z_{4}=z_{1}$} \quad
\color{dc1}{$z_{6}+z_{4}+z_{1}+z_{4}+z_{2}+z_{5}=z_{1}$} \quad
\color{dc1}{$z_{5}+z_{2}+z_{4}+z_{4}+z_{6}+z_{1}=z_{1}$} \quad
\color{dc1}{$z_{4}+z_{5}+z_{4}+z_{6}+z_{1}+z_{2}=z_{1}$} \quad
\color{dc1}{$z_{6}+z_{5}+z_{1}+z_{4}+z_{2}+z_{4}=z_{1}$} \quad
\color{dc1}{$z_{1}+z_{2}+z_{4}+z_{6}+z_{5}+z_{4}=z_{1}$} \quad
\color{dc2}{$z_{6}+z_{1}+z_{6}+z_{2}+z_{3}+z_{5}=z_{2}$} \quad
\color{dc2}{$z_{2}+z_{5}+z_{3}+z_{1}+z_{6}+z_{6}=z_{2}$} \quad
\color{dc2}{$z_{3}+z_{2}+z_{6}+z_{1}+z_{5}+z_{6}=z_{2}$} \quad
\color{dc2}{$z_{6}+z_{2}+z_{5}+z_{1}+z_{3}+z_{6}=z_{2}$} \quad
\color{dc2}{$z_{6}+z_{6}+z_{3}+z_{5}+z_{2}+z_{1}=z_{2}$} \quad
\color{dc2}{$z_{1}+z_{6}+z_{5}+z_{6}+z_{3}+z_{2}=z_{2}$} \quad
\color{dc2}{$z_{5}+z_{1}+z_{6}+z_{3}+z_{6}+z_{2}=z_{2}$} \quad
\color{dc2}{$z_{5}+z_{6}+z_{6}+z_{3}+z_{1}+z_{2}=z_{2}$} \quad
\color{dc2}{$z_{5}+z_{3}+z_{6}+z_{2}+z_{6}+z_{1}=z_{2}$} \quad
\color{dc3}{$z_{2}+z_{4}+z_{5}+z_{2}+z_{4}+z_{6}=z_{2}$} \quad
\color{dc3}{$z_{4}+z_{5}+z_{4}+z_{2}+z_{2}+z_{6}=z_{2}$} \quad
\color{dc3}{$z_{2}+z_{5}+z_{2}+z_{4}+z_{4}+z_{6}=z_{2}$} \quad
\color{dc3}{$z_{5}+z_{2}+z_{6}+z_{4}+z_{2}+z_{4}=z_{2}$} \quad
\color{dc3}{$z_{5}+z_{4}+z_{6}+z_{2}+z_{4}+z_{2}=z_{2}$} \quad
\color{dc3}{$z_{2}+z_{4}+z_{2}+z_{6}+z_{4}+z_{5}=z_{2}$} \quad
\color{dc3}{$z_{4}+z_{2}+z_{4}+z_{5}+z_{6}+z_{2}=z_{2}$} \quad
\color{dc3}{$z_{2}+z_{4}+z_{5}+z_{6}+z_{4}+z_{2}=z_{2}$} \quad
\color{dc3}{$z_{4}+z_{5}+z_{2}+z_{2}+z_{4}+z_{6}=z_{2}$} \quad
\color{dc4}{$z_{2}+z_{1}+z_{6}+z_{5}+z_{5}+z_{3}=z_{1}$} \quad
\color{dc4}{$z_{6}+z_{5}+z_{3}+z_{1}+z_{5}+z_{2}=z_{1}$} \quad
\color{dc4}{$z_{6}+z_{2}+z_{5}+z_{5}+z_{1}+z_{3}=z_{1}$} \quad
\color{dc4}{$z_{3}+z_{6}+z_{1}+z_{5}+z_{2}+z_{5}=z_{1}$} \quad
\color{dc4}{$z_{5}+z_{2}+z_{5}+z_{1}+z_{6}+z_{3}=z_{1}$} \quad
\color{dc4}{$z_{3}+z_{1}+z_{6}+z_{2}+z_{5}+z_{5}=z_{1}$} \quad
\color{dc4}{$z_{3}+z_{6}+z_{5}+z_{2}+z_{5}+z_{1}=z_{1}$} \quad
\color{dc4}{$z_{6}+z_{5}+z_{1}+z_{3}+z_{2}+z_{5}=z_{1}$} \quad
\color{dc4}{$z_{1}+z_{5}+z_{5}+z_{2}+z_{6}+z_{3}=z_{1}$} \quad
\color{dc5}{$z_{2}+z_{5}+z_{5}+z_{5}+z_{4}+z_{4}=z_{4}$} \quad
\color{dc5}{$z_{4}+z_{5}+z_{5}+z_{2}+z_{5}+z_{4}=z_{4}$} \quad
\color{dc5}{$z_{2}+z_{5}+z_{5}+z_{4}+z_{5}+z_{4}=z_{4}$} \quad
\color{dc5}{$z_{2}+z_{5}+z_{4}+z_{5}+z_{5}+z_{4}=z_{4}$} \quad
\color{dc5}{$z_{5}+z_{4}+z_{5}+z_{5}+z_{2}+z_{4}=z_{4}$} \quad
\color{dc5}{$z_{2}+z_{4}+z_{5}+z_{4}+z_{5}+z_{5}=z_{4}$} \quad
\color{dc5}{$z_{5}+z_{4}+z_{2}+z_{5}+z_{5}+z_{4}=z_{4}$} \quad
\color{dc5}{$z_{5}+z_{4}+z_{5}+z_{2}+z_{5}+z_{4}=z_{4}$} \quad
\color{dc5}{$z_{5}+z_{2}+z_{4}+z_{5}+z_{5}+z_{4}=z_{4}$} \quad
\color{dc6}{$z_{4}+z_{1}+z_{4}+z_{5}+z_{1}+z_{1}=z_{2}$} \quad
\color{dc6}{$z_{1}+z_{5}+z_{4}+z_{1}+z_{1}+z_{4}=z_{2}$} \quad
\color{dc6}{$z_{1}+z_{1}+z_{4}+z_{1}+z_{5}+z_{4}=z_{2}$} \quad
\color{dc6}{$z_{4}+z_{1}+z_{1}+z_{5}+z_{4}+z_{1}=z_{2}$} \quad
\color{dc6}{$z_{1}+z_{5}+z_{1}+z_{4}+z_{1}+z_{4}=z_{2}$} \quad

\paragraph{\color{black}{Training, for Operator $+$'s Identity}}\mbox{}\\
{\footnotesize \color{dc0}{$z_{0}+z_{4}+z_{3}+z_{5}+z_{3}+z_{1}=z_{2}$} \quad
\color{dc0}{$z_{4}+z_{0}+z_{3}+z_{5}+z_{3}+z_{1}=z_{2}$} \quad
\color{dc0}{$z_{4}+z_{3}+z_{0}+z_{5}+z_{3}+z_{1}=z_{2}$} \quad
\color{dc0}{$z_{4}+z_{3}+z_{5}+z_{0}+z_{3}+z_{1}=z_{2}$} \quad
\color{dc0}{$z_{4}+z_{3}+z_{5}+z_{3}+z_{0}+z_{1}=z_{2}$} \quad
\color{dc0}{$z_{4}+z_{3}+z_{5}+z_{3}+z_{1}+z_{0}=z_{2}$} \quad
\color{dc0}{$z_{4}+z_{3}+z_{5}+z_{3}+z_{1}=z_{2}\quad\quad$} \quad
\color{dc1}{$z_{0}+z_{1}+z_{5}+z_{6}+z_{6}+z_{1}=z_{5}$} \quad
\color{dc1}{$z_{1}+z_{0}+z_{5}+z_{6}+z_{6}+z_{1}=z_{5}$} \quad
\color{dc1}{$z_{1}+z_{5}+z_{0}+z_{6}+z_{6}+z_{1}=z_{5}$} \quad
\color{dc1}{$z_{1}+z_{5}+z_{6}+z_{0}+z_{6}+z_{1}=z_{5}$} \quad
\color{dc1}{$z_{1}+z_{5}+z_{6}+z_{6}+z_{0}+z_{1}=z_{5}$} \quad
\color{dc1}{$z_{1}+z_{5}+z_{6}+z_{6}+z_{1}+z_{0}=z_{5}$} \quad
\color{dc1}{$z_{1}+z_{5}+z_{6}+z_{6}+z_{1}=z_{5}\quad\quad$} \quad
\color{dc2}{$z_{0}+z_{5}+z_{2}+z_{5}+z_{2}+z_{3}=z_{3}$} \quad
\color{dc2}{$z_{5}+z_{0}+z_{2}+z_{5}+z_{2}+z_{3}=z_{3}$} \quad
\color{dc2}{$z_{5}+z_{2}+z_{0}+z_{5}+z_{2}+z_{3}=z_{3}$} \quad
\color{dc2}{$z_{5}+z_{2}+z_{5}+z_{0}+z_{2}+z_{3}=z_{3}$} \quad
\color{dc2}{$z_{5}+z_{2}+z_{5}+z_{2}+z_{0}+z_{3}=z_{3}$} \quad
\color{dc2}{$z_{5}+z_{2}+z_{5}+z_{2}+z_{3}+z_{0}=z_{3}$} \quad
\color{dc2}{$z_{5}+z_{2}+z_{5}+z_{2}+z_{3}=z_{3}\quad\quad$} \quad
\color{dc3}{$z_{0}+z_{3}+z_{1}+z_{2}+z_{2}+z_{2}=z_{3}$} \quad
\color{dc3}{$z_{3}+z_{0}+z_{1}+z_{2}+z_{2}+z_{2}=z_{3}$} \quad
\color{dc3}{$z_{3}+z_{1}+z_{0}+z_{2}+z_{2}+z_{2}=z_{3}$} \quad
\color{dc3}{$z_{3}+z_{1}+z_{2}+z_{0}+z_{2}+z_{2}=z_{3}$} \quad
\color{dc3}{$z_{3}+z_{1}+z_{2}+z_{2}+z_{0}+z_{2}=z_{3}$} \quad
\color{dc3}{$z_{3}+z_{1}+z_{2}+z_{2}+z_{2}+z_{0}=z_{3}$} \quad
\color{dc3}{$z_{3}+z_{1}+z_{2}+z_{2}+z_{2}=z_{3}\quad\quad$} \quad
\color{dc4}{$z_{0}+z_{2}+z_{4}+z_{4}+z_{3}+z_{4}=z_{3}$} \quad
\color{dc4}{$z_{2}+z_{0}+z_{4}+z_{4}+z_{3}+z_{4}=z_{3}$} \quad
\color{dc4}{$z_{2}+z_{4}+z_{0}+z_{4}+z_{3}+z_{4}=z_{3}$} \quad
\color{dc4}{$z_{2}+z_{4}+z_{4}+z_{0}+z_{3}+z_{4}=z_{3}$} \quad
\color{dc4}{$z_{2}+z_{4}+z_{4}+z_{3}+z_{0}+z_{4}=z_{3}$} \quad
\color{dc4}{$z_{2}+z_{4}+z_{4}+z_{3}+z_{4}+z_{0}=z_{3}$} \quad
\color{dc4}{$z_{2}+z_{4}+z_{4}+z_{3}+z_{4}=z_{3}\quad\quad$} \quad
\color{dc5}{$z_{0}+z_{1}+z_{4}+z_{6}+z_{5}+z_{2}=z_{4}$} \quad
\color{dc5}{$z_{1}+z_{0}+z_{4}+z_{6}+z_{5}+z_{2}=z_{4}$} \quad
\color{dc5}{$z_{1}+z_{4}+z_{0}+z_{6}+z_{5}+z_{2}=z_{4}$} \quad
\color{dc5}{$z_{1}+z_{4}+z_{6}+z_{0}+z_{5}+z_{2}=z_{4}$} \quad
\color{dc5}{$z_{1}+z_{4}+z_{6}+z_{5}+z_{0}+z_{2}=z_{4}$} \quad
\color{dc5}{$z_{1}+z_{4}+z_{6}+z_{5}+z_{2}=z_{4}\quad\quad$} \quad
\color{dc11}{$z_{5}+z_{6}+z_{1}+z_{4}+z_{4}=z_{6}\quad\quad$} \quad
\color{dc12}{$z_{4}+z_{2}+z_{4}+z_{6}+z_{3}=z_{5}\quad\quad$} \quad
\color{dc13}{$z_{5}+z_{1}+z_{1}+z_{5}+z_{3}=z_{1}\quad\quad$} \quad
\color{dc14}{$z_{2}+z_{1}+z_{2}+z_{6}+z_{2}=z_{6}\quad\quad$} \quad
\color{dc15}{$z_{6}+z_{1}+z_{2}+z_{5}+z_{4}=z_{4}\quad\quad$} \quad
\color{dc16}{$z_{1}+z_{4}+z_{6}+z_{1}+z_{3}=z_{1}\quad\quad$} \quad
\color{dc17}{$z_{4}+z_{1}+z_{2}+z_{4}+z_{6}=z_{3}\quad\quad$} \quad
\color{dc18}{$z_{4}+z_{4}+z_{1}+z_{2}+z_{2}=z_{6}\quad\quad$} \quad
\color{dc19}{$z_{2}+z_{1}+z_{3}+z_{5}+z_{4}=z_{1}\quad\quad$} \quad
}

\paragraph{\color{black}{Testing, for Operator $+$'s Identity}}\mbox{}\\
{\footnotesize \color{dc11}{$z_{0}+z_{5}+z_{6}+z_{1}+z_{4}+z_{4}=z_{6}$} \quad
\color{dc11}{$z_{5}+z_{0}+z_{6}+z_{1}+z_{4}+z_{4}=z_{6}$} \quad
\color{dc11}{$z_{5}+z_{6}+z_{0}+z_{1}+z_{4}+z_{4}=z_{6}$} \quad
\color{dc11}{$z_{5}+z_{6}+z_{1}+z_{0}+z_{4}+z_{4}=z_{6}$} \quad
\color{dc11}{$z_{5}+z_{6}+z_{1}+z_{4}+z_{0}+z_{4}=z_{6}$} \quad
\color{dc11}{$z_{5}+z_{6}+z_{1}+z_{4}+z_{4}+z_{0}=z_{6}$} \quad
\color{dc12}{$z_{0}+z_{4}+z_{2}+z_{4}+z_{6}+z_{3}=z_{5}$} \quad
\color{dc12}{$z_{4}+z_{0}+z_{2}+z_{4}+z_{6}+z_{3}=z_{5}$} \quad
\color{dc12}{$z_{4}+z_{2}+z_{0}+z_{4}+z_{6}+z_{3}=z_{5}$} \quad
\color{dc12}{$z_{4}+z_{2}+z_{4}+z_{0}+z_{6}+z_{3}=z_{5}$} \quad
\color{dc12}{$z_{4}+z_{2}+z_{4}+z_{6}+z_{0}+z_{3}=z_{5}$} \quad
\color{dc12}{$z_{4}+z_{2}+z_{4}+z_{6}+z_{3}+z_{0}=z_{5}$} \quad
\color{dc13}{$z_{0}+z_{5}+z_{1}+z_{1}+z_{5}+z_{3}=z_{1}$} \quad
\color{dc13}{$z_{5}+z_{0}+z_{1}+z_{1}+z_{5}+z_{3}=z_{1}$} \quad
\color{dc13}{$z_{5}+z_{1}+z_{0}+z_{1}+z_{5}+z_{3}=z_{1}$} \quad
\color{dc13}{$z_{5}+z_{1}+z_{1}+z_{0}+z_{5}+z_{3}=z_{1}$} \quad
\color{dc13}{$z_{5}+z_{1}+z_{1}+z_{5}+z_{0}+z_{3}=z_{1}$} \quad
\color{dc13}{$z_{5}+z_{1}+z_{1}+z_{5}+z_{3}+z_{0}=z_{1}$} \quad
\color{dc14}{$z_{0}+z_{2}+z_{1}+z_{2}+z_{6}+z_{2}=z_{6}$} \quad
\color{dc14}{$z_{2}+z_{0}+z_{1}+z_{2}+z_{6}+z_{2}=z_{6}$} \quad
\color{dc14}{$z_{2}+z_{1}+z_{0}+z_{2}+z_{6}+z_{2}=z_{6}$} \quad
\color{dc14}{$z_{2}+z_{1}+z_{2}+z_{0}+z_{6}+z_{2}=z_{6}$} \quad
\color{dc14}{$z_{2}+z_{1}+z_{2}+z_{6}+z_{0}+z_{2}=z_{6}$} \quad
\color{dc14}{$z_{2}+z_{1}+z_{2}+z_{6}+z_{2}+z_{0}=z_{6}$} \quad
\color{dc15}{$z_{0}+z_{6}+z_{1}+z_{2}+z_{5}+z_{4}=z_{4}$} \quad
\color{dc15}{$z_{6}+z_{0}+z_{1}+z_{2}+z_{5}+z_{4}=z_{4}$} \quad
\color{dc15}{$z_{6}+z_{1}+z_{0}+z_{2}+z_{5}+z_{4}=z_{4}$} \quad
\color{dc15}{$z_{6}+z_{1}+z_{2}+z_{0}+z_{5}+z_{4}=z_{4}$} \quad
\color{dc15}{$z_{6}+z_{1}+z_{2}+z_{5}+z_{0}+z_{4}=z_{4}$} \quad
\color{dc15}{$z_{6}+z_{1}+z_{2}+z_{5}+z_{4}+z_{0}=z_{4}$} \quad
\color{dc16}{$z_{0}+z_{1}+z_{4}+z_{6}+z_{1}+z_{3}=z_{1}$} \quad
\color{dc16}{$z_{1}+z_{0}+z_{4}+z_{6}+z_{1}+z_{3}=z_{1}$} \quad
\color{dc16}{$z_{1}+z_{4}+z_{0}+z_{6}+z_{1}+z_{3}=z_{1}$} \quad
\color{dc16}{$z_{1}+z_{4}+z_{6}+z_{0}+z_{1}+z_{3}=z_{1}$} \quad
\color{dc16}{$z_{1}+z_{4}+z_{6}+z_{1}+z_{0}+z_{3}=z_{1}$} \quad
\color{dc16}{$z_{1}+z_{4}+z_{6}+z_{1}+z_{3}+z_{0}=z_{1}$} \quad
\color{dc17}{$z_{0}+z_{4}+z_{1}+z_{2}+z_{4}+z_{6}=z_{3}$} \quad
\color{dc17}{$z_{4}+z_{0}+z_{1}+z_{2}+z_{4}+z_{6}=z_{3}$} \quad
\color{dc17}{$z_{4}+z_{1}+z_{0}+z_{2}+z_{4}+z_{6}=z_{3}$} \quad
\color{dc17}{$z_{4}+z_{1}+z_{2}+z_{0}+z_{4}+z_{6}=z_{3}$} \quad
\color{dc17}{$z_{4}+z_{1}+z_{2}+z_{4}+z_{0}+z_{6}=z_{3}$} \quad
\color{dc17}{$z_{4}+z_{1}+z_{2}+z_{4}+z_{6}+z_{0}=z_{3}$} \quad
\color{dc18}{$z_{0}+z_{4}+z_{4}+z_{1}+z_{2}+z_{2}=z_{6}$} \quad
\color{dc18}{$z_{4}+z_{0}+z_{4}+z_{1}+z_{2}+z_{2}=z_{6}$} \quad
\color{dc18}{$z_{4}+z_{4}+z_{0}+z_{1}+z_{2}+z_{2}=z_{6}$} \quad
\color{dc18}{$z_{4}+z_{4}+z_{1}+z_{0}+z_{2}+z_{2}=z_{6}$} \quad
\color{dc18}{$z_{4}+z_{4}+z_{1}+z_{2}+z_{0}+z_{2}=z_{6}$} \quad
\color{dc18}{$z_{4}+z_{4}+z_{1}+z_{2}+z_{2}+z_{0}=z_{6}$} \quad
\color{dc19}{$z_{0}+z_{2}+z_{1}+z_{3}+z_{5}+z_{4}=z_{1}$} \quad
\color{dc19}{$z_{2}+z_{0}+z_{1}+z_{3}+z_{5}+z_{4}=z_{1}$} \quad
}

\paragraph{\color{black}{Training, for Operator $\oplus$'s Commutativity}}\mbox{}\\
{\footnotesize \color{dc7}{$z_{2}\oplus z_{2}\oplus z_{4}\oplus z_{3}\oplus z_{6}\oplus z_{4}=r_{4}$} \quad
\color{dc7}{$z_{4}\oplus z_{3}\oplus z_{6}\oplus z_{4}\oplus z_{2}\oplus z_{2}=r_{4}$} \quad
\color{dc7}{$z_{3}\oplus z_{6}\oplus z_{4}\oplus z_{4}\oplus z_{2}\oplus z_{2}=r_{4}$} \quad
\color{dc7}{$z_{6}\oplus z_{2}\oplus z_{3}\oplus z_{2}\oplus z_{4}\oplus z_{4}=r_{4}$} \quad
\color{dc7}{$z_{2}\oplus z_{3}\oplus z_{4}\oplus z_{2}\oplus z_{6}\oplus z_{4}=r_{4}$} \quad
\color{dc7}{$z_{2}\oplus z_{6}\oplus z_{4}\oplus z_{4}\oplus z_{3}\oplus z_{2}=r_{4}$} \quad
\color{dc7}{$z_{6}\oplus z_{2}\oplus z_{4}\oplus z_{2}\oplus z_{4}\oplus z_{3}=r_{4}$} \quad
\color{dc7}{$z_{4}\oplus z_{2}\oplus z_{6}\oplus z_{2}\oplus z_{3}\oplus z_{4}=r_{4}$} \quad
\color{dc7}{$z_{6}\oplus z_{2}\oplus z_{2}\oplus z_{3}\oplus z_{4}\oplus z_{4}=r_{4}$} \quad
\color{dc7}{$z_{6}\oplus z_{4}\oplus z_{2}\oplus z_{2}\oplus z_{3}\oplus z_{4}=r_{4}$} \quad
\color{dc8}{$z_{6}\oplus z_{6}\oplus z_{4}\oplus z_{4}\oplus z_{2}\oplus z_{3}=r_{5}$} \quad
\color{dc8}{$z_{6}\oplus z_{4}\oplus z_{2}\oplus z_{6}\oplus z_{4}\oplus z_{3}=r_{5}$} \quad
\color{dc8}{$z_{4}\oplus z_{6}\oplus z_{2}\oplus z_{3}\oplus z_{6}\oplus z_{4}=r_{5}$} \quad
\color{dc8}{$z_{4}\oplus z_{2}\oplus z_{6}\oplus z_{4}\oplus z_{6}\oplus z_{3}=r_{5}$} \quad
\color{dc8}{$z_{6}\oplus z_{4}\oplus z_{6}\oplus z_{3}\oplus z_{2}\oplus z_{4}=r_{5}$} \quad
\color{dc8}{$z_{4}\oplus z_{4}\oplus z_{6}\oplus z_{2}\oplus z_{3}\oplus z_{6}=r_{5}$} \quad
\color{dc8}{$z_{6}\oplus z_{3}\oplus z_{4}\oplus z_{6}\oplus z_{4}\oplus z_{2}=r_{5}$} \quad
\color{dc8}{$z_{3}\oplus z_{6}\oplus z_{2}\oplus z_{4}\oplus z_{4}\oplus z_{6}=r_{5}$} \quad
\color{dc8}{$z_{6}\oplus z_{3}\oplus z_{4}\oplus z_{4}\oplus z_{6}\oplus z_{2}=r_{5}$} \quad
\color{dc8}{$z_{6}\oplus z_{2}\oplus z_{4}\oplus z_{4}\oplus z_{3}\oplus z_{6}=r_{5}$} \quad
\color{dc9}{$z_{5}\oplus z_{5}\oplus z_{6}\oplus z_{2}\oplus z_{3}\oplus z_{3}=r_{0}$} \quad
\color{dc9}{$z_{5}\oplus z_{2}\oplus z_{3}\oplus z_{5}\oplus z_{6}\oplus z_{3}=r_{0}$} \quad
\color{dc9}{$z_{6}\oplus z_{2}\oplus z_{3}\oplus z_{5}\oplus z_{3}\oplus z_{5}=r_{0}$} \quad
\color{dc9}{$z_{5}\oplus z_{2}\oplus z_{3}\oplus z_{3}\oplus z_{5}\oplus z_{6}=r_{0}$} \quad
\color{dc9}{$z_{3}\oplus z_{5}\oplus z_{5}\oplus z_{2}\oplus z_{3}\oplus z_{6}=r_{0}$} \quad
\color{dc9}{$z_{5}\oplus z_{6}\oplus z_{3}\oplus z_{2}\oplus z_{3}\oplus z_{5}=r_{0}$} \quad
\color{dc9}{$z_{2}\oplus z_{5}\oplus z_{3}\oplus z_{3}\oplus z_{6}\oplus z_{5}=r_{0}$} \quad
\color{dc9}{$z_{6}\oplus z_{2}\oplus z_{3}\oplus z_{3}\oplus z_{5}\oplus z_{5}=r_{0}$} \quad
\color{dc9}{$z_{3}\oplus z_{5}\oplus z_{3}\oplus z_{6}\oplus z_{5}\oplus z_{2}=r_{0}$} \quad
\color{dc9}{$z_{6}\oplus z_{3}\oplus z_{2}\oplus z_{3}\oplus z_{5}\oplus z_{5}=r_{0}$} \quad
\color{dc10}{$z_{5}\oplus z_{5}\oplus z_{5}\oplus z_{3}\oplus z_{5}\oplus z_{2}=r_{3}$} \quad
\color{dc10}{$z_{5}\oplus z_{5}\oplus z_{2}\oplus z_{5}\oplus z_{3}\oplus z_{5}=r_{3}$} \quad
\color{dc10}{$z_{3}\oplus z_{5}\oplus z_{5}\oplus z_{2}\oplus z_{5}\oplus z_{5}=r_{3}$} \quad
\color{dc10}{$z_{5}\oplus z_{2}\oplus z_{5}\oplus z_{5}\oplus z_{3}\oplus z_{5}=r_{3}$} \quad
\color{dc10}{$z_{5}\oplus z_{3}\oplus z_{2}\oplus z_{5}\oplus z_{5}\oplus z_{5}=r_{3}$} \quad
\color{dc10}{$z_{3}\oplus z_{5}\oplus z_{5}\oplus z_{5}\oplus z_{2}\oplus z_{5}=r_{3}$} \quad
\color{dc10}{$z_{5}\oplus z_{5}\oplus z_{3}\oplus z_{5}\oplus z_{5}\oplus z_{2}=r_{3}$} \quad
\color{dc10}{$z_{5}\oplus z_{5}\oplus z_{5}\oplus z_{3}\oplus z_{2}\oplus z_{5}=r_{3}$} \quad
\color{dc10}{$z_{5}\oplus z_{2}\oplus z_{3}\oplus z_{5}\oplus z_{5}\oplus z_{5}=r_{3}$} \quad
\color{dc10}{$z_{2}\oplus z_{5}\oplus z_{3}\oplus z_{5}\oplus z_{5}\oplus z_{5}=r_{3}$} \quad
\color{dc11}{$z_{6}\oplus z_{6}\oplus z_{2}\oplus z_{5}\oplus z_{5}\oplus z_{3}=r_{3}$} \quad
\color{dc11}{$z_{6}\oplus z_{6}\oplus z_{5}\oplus z_{2}\oplus z_{5}\oplus z_{3}=r_{3}$} \quad
\color{dc11}{$z_{5}\oplus z_{6}\oplus z_{3}\oplus z_{2}\oplus z_{5}\oplus z_{6}=r_{3}$} \quad
\color{dc11}{$z_{5}\oplus z_{3}\oplus z_{2}\oplus z_{6}\oplus z_{6}\oplus z_{5}=r_{3}$} \quad
\color{dc1}{$z_{6}\oplus z_{5}\oplus z_{4}\oplus z_{1}\oplus z_{2}\oplus z_{4}=r_{3}$} \quad
\color{dc2}{$z_{6}\oplus z_{2}\oplus z_{3}\oplus z_{5}\oplus z_{6}\oplus z_{1}=r_{1}$} \quad
\color{dc3}{$z_{4}\oplus z_{6}\oplus z_{2}\oplus z_{2}\oplus z_{4}\oplus z_{5}=r_{3}$} \quad
\color{dc4}{$z_{2}\oplus z_{3}\oplus z_{6}\oplus z_{1}\oplus z_{5}\oplus z_{5}=r_{5}$} \quad
\color{dc5}{$z_{5}\oplus z_{4}\oplus z_{5}\oplus z_{5}\oplus z_{4}\oplus z_{2}=r_{2}$} \quad
\color{dc6}{$z_{4}\oplus z_{1}\oplus z_{1}\oplus z_{4}\oplus z_{5}\oplus z_{1}=r_{4}$} \quad
}

\paragraph{\color{black}{Testing, for Operator $\oplus$'s Commutativity}}\mbox{}\\
{\footnotesize \color{dc1}{$z_{4}\oplus z_{4}\oplus z_{5}\oplus z_{6}\oplus z_{2}\oplus z_{1}=r_{3}$} \quad
\color{dc1}{$z_{4}\oplus z_{2}\oplus z_{6}\oplus z_{5}\oplus z_{4}\oplus z_{1}=r_{3}$} \quad
\color{dc1}{$z_{4}\oplus z_{4}\oplus z_{2}\oplus z_{1}\oplus z_{6}\oplus z_{5}=r_{3}$} \quad
\color{dc1}{$z_{6}\oplus z_{2}\oplus z_{1}\oplus z_{5}\oplus z_{4}\oplus z_{4}=r_{3}$} \quad
\color{dc1}{$z_{6}\oplus z_{4}\oplus z_{1}\oplus z_{4}\oplus z_{2}\oplus z_{5}=r_{3}$} \quad
\color{dc1}{$z_{5}\oplus z_{2}\oplus z_{4}\oplus z_{4}\oplus z_{6}\oplus z_{1}=r_{3}$} \quad
\color{dc1}{$z_{4}\oplus z_{5}\oplus z_{4}\oplus z_{6}\oplus z_{1}\oplus z_{2}=r_{3}$} \quad
\color{dc1}{$z_{6}\oplus z_{5}\oplus z_{1}\oplus z_{4}\oplus z_{2}\oplus z_{4}=r_{3}$} \quad
\color{dc1}{$z_{1}\oplus z_{2}\oplus z_{4}\oplus z_{6}\oplus z_{5}\oplus z_{4}=r_{3}$} \quad
\color{dc2}{$z_{6}\oplus z_{1}\oplus z_{6}\oplus z_{2}\oplus z_{3}\oplus z_{5}=r_{1}$} \quad
\color{dc2}{$z_{2}\oplus z_{5}\oplus z_{3}\oplus z_{1}\oplus z_{6}\oplus z_{6}=r_{1}$} \quad
\color{dc2}{$z_{3}\oplus z_{2}\oplus z_{6}\oplus z_{1}\oplus z_{5}\oplus z_{6}=r_{1}$} \quad
\color{dc2}{$z_{6}\oplus z_{2}\oplus z_{5}\oplus z_{1}\oplus z_{3}\oplus z_{6}=r_{1}$} \quad
\color{dc2}{$z_{6}\oplus z_{6}\oplus z_{3}\oplus z_{5}\oplus z_{2}\oplus z_{1}=r_{1}$} \quad
\color{dc2}{$z_{1}\oplus z_{6}\oplus z_{5}\oplus z_{6}\oplus z_{3}\oplus z_{2}=r_{1}$} \quad
\color{dc2}{$z_{5}\oplus z_{1}\oplus z_{6}\oplus z_{3}\oplus z_{6}\oplus z_{2}=r_{1}$} \quad
\color{dc2}{$z_{5}\oplus z_{6}\oplus z_{6}\oplus z_{3}\oplus z_{1}\oplus z_{2}=r_{1}$} \quad
\color{dc2}{$z_{5}\oplus z_{3}\oplus z_{6}\oplus z_{2}\oplus z_{6}\oplus z_{1}=r_{1}$} \quad
\color{dc3}{$z_{2}\oplus z_{4}\oplus z_{5}\oplus z_{2}\oplus z_{4}\oplus z_{6}=r_{3}$} \quad
\color{dc3}{$z_{4}\oplus z_{5}\oplus z_{4}\oplus z_{2}\oplus z_{2}\oplus z_{6}=r_{3}$} \quad
\color{dc3}{$z_{2}\oplus z_{5}\oplus z_{2}\oplus z_{4}\oplus z_{4}\oplus z_{6}=r_{3}$} \quad
\color{dc3}{$z_{5}\oplus z_{2}\oplus z_{6}\oplus z_{4}\oplus z_{2}\oplus z_{4}=r_{3}$} \quad
\color{dc3}{$z_{5}\oplus z_{4}\oplus z_{6}\oplus z_{2}\oplus z_{4}\oplus z_{2}=r_{3}$} \quad
\color{dc3}{$z_{2}\oplus z_{4}\oplus z_{2}\oplus z_{6}\oplus z_{4}\oplus z_{5}=r_{3}$} \quad
\color{dc3}{$z_{4}\oplus z_{2}\oplus z_{4}\oplus z_{5}\oplus z_{6}\oplus z_{2}=r_{3}$} \quad
\color{dc3}{$z_{2}\oplus z_{4}\oplus z_{5}\oplus z_{6}\oplus z_{4}\oplus z_{2}=r_{3}$} \quad
\color{dc3}{$z_{4}\oplus z_{5}\oplus z_{2}\oplus z_{2}\oplus z_{4}\oplus z_{6}=r_{3}$} \quad
\color{dc4}{$z_{2}\oplus z_{1}\oplus z_{6}\oplus z_{5}\oplus z_{5}\oplus z_{3}=r_{5}$} \quad
\color{dc4}{$z_{6}\oplus z_{5}\oplus z_{3}\oplus z_{1}\oplus z_{5}\oplus z_{2}=r_{5}$} \quad
\color{dc4}{$z_{6}\oplus z_{2}\oplus z_{5}\oplus z_{5}\oplus z_{1}\oplus z_{3}=r_{5}$} \quad
\color{dc4}{$z_{3}\oplus z_{6}\oplus z_{1}\oplus z_{5}\oplus z_{2}\oplus z_{5}=r_{5}$} \quad
\color{dc4}{$z_{5}\oplus z_{2}\oplus z_{5}\oplus z_{1}\oplus z_{6}\oplus z_{3}=r_{5}$} \quad
\color{dc4}{$z_{3}\oplus z_{1}\oplus z_{6}\oplus z_{2}\oplus z_{5}\oplus z_{5}=r_{5}$} \quad
\color{dc4}{$z_{3}\oplus z_{6}\oplus z_{5}\oplus z_{2}\oplus z_{5}\oplus z_{1}=r_{5}$} \quad
\color{dc4}{$z_{6}\oplus z_{5}\oplus z_{1}\oplus z_{3}\oplus z_{2}\oplus z_{5}=r_{5}$} \quad
\color{dc4}{$z_{1}\oplus z_{5}\oplus z_{5}\oplus z_{2}\oplus z_{6}\oplus z_{3}=r_{5}$} \quad
\color{dc5}{$z_{2}\oplus z_{5}\oplus z_{5}\oplus z_{5}\oplus z_{4}\oplus z_{4}=r_{2}$} \quad
\color{dc5}{$z_{4}\oplus z_{5}\oplus z_{5}\oplus z_{2}\oplus z_{5}\oplus z_{4}=r_{2}$} \quad
\color{dc5}{$z_{2}\oplus z_{5}\oplus z_{5}\oplus z_{4}\oplus z_{5}\oplus z_{4}=r_{2}$} \quad
\color{dc5}{$z_{2}\oplus z_{5}\oplus z_{4}\oplus z_{5}\oplus z_{5}\oplus z_{4}=r_{2}$} \quad
\color{dc5}{$z_{5}\oplus z_{4}\oplus z_{5}\oplus z_{5}\oplus z_{2}\oplus z_{4}=r_{2}$} \quad
\color{dc5}{$z_{2}\oplus z_{4}\oplus z_{5}\oplus z_{4}\oplus z_{5}\oplus z_{5}=r_{2}$} \quad
\color{dc5}{$z_{5}\oplus z_{4}\oplus z_{2}\oplus z_{5}\oplus z_{5}\oplus z_{4}=r_{2}$} \quad
\color{dc5}{$z_{5}\oplus z_{4}\oplus z_{5}\oplus z_{2}\oplus z_{5}\oplus z_{4}=r_{2}$} \quad
\color{dc5}{$z_{5}\oplus z_{2}\oplus z_{4}\oplus z_{5}\oplus z_{5}\oplus z_{4}=r_{2}$} \quad
\color{dc6}{$z_{4}\oplus z_{1}\oplus z_{4}\oplus z_{5}\oplus z_{1}\oplus z_{1}=r_{4}$} \quad
\color{dc6}{$z_{1}\oplus z_{5}\oplus z_{4}\oplus z_{1}\oplus z_{1}\oplus z_{4}=r_{4}$} \quad
\color{dc6}{$z_{1}\oplus z_{1}\oplus z_{4}\oplus z_{1}\oplus z_{5}\oplus z_{4}=r_{4}$} \quad
\color{dc6}{$z_{4}\oplus z_{1}\oplus z_{1}\oplus z_{5}\oplus z_{4}\oplus z_{1}=r_{4}$} \quad
\color{dc6}{$z_{1}\oplus z_{5}\oplus z_{1}\oplus z_{4}\oplus z_{1}\oplus z_{4}=r_{4}$} \quad

\paragraph{\color{black}{Training, for Operator $\oplus$'s Identity}}\mbox{}\\
{\footnotesize \color{dc0}{$z_{0}\oplus z_{4}\oplus z_{3}\oplus z_{5}\oplus z_{3}\oplus z_{1}=r_{6}$} \quad
\color{dc0}{$z_{4}\oplus z_{0}\oplus z_{3}\oplus z_{5}\oplus z_{3}\oplus z_{1}=r_{6}$} \quad
\color{dc0}{$z_{4}\oplus z_{3}\oplus z_{0}\oplus z_{5}\oplus z_{3}\oplus z_{1}=r_{6}$} \quad
\color{dc0}{$z_{4}\oplus z_{3}\oplus z_{5}\oplus z_{0}\oplus z_{3}\oplus z_{1}=r_{6}$} \quad
\color{dc0}{$z_{4}\oplus z_{3}\oplus z_{5}\oplus z_{3}\oplus z_{0}\oplus z_{1}=r_{6}$} \quad
\color{dc0}{$z_{4}\oplus z_{3}\oplus z_{5}\oplus z_{3}\oplus z_{1}\oplus z_{0}=r_{6}$} \quad
\color{dc0}{$z_{4}\oplus z_{3}\oplus z_{5}\oplus z_{3}\oplus z_{1}=r_{6}\quad\quad$} \quad
\color{dc1}{$z_{0}\oplus z_{1}\oplus z_{5}\oplus z_{6}\oplus z_{6}\oplus z_{1}=r_{0}$} \quad
\color{dc1}{$z_{1}\oplus z_{0}\oplus z_{5}\oplus z_{6}\oplus z_{6}\oplus z_{1}=r_{0}$} \quad
\color{dc1}{$z_{1}\oplus z_{5}\oplus z_{0}\oplus z_{6}\oplus z_{6}\oplus z_{1}=r_{0}$} \quad
\color{dc1}{$z_{1}\oplus z_{5}\oplus z_{6}\oplus z_{0}\oplus z_{6}\oplus z_{1}=r_{0}$} \quad
\color{dc1}{$z_{1}\oplus z_{5}\oplus z_{6}\oplus z_{6}\oplus z_{0}\oplus z_{1}=r_{0}$} \quad
\color{dc1}{$z_{1}\oplus z_{5}\oplus z_{6}\oplus z_{6}\oplus z_{1}\oplus z_{0}=r_{0}$} \quad
\color{dc1}{$z_{1}\oplus z_{5}\oplus z_{6}\oplus z_{6}\oplus z_{1}=r_{0}\quad\quad$} \quad
\color{dc2}{$z_{0}\oplus z_{5}\oplus z_{2}\oplus z_{5}\oplus z_{2}\oplus z_{3}=r_{0}$} \quad
\color{dc2}{$z_{5}\oplus z_{0}\oplus z_{2}\oplus z_{5}\oplus z_{2}\oplus z_{3}=r_{0}$} \quad
\color{dc2}{$z_{5}\oplus z_{2}\oplus z_{0}\oplus z_{5}\oplus z_{2}\oplus z_{3}=r_{0}$} \quad
\color{dc2}{$z_{5}\oplus z_{2}\oplus z_{5}\oplus z_{0}\oplus z_{2}\oplus z_{3}=r_{0}$} \quad
\color{dc2}{$z_{5}\oplus z_{2}\oplus z_{5}\oplus z_{2}\oplus z_{0}\oplus z_{3}=r_{0}$} \quad
\color{dc2}{$z_{5}\oplus z_{2}\oplus z_{5}\oplus z_{2}\oplus z_{3}\oplus z_{0}=r_{0}$} \quad
\color{dc2}{$z_{5}\oplus z_{2}\oplus z_{5}\oplus z_{2}\oplus z_{3}=r_{0}\quad\quad$} \quad
\color{dc3}{$z_{0}\oplus z_{3}\oplus z_{1}\oplus z_{2}\oplus z_{2}\oplus z_{2}=r_{4}$} \quad
\color{dc3}{$z_{3}\oplus z_{0}\oplus z_{1}\oplus z_{2}\oplus z_{2}\oplus z_{2}=r_{4}$} \quad
\color{dc3}{$z_{3}\oplus z_{1}\oplus z_{0}\oplus z_{2}\oplus z_{2}\oplus z_{2}=r_{4}$} \quad
\color{dc3}{$z_{3}\oplus z_{1}\oplus z_{2}\oplus z_{0}\oplus z_{2}\oplus z_{2}=r_{4}$} \quad
\color{dc3}{$z_{3}\oplus z_{1}\oplus z_{2}\oplus z_{2}\oplus z_{0}\oplus z_{2}=r_{4}$} \quad
\color{dc3}{$z_{3}\oplus z_{1}\oplus z_{2}\oplus z_{2}\oplus z_{2}\oplus z_{0}=r_{4}$} \quad
\color{dc3}{$z_{3}\oplus z_{1}\oplus z_{2}\oplus z_{2}\oplus z_{2}=r_{4}\quad\quad$} \quad
\color{dc4}{$z_{0}\oplus z_{2}\oplus z_{4}\oplus z_{4}\oplus z_{3}\oplus z_{4}=r_{2}$} \quad
\color{dc4}{$z_{2}\oplus z_{0}\oplus z_{4}\oplus z_{4}\oplus z_{3}\oplus z_{4}=r_{2}$} \quad
\color{dc4}{$z_{2}\oplus z_{4}\oplus z_{0}\oplus z_{4}\oplus z_{3}\oplus z_{4}=r_{2}$} \quad
\color{dc4}{$z_{2}\oplus z_{4}\oplus z_{4}\oplus z_{0}\oplus z_{3}\oplus z_{4}=r_{2}$} \quad
\color{dc4}{$z_{2}\oplus z_{4}\oplus z_{4}\oplus z_{3}\oplus z_{0}\oplus z_{4}=r_{2}$} \quad
\color{dc4}{$z_{2}\oplus z_{4}\oplus z_{4}\oplus z_{3}\oplus z_{4}\oplus z_{0}=r_{2}$} \quad
\color{dc4}{$z_{2}\oplus z_{4}\oplus z_{4}\oplus z_{3}\oplus z_{4}=r_{2}\quad\quad$} \quad
\color{dc5}{$z_{0}\oplus z_{1}\oplus z_{4}\oplus z_{6}\oplus z_{5}\oplus z_{2}=r_{1}$} \quad
\color{dc5}{$z_{1}\oplus z_{0}\oplus z_{4}\oplus z_{6}\oplus z_{5}\oplus z_{2}=r_{1}$} \quad
\color{dc5}{$z_{1}\oplus z_{4}\oplus z_{0}\oplus z_{6}\oplus z_{5}\oplus z_{2}=r_{1}$} \quad
\color{dc5}{$z_{1}\oplus z_{4}\oplus z_{6}\oplus z_{0}\oplus z_{5}\oplus z_{2}=r_{1}$} \quad
\color{dc5}{$z_{1}\oplus z_{4}\oplus z_{6}\oplus z_{5}\oplus z_{0}\oplus z_{2}=r_{1}$} \quad
\color{dc5}{$z_{1}\oplus z_{4}\oplus z_{6}\oplus z_{5}\oplus z_{2}=r_{1}\quad\quad$} \quad
\color{dc11}{$z_{5}\oplus z_{6}\oplus z_{1}\oplus z_{4}\oplus z_{4}=r_{6}\quad\quad$} \quad
\color{dc12}{$z_{4}\oplus z_{2}\oplus z_{4}\oplus z_{6}\oplus z_{3}=r_{6}\quad\quad$} \quad
\color{dc13}{$z_{5}\oplus z_{1}\oplus z_{1}\oplus z_{5}\oplus z_{3}=r_{0}\quad\quad$} \quad
\color{dc14}{$z_{2}\oplus z_{1}\oplus z_{2}\oplus z_{6}\oplus z_{2}=r_{1}\quad\quad$} \quad
\color{dc15}{$z_{6}\oplus z_{1}\oplus z_{2}\oplus z_{5}\oplus z_{4}=r_{1}\quad\quad$} \quad
\color{dc16}{$z_{1}\oplus z_{4}\oplus z_{6}\oplus z_{1}\oplus z_{3}=r_{5}\quad\quad$} \quad
\color{dc17}{$z_{4}\oplus z_{1}\oplus z_{2}\oplus z_{4}\oplus z_{6}=r_{1}\quad\quad$} \quad
\color{dc18}{$z_{4}\oplus z_{4}\oplus z_{1}\oplus z_{2}\oplus z_{2}=r_{5}\quad\quad$} \quad
\color{dc19}{$z_{2}\oplus z_{1}\oplus z_{3}\oplus z_{5}\oplus z_{4}=r_{0}\quad\quad$} \quad
}

\paragraph{\color{black}{Testing, for Operator $\oplus$'s Identity}}\mbox{}\\
{\footnotesize \color{dc11}{$z_{0}\oplus z_{5}\oplus z_{6}\oplus z_{1}\oplus z_{4}\oplus z_{4}=r_{6}$} \quad
\color{dc11}{$z_{5}\oplus z_{0}\oplus z_{6}\oplus z_{1}\oplus z_{4}\oplus z_{4}=r_{6}$} \quad
\color{dc11}{$z_{5}\oplus z_{6}\oplus z_{0}\oplus z_{1}\oplus z_{4}\oplus z_{4}=r_{6}$} \quad
\color{dc11}{$z_{5}\oplus z_{6}\oplus z_{1}\oplus z_{0}\oplus z_{4}\oplus z_{4}=r_{6}$} \quad
\color{dc11}{$z_{5}\oplus z_{6}\oplus z_{1}\oplus z_{4}\oplus z_{0}\oplus z_{4}=r_{6}$} \quad
\color{dc11}{$z_{5}\oplus z_{6}\oplus z_{1}\oplus z_{4}\oplus z_{4}\oplus z_{0}=r_{6}$} \quad
\color{dc12}{$z_{0}\oplus z_{4}\oplus z_{2}\oplus z_{4}\oplus z_{6}\oplus z_{3}=r_{6}$} \quad
\color{dc12}{$z_{4}\oplus z_{0}\oplus z_{2}\oplus z_{4}\oplus z_{6}\oplus z_{3}=r_{6}$} \quad
\color{dc12}{$z_{4}\oplus z_{2}\oplus z_{0}\oplus z_{4}\oplus z_{6}\oplus z_{3}=r_{6}$} \quad
\color{dc12}{$z_{4}\oplus z_{2}\oplus z_{4}\oplus z_{0}\oplus z_{6}\oplus z_{3}=r_{6}$} \quad
\color{dc12}{$z_{4}\oplus z_{2}\oplus z_{4}\oplus z_{6}\oplus z_{0}\oplus z_{3}=r_{6}$} \quad
\color{dc12}{$z_{4}\oplus z_{2}\oplus z_{4}\oplus z_{6}\oplus z_{3}\oplus z_{0}=r_{6}$} \quad
\color{dc13}{$z_{0}\oplus z_{5}\oplus z_{1}\oplus z_{1}\oplus z_{5}\oplus z_{3}=r_{0}$} \quad
\color{dc13}{$z_{5}\oplus z_{0}\oplus z_{1}\oplus z_{1}\oplus z_{5}\oplus z_{3}=r_{0}$} \quad
\color{dc13}{$z_{5}\oplus z_{1}\oplus z_{0}\oplus z_{1}\oplus z_{5}\oplus z_{3}=r_{0}$} \quad
\color{dc13}{$z_{5}\oplus z_{1}\oplus z_{1}\oplus z_{0}\oplus z_{5}\oplus z_{3}=r_{0}$} \quad
\color{dc13}{$z_{5}\oplus z_{1}\oplus z_{1}\oplus z_{5}\oplus z_{0}\oplus z_{3}=r_{0}$} \quad
\color{dc13}{$z_{5}\oplus z_{1}\oplus z_{1}\oplus z_{5}\oplus z_{3}\oplus z_{0}=r_{0}$} \quad
\color{dc14}{$z_{0}\oplus z_{2}\oplus z_{1}\oplus z_{2}\oplus z_{6}\oplus z_{2}=r_{1}$} \quad
\color{dc14}{$z_{2}\oplus z_{0}\oplus z_{1}\oplus z_{2}\oplus z_{6}\oplus z_{2}=r_{1}$} \quad
\color{dc14}{$z_{2}\oplus z_{1}\oplus z_{0}\oplus z_{2}\oplus z_{6}\oplus z_{2}=r_{1}$} \quad
\color{dc14}{$z_{2}\oplus z_{1}\oplus z_{2}\oplus z_{0}\oplus z_{6}\oplus z_{2}=r_{1}$} \quad
\color{dc14}{$z_{2}\oplus z_{1}\oplus z_{2}\oplus z_{6}\oplus z_{0}\oplus z_{2}=r_{1}$} \quad
\color{dc14}{$z_{2}\oplus z_{1}\oplus z_{2}\oplus z_{6}\oplus z_{2}\oplus z_{0}=r_{1}$} \quad
\color{dc15}{$z_{0}\oplus z_{6}\oplus z_{1}\oplus z_{2}\oplus z_{5}\oplus z_{4}=r_{1}$} \quad
\color{dc15}{$z_{6}\oplus z_{0}\oplus z_{1}\oplus z_{2}\oplus z_{5}\oplus z_{4}=r_{1}$} \quad
\color{dc15}{$z_{6}\oplus z_{1}\oplus z_{0}\oplus z_{2}\oplus z_{5}\oplus z_{4}=r_{1}$} \quad
\color{dc15}{$z_{6}\oplus z_{1}\oplus z_{2}\oplus z_{0}\oplus z_{5}\oplus z_{4}=r_{1}$} \quad
\color{dc15}{$z_{6}\oplus z_{1}\oplus z_{2}\oplus z_{5}\oplus z_{0}\oplus z_{4}=r_{1}$} \quad
\color{dc15}{$z_{6}\oplus z_{1}\oplus z_{2}\oplus z_{5}\oplus z_{4}\oplus z_{0}=r_{1}$} \quad
\color{dc16}{$z_{0}\oplus z_{1}\oplus z_{4}\oplus z_{6}\oplus z_{1}\oplus z_{3}=r_{5}$} \quad
\color{dc16}{$z_{1}\oplus z_{0}\oplus z_{4}\oplus z_{6}\oplus z_{1}\oplus z_{3}=r_{5}$} \quad
\color{dc16}{$z_{1}\oplus z_{4}\oplus z_{0}\oplus z_{6}\oplus z_{1}\oplus z_{3}=r_{5}$} \quad
\color{dc16}{$z_{1}\oplus z_{4}\oplus z_{6}\oplus z_{0}\oplus z_{1}\oplus z_{3}=r_{5}$} \quad
\color{dc16}{$z_{1}\oplus z_{4}\oplus z_{6}\oplus z_{1}\oplus z_{0}\oplus z_{3}=r_{5}$} \quad
\color{dc16}{$z_{1}\oplus z_{4}\oplus z_{6}\oplus z_{1}\oplus z_{3}\oplus z_{0}=r_{5}$} \quad
\color{dc17}{$z_{0}\oplus z_{4}\oplus z_{1}\oplus z_{2}\oplus z_{4}\oplus z_{6}=r_{1}$} \quad
\color{dc17}{$z_{4}\oplus z_{0}\oplus z_{1}\oplus z_{2}\oplus z_{4}\oplus z_{6}=r_{1}$} \quad
\color{dc17}{$z_{4}\oplus z_{1}\oplus z_{0}\oplus z_{2}\oplus z_{4}\oplus z_{6}=r_{1}$} \quad
\color{dc17}{$z_{4}\oplus z_{1}\oplus z_{2}\oplus z_{0}\oplus z_{4}\oplus z_{6}=r_{1}$} \quad
\color{dc17}{$z_{4}\oplus z_{1}\oplus z_{2}\oplus z_{4}\oplus z_{0}\oplus z_{6}=r_{1}$} \quad
\color{dc17}{$z_{4}\oplus z_{1}\oplus z_{2}\oplus z_{4}\oplus z_{6}\oplus z_{0}=r_{1}$} \quad
\color{dc18}{$z_{0}\oplus z_{4}\oplus z_{4}\oplus z_{1}\oplus z_{2}\oplus z_{2}=r_{5}$} \quad
\color{dc18}{$z_{4}\oplus z_{0}\oplus z_{4}\oplus z_{1}\oplus z_{2}\oplus z_{2}=r_{5}$} \quad
\color{dc18}{$z_{4}\oplus z_{4}\oplus z_{0}\oplus z_{1}\oplus z_{2}\oplus z_{2}=r_{5}$} \quad
\color{dc18}{$z_{4}\oplus z_{4}\oplus z_{1}\oplus z_{0}\oplus z_{2}\oplus z_{2}=r_{5}$} \quad
\color{dc18}{$z_{4}\oplus z_{4}\oplus z_{1}\oplus z_{2}\oplus z_{0}\oplus z_{2}=r_{5}$} \quad
\color{dc18}{$z_{4}\oplus z_{4}\oplus z_{1}\oplus z_{2}\oplus z_{2}\oplus z_{0}=r_{5}$} \quad
\color{dc19}{$z_{0}\oplus z_{2}\oplus z_{1}\oplus z_{3}\oplus z_{5}\oplus z_{4}=r_{0}$} \quad
\color{dc19}{$z_{2}\oplus z_{0}\oplus z_{1}\oplus z_{3}\oplus z_{5}\oplus z_{4}=r_{0}$} \quad
}

\paragraph{\color{black}{Training, for Operator $\ominus$}}\mbox{}\\
{\footnotesize \color{dc7}{$z_{2}\ominus z_{2}\ominus z_{4}\ominus z_{3}\ominus z_{6}\ominus z_{4}=3$} \quad
\color{dc7}{$z_{4}\ominus z_{3}\ominus z_{6}\ominus z_{4}\ominus z_{2}\ominus z_{2}=4$} \quad
\color{dc7}{$z_{3}\ominus z_{6}\ominus z_{4}\ominus z_{4}\ominus z_{2}\ominus z_{2}=4$} \quad
\color{dc7}{$z_{6}\ominus z_{2}\ominus z_{3}\ominus z_{2}\ominus z_{4}\ominus z_{4}=3$} \quad
\color{dc7}{$z_{2}\ominus z_{3}\ominus z_{4}\ominus z_{2}\ominus z_{6}\ominus z_{4}=2$} \quad
\color{dc7}{$z_{2}\ominus z_{6}\ominus z_{4}\ominus z_{4}\ominus z_{3}\ominus z_{2}=4$} \quad
\color{dc7}{$z_{6}\ominus z_{2}\ominus z_{4}\ominus z_{2}\ominus z_{4}\ominus z_{3}=3$} \quad
\color{dc7}{$z_{4}\ominus z_{2}\ominus z_{6}\ominus z_{2}\ominus z_{3}\ominus z_{4}=2$} \quad
\color{dc7}{$z_{6}\ominus z_{2}\ominus z_{2}\ominus z_{3}\ominus z_{4}\ominus z_{4}=3$} \quad
\color{dc7}{$z_{6}\ominus z_{4}\ominus z_{2}\ominus z_{2}\ominus z_{3}\ominus z_{4}=3$} \quad
\color{dc8}{$z_{6}\ominus z_{6}\ominus z_{4}\ominus z_{4}\ominus z_{2}\ominus z_{3}=4$} \quad
\color{dc8}{$z_{6}\ominus z_{4}\ominus z_{2}\ominus z_{6}\ominus z_{4}\ominus z_{3}=4$} \quad
\color{dc8}{$z_{4}\ominus z_{6}\ominus z_{2}\ominus z_{3}\ominus z_{6}\ominus z_{4}=2$} \quad
\color{dc8}{$z_{4}\ominus z_{2}\ominus z_{6}\ominus z_{4}\ominus z_{6}\ominus z_{3}=3$} \quad
\color{dc8}{$z_{6}\ominus z_{4}\ominus z_{6}\ominus z_{3}\ominus z_{2}\ominus z_{4}=3$} \quad
\color{dc8}{$z_{4}\ominus z_{4}\ominus z_{6}\ominus z_{2}\ominus z_{3}\ominus z_{6}=2$} \quad
\color{dc8}{$z_{6}\ominus z_{3}\ominus z_{4}\ominus z_{6}\ominus z_{4}\ominus z_{2}=3$} \quad
\color{dc8}{$z_{3}\ominus z_{6}\ominus z_{2}\ominus z_{4}\ominus z_{4}\ominus z_{6}=2$} \quad
\color{dc8}{$z_{6}\ominus z_{3}\ominus z_{4}\ominus z_{4}\ominus z_{6}\ominus z_{2}=3$} \quad
\color{dc8}{$z_{6}\ominus z_{2}\ominus z_{4}\ominus z_{4}\ominus z_{3}\ominus z_{6}=3$} \quad
\color{dc9}{$z_{5}\ominus z_{5}\ominus z_{6}\ominus z_{2}\ominus z_{3}\ominus z_{3}=3$} \quad
\color{dc9}{$z_{5}\ominus z_{2}\ominus z_{3}\ominus z_{5}\ominus z_{6}\ominus z_{3}=2$} \quad
\color{dc9}{$z_{6}\ominus z_{2}\ominus z_{3}\ominus z_{5}\ominus z_{3}\ominus z_{5}=2$} \quad
\color{dc9}{$z_{5}\ominus z_{2}\ominus z_{3}\ominus z_{3}\ominus z_{5}\ominus z_{6}=2$} \quad
\color{dc9}{$z_{3}\ominus z_{5}\ominus z_{5}\ominus z_{2}\ominus z_{3}\ominus z_{6}=2$} \quad
\color{dc9}{$z_{5}\ominus z_{6}\ominus z_{3}\ominus z_{2}\ominus z_{3}\ominus z_{5}=2$} \quad
\color{dc9}{$z_{2}\ominus z_{5}\ominus z_{3}\ominus z_{3}\ominus z_{6}\ominus z_{5}=3$} \quad
\color{dc9}{$z_{6}\ominus z_{2}\ominus z_{3}\ominus z_{3}\ominus z_{5}\ominus z_{5}=3$} \quad
\color{dc9}{$z_{3}\ominus z_{5}\ominus z_{3}\ominus z_{6}\ominus z_{5}\ominus z_{2}=3$} \quad
\color{dc9}{$z_{6}\ominus z_{3}\ominus z_{2}\ominus z_{3}\ominus z_{5}\ominus z_{5}=3$} \quad
\color{dc10}{$z_{5}\ominus z_{5}\ominus z_{5}\ominus z_{3}\ominus z_{5}\ominus z_{2}=4$} \quad
\color{dc10}{$z_{5}\ominus z_{5}\ominus z_{2}\ominus z_{5}\ominus z_{3}\ominus z_{5}=3$} \quad
\color{dc10}{$z_{3}\ominus z_{5}\ominus z_{5}\ominus z_{2}\ominus z_{5}\ominus z_{5}=3$} \quad
\color{dc10}{$z_{5}\ominus z_{2}\ominus z_{5}\ominus z_{5}\ominus z_{3}\ominus z_{5}=3$} \quad
\color{dc10}{$z_{5}\ominus z_{3}\ominus z_{2}\ominus z_{5}\ominus z_{5}\ominus z_{5}=4$} \quad
\color{dc10}{$z_{3}\ominus z_{5}\ominus z_{5}\ominus z_{5}\ominus z_{2}\ominus z_{5}=3$} \quad
\color{dc10}{$z_{5}\ominus z_{5}\ominus z_{3}\ominus z_{5}\ominus z_{5}\ominus z_{2}=4$} \quad
\color{dc10}{$z_{5}\ominus z_{5}\ominus z_{5}\ominus z_{3}\ominus z_{2}\ominus z_{5}=4$} \quad
\color{dc10}{$z_{5}\ominus z_{2}\ominus z_{3}\ominus z_{5}\ominus z_{5}\ominus z_{5}=3$} \quad
\color{dc10}{$z_{2}\ominus z_{5}\ominus z_{3}\ominus z_{5}\ominus z_{5}\ominus z_{5}=3$} \quad
\color{dc11}{$z_{6}\ominus z_{6}\ominus z_{2}\ominus z_{5}\ominus z_{5}\ominus z_{3}=4$} \quad
\color{dc11}{$z_{6}\ominus z_{6}\ominus z_{5}\ominus z_{2}\ominus z_{5}\ominus z_{3}=4$} \quad
\color{dc11}{$z_{5}\ominus z_{6}\ominus z_{3}\ominus z_{2}\ominus z_{5}\ominus z_{6}=2$} \quad
\color{dc11}{$z_{5}\ominus z_{3}\ominus z_{2}\ominus z_{6}\ominus z_{6}\ominus z_{5}=4$} \quad
\color{dc1}{$z_{6}\ominus z_{5}\ominus z_{4}\ominus z_{1}\ominus z_{2}\ominus z_{4}=3$} \quad
\color{dc2}{$z_{6}\ominus z_{2}\ominus z_{3}\ominus z_{5}\ominus z_{6}\ominus z_{1}=2$} \quad
\color{dc3}{$z_{4}\ominus z_{6}\ominus z_{2}\ominus z_{2}\ominus z_{4}\ominus z_{5}=2$} \quad
\color{dc4}{$z_{2}\ominus z_{3}\ominus z_{6}\ominus z_{1}\ominus z_{5}\ominus z_{5}=2$} \quad
\color{dc5}{$z_{5}\ominus z_{4}\ominus z_{5}\ominus z_{5}\ominus z_{4}\ominus z_{2}=4$} \quad
\color{dc6}{$z_{4}\ominus z_{1}\ominus z_{1}\ominus z_{4}\ominus z_{5}\ominus z_{1}=3$} \quad
\color{dc0}{$z_{0}\ominus z_{4}\ominus z_{3}\ominus z_{5}\ominus z_{3}\ominus z_{1}=3$} \quad
\color{dc0}{$z_{4}\ominus z_{0}\ominus z_{3}\ominus z_{5}\ominus z_{3}\ominus z_{1}=3$} \quad
\color{dc0}{$z_{4}\ominus z_{3}\ominus z_{0}\ominus z_{5}\ominus z_{3}\ominus z_{1}=4$} \quad
\color{dc0}{$z_{4}\ominus z_{3}\ominus z_{5}\ominus z_{0}\ominus z_{3}\ominus z_{1}=3$} \quad
\color{dc0}{$z_{4}\ominus z_{3}\ominus z_{5}\ominus z_{3}\ominus z_{0}\ominus z_{1}=3$} \quad
\color{dc0}{$z_{4}\ominus z_{3}\ominus z_{5}\ominus z_{3}\ominus z_{1}\ominus z_{0}=4$} \quad
\color{dc0}{$z_{4}\ominus z_{3}\ominus z_{5}\ominus z_{3}\ominus z_{1}=3~$} \quad
\color{dc1}{$z_{0}\ominus z_{1}\ominus z_{5}\ominus z_{6}\ominus z_{6}\ominus z_{1}=2$} \quad
\color{dc1}{$z_{1}\ominus z_{0}\ominus z_{5}\ominus z_{6}\ominus z_{6}\ominus z_{1}=3$} \quad
\color{dc1}{$z_{1}\ominus z_{5}\ominus z_{0}\ominus z_{6}\ominus z_{6}\ominus z_{1}=3$} \quad
\color{dc1}{$z_{1}\ominus z_{5}\ominus z_{6}\ominus z_{0}\ominus z_{6}\ominus z_{1}=2$} \quad
\color{dc1}{$z_{1}\ominus z_{5}\ominus z_{6}\ominus z_{6}\ominus z_{0}\ominus z_{1}=2$} \quad
\color{dc1}{$z_{1}\ominus z_{5}\ominus z_{6}\ominus z_{6}\ominus z_{1}\ominus z_{0}=3$} \quad
\color{dc1}{$z_{1}\ominus z_{5}\ominus z_{6}\ominus z_{6}\ominus z_{1}=2~$} \quad
\color{dc2}{$z_{0}\ominus z_{5}\ominus z_{2}\ominus z_{5}\ominus z_{2}\ominus z_{3}=2$} \quad
\color{dc2}{$z_{5}\ominus z_{0}\ominus z_{2}\ominus z_{5}\ominus z_{2}\ominus z_{3}=2$} \quad
\color{dc2}{$z_{5}\ominus z_{2}\ominus z_{0}\ominus z_{5}\ominus z_{2}\ominus z_{3}=3$} \quad
\color{dc2}{$z_{5}\ominus z_{2}\ominus z_{5}\ominus z_{0}\ominus z_{2}\ominus z_{3}=2$} \quad
\color{dc2}{$z_{5}\ominus z_{2}\ominus z_{5}\ominus z_{2}\ominus z_{0}\ominus z_{3}=3$} \quad
\color{dc2}{$z_{5}\ominus z_{2}\ominus z_{5}\ominus z_{2}\ominus z_{3}\ominus z_{0}=3$} \quad
\color{dc2}{$z_{5}\ominus z_{2}\ominus z_{5}\ominus z_{2}\ominus z_{3}=2~$} \quad
\color{dc3}{$z_{0}\ominus z_{3}\ominus z_{1}\ominus z_{2}\ominus z_{2}\ominus z_{2}=3$} \quad
\color{dc3}{$z_{3}\ominus z_{0}\ominus z_{1}\ominus z_{2}\ominus z_{2}\ominus z_{2}=3$} \quad
\color{dc3}{$z_{3}\ominus z_{1}\ominus z_{0}\ominus z_{2}\ominus z_{2}\ominus z_{2}=4$} \quad
\color{dc3}{$z_{3}\ominus z_{1}\ominus z_{2}\ominus z_{0}\ominus z_{2}\ominus z_{2}=3$} \quad
\color{dc3}{$z_{3}\ominus z_{1}\ominus z_{2}\ominus z_{2}\ominus z_{0}\ominus z_{2}=3$} \quad
\color{dc3}{$z_{3}\ominus z_{1}\ominus z_{2}\ominus z_{2}\ominus z_{2}\ominus z_{0}=4$} \quad
\color{dc3}{$z_{3}\ominus z_{1}\ominus z_{2}\ominus z_{2}\ominus z_{2}=3~$} \quad
\color{dc4}{$z_{0}\ominus z_{2}\ominus z_{4}\ominus z_{4}\ominus z_{3}\ominus z_{4}=2$} \quad
\color{dc4}{$z_{2}\ominus z_{0}\ominus z_{4}\ominus z_{4}\ominus z_{3}\ominus z_{4}=3$} \quad
\color{dc4}{$z_{2}\ominus z_{4}\ominus z_{0}\ominus z_{4}\ominus z_{3}\ominus z_{4}=2$} \quad
\color{dc4}{$z_{2}\ominus z_{4}\ominus z_{4}\ominus z_{0}\ominus z_{3}\ominus z_{4}=2$} \quad
\color{dc4}{$z_{2}\ominus z_{4}\ominus z_{4}\ominus z_{3}\ominus z_{0}\ominus z_{4}=3$} \quad
\color{dc4}{$z_{2}\ominus z_{4}\ominus z_{4}\ominus z_{3}\ominus z_{4}\ominus z_{0}=3$} \quad
\color{dc4}{$z_{2}\ominus z_{4}\ominus z_{4}\ominus z_{3}\ominus z_{4}=2~$} \quad
\color{dc5}{$z_{0}\ominus z_{1}\ominus z_{4}\ominus z_{6}\ominus z_{5}\ominus z_{2}=2$} \quad
\color{dc5}{$z_{1}\ominus z_{0}\ominus z_{4}\ominus z_{6}\ominus z_{5}\ominus z_{2}=3$} \quad
\color{dc5}{$z_{1}\ominus z_{4}\ominus z_{0}\ominus z_{6}\ominus z_{5}\ominus z_{2}=3$} \quad
\color{dc5}{$z_{1}\ominus z_{4}\ominus z_{6}\ominus z_{0}\ominus z_{5}\ominus z_{2}=2$} \quad
\color{dc5}{$z_{1}\ominus z_{4}\ominus z_{6}\ominus z_{5}\ominus z_{0}\ominus z_{2}=2$} \quad
\color{dc5}{$z_{1}\ominus z_{4}\ominus z_{6}\ominus z_{5}\ominus z_{2}=2~$} \quad
\color{dc11}{$z_{5}\ominus z_{6}\ominus z_{1}\ominus z_{4}\ominus z_{4}=2~$} \quad
\color{dc12}{$z_{4}\ominus z_{2}\ominus z_{4}\ominus z_{6}\ominus z_{3}=2~$} \quad
\color{dc13}{$z_{5}\ominus z_{1}\ominus z_{1}\ominus z_{5}\ominus z_{3}=3~$} \quad
\color{dc14}{$z_{2}\ominus z_{1}\ominus z_{2}\ominus z_{6}\ominus z_{2}=2~$} \quad
\color{dc15}{$z_{6}\ominus z_{1}\ominus z_{2}\ominus z_{5}\ominus z_{4}=2~$} \quad
\color{dc16}{$z_{1}\ominus z_{4}\ominus z_{6}\ominus z_{1}\ominus z_{3}=1~$} \quad
\color{dc17}{$z_{4}\ominus z_{1}\ominus z_{2}\ominus z_{4}\ominus z_{6}=1~$} \quad
\color{dc18}{$z_{4}\ominus z_{4}\ominus z_{1}\ominus z_{2}\ominus z_{2}=3~$} \quad
\color{dc19}{$z_{2}\ominus z_{1}\ominus z_{3}\ominus z_{5}\ominus z_{4}=2~$} \quad
}

\paragraph{\color{black}{Testing, for Operator $\ominus$}}\mbox{}\\
{\footnotesize \color{dc1}{$z_{4}\ominus z_{4}\ominus z_{5}\ominus z_{6}\ominus z_{2}\ominus z_{1}=3$} \quad
\color{dc1}{$z_{4}\ominus z_{2}\ominus z_{6}\ominus z_{5}\ominus z_{4}\ominus z_{1}=4$} \quad
\color{dc1}{$z_{4}\ominus z_{4}\ominus z_{2}\ominus z_{1}\ominus z_{6}\ominus z_{5}=4$} \quad
\color{dc1}{$z_{6}\ominus z_{2}\ominus z_{1}\ominus z_{5}\ominus z_{4}\ominus z_{4}=4$} \quad
\color{dc1}{$z_{6}\ominus z_{4}\ominus z_{1}\ominus z_{4}\ominus z_{2}\ominus z_{5}=3$} \quad
\color{dc1}{$z_{5}\ominus z_{2}\ominus z_{4}\ominus z_{4}\ominus z_{6}\ominus z_{1}=3$} \quad
\color{dc1}{$z_{4}\ominus z_{5}\ominus z_{4}\ominus z_{6}\ominus z_{1}\ominus z_{2}=2$} \quad
\color{dc1}{$z_{6}\ominus z_{5}\ominus z_{1}\ominus z_{4}\ominus z_{2}\ominus z_{4}=3$} \quad
\color{dc1}{$z_{1}\ominus z_{2}\ominus z_{4}\ominus z_{6}\ominus z_{5}\ominus z_{4}=2$} \quad
\color{dc2}{$z_{6}\ominus z_{1}\ominus z_{6}\ominus z_{2}\ominus z_{3}\ominus z_{5}=2$} \quad
\color{dc2}{$z_{2}\ominus z_{5}\ominus z_{3}\ominus z_{1}\ominus z_{6}\ominus z_{6}=3$} \quad
\color{dc2}{$z_{3}\ominus z_{2}\ominus z_{6}\ominus z_{1}\ominus z_{5}\ominus z_{6}=2$} \quad
\color{dc2}{$z_{6}\ominus z_{2}\ominus z_{5}\ominus z_{1}\ominus z_{3}\ominus z_{6}=2$} \quad
\color{dc2}{$z_{6}\ominus z_{6}\ominus z_{3}\ominus z_{5}\ominus z_{2}\ominus z_{1}=4$} \quad
\color{dc2}{$z_{1}\ominus z_{6}\ominus z_{5}\ominus z_{6}\ominus z_{3}\ominus z_{2}=3$} \quad
\color{dc2}{$z_{5}\ominus z_{1}\ominus z_{6}\ominus z_{3}\ominus z_{6}\ominus z_{2}=3$} \quad
\color{dc2}{$z_{5}\ominus z_{6}\ominus z_{6}\ominus z_{3}\ominus z_{1}\ominus z_{2}=3$} \quad
\color{dc2}{$z_{5}\ominus z_{3}\ominus z_{6}\ominus z_{2}\ominus z_{6}\ominus z_{1}=3$} \quad
\color{dc3}{$z_{2}\ominus z_{4}\ominus z_{5}\ominus z_{2}\ominus z_{4}\ominus z_{6}=1$} \quad
\color{dc3}{$z_{4}\ominus z_{5}\ominus z_{4}\ominus z_{2}\ominus z_{2}\ominus z_{6}=3$} \quad
\color{dc3}{$z_{2}\ominus z_{5}\ominus z_{2}\ominus z_{4}\ominus z_{4}\ominus z_{6}=2$} \quad
\color{dc3}{$z_{5}\ominus z_{2}\ominus z_{6}\ominus z_{4}\ominus z_{2}\ominus z_{4}=3$} \quad
\color{dc3}{$z_{5}\ominus z_{4}\ominus z_{6}\ominus z_{2}\ominus z_{4}\ominus z_{2}=3$} \quad
\color{dc3}{$z_{2}\ominus z_{4}\ominus z_{2}\ominus z_{6}\ominus z_{4}\ominus z_{5}=2$} \quad
\color{dc3}{$z_{4}\ominus z_{2}\ominus z_{4}\ominus z_{5}\ominus z_{6}\ominus z_{2}=2$} \quad
\color{dc3}{$z_{2}\ominus z_{4}\ominus z_{5}\ominus z_{6}\ominus z_{4}\ominus z_{2}=2$} \quad
\color{dc3}{$z_{4}\ominus z_{5}\ominus z_{2}\ominus z_{2}\ominus z_{4}\ominus z_{6}=2$} \quad
\color{dc4}{$z_{2}\ominus z_{1}\ominus z_{6}\ominus z_{5}\ominus z_{5}\ominus z_{3}=4$} \quad
\color{dc4}{$z_{6}\ominus z_{5}\ominus z_{3}\ominus z_{1}\ominus z_{5}\ominus z_{2}=4$} \quad
\color{dc4}{$z_{6}\ominus z_{2}\ominus z_{5}\ominus z_{5}\ominus z_{1}\ominus z_{3}=3$} \quad
\color{dc4}{$z_{3}\ominus z_{6}\ominus z_{1}\ominus z_{5}\ominus z_{2}\ominus z_{5}=2$} \quad
\color{dc4}{$z_{5}\ominus z_{2}\ominus z_{5}\ominus z_{1}\ominus z_{6}\ominus z_{3}=3$} \quad
\color{dc4}{$z_{3}\ominus z_{1}\ominus z_{6}\ominus z_{2}\ominus z_{5}\ominus z_{5}=3$} \quad
\color{dc4}{$z_{3}\ominus z_{6}\ominus z_{5}\ominus z_{2}\ominus z_{5}\ominus z_{1}=3$} \quad
\color{dc4}{$z_{6}\ominus z_{5}\ominus z_{1}\ominus z_{3}\ominus z_{2}\ominus z_{5}=3$} \quad
\color{dc4}{$z_{1}\ominus z_{5}\ominus z_{5}\ominus z_{2}\ominus z_{6}\ominus z_{3}=3$} \quad
\color{dc5}{$z_{2}\ominus z_{5}\ominus z_{5}\ominus z_{5}\ominus z_{4}\ominus z_{4}=4$} \quad
\color{dc5}{$z_{4}\ominus z_{5}\ominus z_{5}\ominus z_{2}\ominus z_{5}\ominus z_{4}=3$} \quad
\color{dc5}{$z_{2}\ominus z_{5}\ominus z_{5}\ominus z_{4}\ominus z_{5}\ominus z_{4}=3$} \quad
\color{dc5}{$z_{2}\ominus z_{5}\ominus z_{4}\ominus z_{5}\ominus z_{5}\ominus z_{4}=3$} \quad
\color{dc5}{$z_{5}\ominus z_{4}\ominus z_{5}\ominus z_{5}\ominus z_{2}\ominus z_{4}=3$} \quad
\color{dc5}{$z_{2}\ominus z_{4}\ominus z_{5}\ominus z_{4}\ominus z_{5}\ominus z_{5}=2$} \quad
\color{dc5}{$z_{5}\ominus z_{4}\ominus z_{2}\ominus z_{5}\ominus z_{5}\ominus z_{4}=4$} \quad
\color{dc5}{$z_{5}\ominus z_{4}\ominus z_{5}\ominus z_{2}\ominus z_{5}\ominus z_{4}=3$} \quad
\color{dc5}{$z_{5}\ominus z_{2}\ominus z_{4}\ominus z_{5}\ominus z_{5}\ominus z_{4}=3$} \quad
\color{dc6}{$z_{4}\ominus z_{1}\ominus z_{4}\ominus z_{5}\ominus z_{1}\ominus z_{1}=3$} \quad
\color{dc6}{$z_{1}\ominus z_{5}\ominus z_{4}\ominus z_{1}\ominus z_{1}\ominus z_{4}=3$} \quad
\color{dc6}{$z_{1}\ominus z_{1}\ominus z_{4}\ominus z_{1}\ominus z_{5}\ominus z_{4}=3$} \quad
\color{dc6}{$z_{4}\ominus z_{1}\ominus z_{1}\ominus z_{5}\ominus z_{4}\ominus z_{1}=4$} \quad
\color{dc6}{$z_{1}\ominus z_{5}\ominus z_{1}\ominus z_{4}\ominus z_{1}\ominus z_{4}=2$} \quad
\color{dc11}{$z_{0}\ominus z_{5}\ominus z_{6}\ominus z_{1}\ominus z_{4}\ominus z_{4}=2$} \quad
\color{dc11}{$z_{5}\ominus z_{0}\ominus z_{6}\ominus z_{1}\ominus z_{4}\ominus z_{4}=3$} \quad
\color{dc11}{$z_{5}\ominus z_{6}\ominus z_{0}\ominus z_{1}\ominus z_{4}\ominus z_{4}=2$} \quad
\color{dc11}{$z_{5}\ominus z_{6}\ominus z_{1}\ominus z_{0}\ominus z_{4}\ominus z_{4}=3$} \quad
\color{dc11}{$z_{5}\ominus z_{6}\ominus z_{1}\ominus z_{4}\ominus z_{0}\ominus z_{4}=2$} \quad
\color{dc11}{$z_{5}\ominus z_{6}\ominus z_{1}\ominus z_{4}\ominus z_{4}\ominus z_{0}=3$} \quad
\color{dc12}{$z_{0}\ominus z_{4}\ominus z_{2}\ominus z_{4}\ominus z_{6}\ominus z_{3}=2$} \quad
\color{dc12}{$z_{4}\ominus z_{0}\ominus z_{2}\ominus z_{4}\ominus z_{6}\ominus z_{3}=2$} \quad
\color{dc12}{$z_{4}\ominus z_{2}\ominus z_{0}\ominus z_{4}\ominus z_{6}\ominus z_{3}=3$} \quad
\color{dc12}{$z_{4}\ominus z_{2}\ominus z_{4}\ominus z_{0}\ominus z_{6}\ominus z_{3}=3$} \quad
\color{dc12}{$z_{4}\ominus z_{2}\ominus z_{4}\ominus z_{6}\ominus z_{0}\ominus z_{3}=2$} \quad
\color{dc12}{$z_{4}\ominus z_{2}\ominus z_{4}\ominus z_{6}\ominus z_{3}\ominus z_{0}=3$} \quad
\color{dc13}{$z_{0}\ominus z_{5}\ominus z_{1}\ominus z_{1}\ominus z_{5}\ominus z_{3}=3$} \quad
\color{dc13}{$z_{5}\ominus z_{0}\ominus z_{1}\ominus z_{1}\ominus z_{5}\ominus z_{3}=3$} \quad
\color{dc13}{$z_{5}\ominus z_{1}\ominus z_{0}\ominus z_{1}\ominus z_{5}\ominus z_{3}=3$} \quad
\color{dc13}{$z_{5}\ominus z_{1}\ominus z_{1}\ominus z_{0}\ominus z_{5}\ominus z_{3}=4$} \quad
\color{dc13}{$z_{5}\ominus z_{1}\ominus z_{1}\ominus z_{5}\ominus z_{0}\ominus z_{3}=3$} \quad
\color{dc13}{$z_{5}\ominus z_{1}\ominus z_{1}\ominus z_{5}\ominus z_{3}\ominus z_{0}=4$} \quad
\color{dc14}{$z_{0}\ominus z_{2}\ominus z_{1}\ominus z_{2}\ominus z_{6}\ominus z_{2}=2$} \quad
\color{dc14}{$z_{2}\ominus z_{0}\ominus z_{1}\ominus z_{2}\ominus z_{6}\ominus z_{2}=2$} \quad
\color{dc14}{$z_{2}\ominus z_{1}\ominus z_{0}\ominus z_{2}\ominus z_{6}\ominus z_{2}=3$} \quad
\color{dc14}{$z_{2}\ominus z_{1}\ominus z_{2}\ominus z_{0}\ominus z_{6}\ominus z_{2}=3$} \quad
\color{dc14}{$z_{2}\ominus z_{1}\ominus z_{2}\ominus z_{6}\ominus z_{0}\ominus z_{2}=2$} \quad
\color{dc14}{$z_{2}\ominus z_{1}\ominus z_{2}\ominus z_{6}\ominus z_{2}\ominus z_{0}=3$} \quad
\color{dc15}{$z_{0}\ominus z_{6}\ominus z_{1}\ominus z_{2}\ominus z_{5}\ominus z_{4}=2$} \quad
\color{dc15}{$z_{6}\ominus z_{0}\ominus z_{1}\ominus z_{2}\ominus z_{5}\ominus z_{4}=2$} \quad
\color{dc15}{$z_{6}\ominus z_{1}\ominus z_{0}\ominus z_{2}\ominus z_{5}\ominus z_{4}=3$} \quad
\color{dc15}{$z_{6}\ominus z_{1}\ominus z_{2}\ominus z_{0}\ominus z_{5}\ominus z_{4}=3$} \quad
\color{dc15}{$z_{6}\ominus z_{1}\ominus z_{2}\ominus z_{5}\ominus z_{0}\ominus z_{4}=2$} \quad
\color{dc15}{$z_{6}\ominus z_{1}\ominus z_{2}\ominus z_{5}\ominus z_{4}\ominus z_{0}=3$} \quad
\color{dc16}{$z_{0}\ominus z_{1}\ominus z_{4}\ominus z_{6}\ominus z_{1}\ominus z_{3}=1$} \quad
\color{dc16}{$z_{1}\ominus z_{0}\ominus z_{4}\ominus z_{6}\ominus z_{1}\ominus z_{3}=2$} \quad
\color{dc16}{$z_{1}\ominus z_{4}\ominus z_{0}\ominus z_{6}\ominus z_{1}\ominus z_{3}=2$} \quad
\color{dc16}{$z_{1}\ominus z_{4}\ominus z_{6}\ominus z_{0}\ominus z_{1}\ominus z_{3}=1$} \quad
\color{dc16}{$z_{1}\ominus z_{4}\ominus z_{6}\ominus z_{1}\ominus z_{0}\ominus z_{3}=2$} \quad
\color{dc16}{$z_{1}\ominus z_{4}\ominus z_{6}\ominus z_{1}\ominus z_{3}\ominus z_{0}=2$} \quad
\color{dc17}{$z_{0}\ominus z_{4}\ominus z_{1}\ominus z_{2}\ominus z_{4}\ominus z_{6}=1$} \quad
\color{dc17}{$z_{4}\ominus z_{0}\ominus z_{1}\ominus z_{2}\ominus z_{4}\ominus z_{6}=1$} \quad
\color{dc17}{$z_{4}\ominus z_{1}\ominus z_{0}\ominus z_{2}\ominus z_{4}\ominus z_{6}=2$} \quad
\color{dc17}{$z_{4}\ominus z_{1}\ominus z_{2}\ominus z_{0}\ominus z_{4}\ominus z_{6}=2$} \quad
\color{dc17}{$z_{4}\ominus z_{1}\ominus z_{2}\ominus z_{4}\ominus z_{0}\ominus z_{6}=2$} \quad
\color{dc17}{$z_{4}\ominus z_{1}\ominus z_{2}\ominus z_{4}\ominus z_{6}\ominus z_{0}=2$} \quad
\color{dc18}{$z_{0}\ominus z_{4}\ominus z_{4}\ominus z_{1}\ominus z_{2}\ominus z_{2}=3$} \quad
\color{dc18}{$z_{4}\ominus z_{0}\ominus z_{4}\ominus z_{1}\ominus z_{2}\ominus z_{2}=3$} \quad
\color{dc18}{$z_{4}\ominus z_{4}\ominus z_{0}\ominus z_{1}\ominus z_{2}\ominus z_{2}=3$} \quad
\color{dc18}{$z_{4}\ominus z_{4}\ominus z_{1}\ominus z_{0}\ominus z_{2}\ominus z_{2}=4$} \quad
\color{dc18}{$z_{4}\ominus z_{4}\ominus z_{1}\ominus z_{2}\ominus z_{0}\ominus z_{2}=3$} \quad
\color{dc18}{$z_{4}\ominus z_{4}\ominus z_{1}\ominus z_{2}\ominus z_{2}\ominus z_{0}=4$} \quad
\color{dc19}{$z_{0}\ominus z_{2}\ominus z_{1}\ominus z_{3}\ominus z_{5}\ominus z_{4}=2$} \quad
\color{dc19}{$z_{2}\ominus z_{0}\ominus z_{1}\ominus z_{3}\ominus z_{5}\ominus z_{4}=2$} \quad
}

\paragraph{\color{black}{Training, for Operator $\triangleleft$}}\mbox{}\\
{\footnotesize \color{dc7}{$z_{2}\triangleleft z_{2}\triangleleft z_{4}\triangleleft z_{3}\triangleleft z_{6}\triangleleft z_{4}=z_{2}$} \quad
\color{dc7}{$z_{4}\triangleleft z_{3}\triangleleft z_{6}\triangleleft z_{4}\triangleleft z_{2}\triangleleft z_{2}=z_{4}$} \quad
\color{dc7}{$z_{3}\triangleleft z_{6}\triangleleft z_{4}\triangleleft z_{4}\triangleleft z_{2}\triangleleft z_{2}=z_{3}$} \quad
\color{dc7}{$z_{6}\triangleleft z_{2}\triangleleft z_{3}\triangleleft z_{2}\triangleleft z_{4}\triangleleft z_{4}=z_{6}$} \quad
\color{dc7}{$z_{2}\triangleleft z_{3}\triangleleft z_{4}\triangleleft z_{2}\triangleleft z_{6}\triangleleft z_{4}=z_{2}$} \quad
\color{dc7}{$z_{2}\triangleleft z_{6}\triangleleft z_{4}\triangleleft z_{4}\triangleleft z_{3}\triangleleft z_{2}=z_{2}$} \quad
\color{dc7}{$z_{6}\triangleleft z_{2}\triangleleft z_{4}\triangleleft z_{2}\triangleleft z_{4}\triangleleft z_{3}=z_{6}$} \quad
\color{dc7}{$z_{4}\triangleleft z_{2}\triangleleft z_{6}\triangleleft z_{2}\triangleleft z_{3}\triangleleft z_{4}=z_{4}$} \quad
\color{dc7}{$z_{6}\triangleleft z_{2}\triangleleft z_{2}\triangleleft z_{3}\triangleleft z_{4}\triangleleft z_{4}=z_{6}$} \quad
\color{dc7}{$z_{6}\triangleleft z_{4}\triangleleft z_{2}\triangleleft z_{2}\triangleleft z_{3}\triangleleft z_{4}=z_{6}$} \quad
\color{dc8}{$z_{6}\triangleleft z_{6}\triangleleft z_{4}\triangleleft z_{4}\triangleleft z_{2}\triangleleft z_{3}=z_{6}$} \quad
\color{dc8}{$z_{6}\triangleleft z_{4}\triangleleft z_{2}\triangleleft z_{6}\triangleleft z_{4}\triangleleft z_{3}=z_{6}$} \quad
\color{dc8}{$z_{4}\triangleleft z_{6}\triangleleft z_{2}\triangleleft z_{3}\triangleleft z_{6}\triangleleft z_{4}=z_{4}$} \quad
\color{dc8}{$z_{4}\triangleleft z_{2}\triangleleft z_{6}\triangleleft z_{4}\triangleleft z_{6}\triangleleft z_{3}=z_{4}$} \quad
\color{dc8}{$z_{6}\triangleleft z_{4}\triangleleft z_{6}\triangleleft z_{3}\triangleleft z_{2}\triangleleft z_{4}=z_{6}$} \quad
\color{dc8}{$z_{4}\triangleleft z_{4}\triangleleft z_{6}\triangleleft z_{2}\triangleleft z_{3}\triangleleft z_{6}=z_{4}$} \quad
\color{dc8}{$z_{6}\triangleleft z_{3}\triangleleft z_{4}\triangleleft z_{6}\triangleleft z_{4}\triangleleft z_{2}=z_{6}$} \quad
\color{dc8}{$z_{3}\triangleleft z_{6}\triangleleft z_{2}\triangleleft z_{4}\triangleleft z_{4}\triangleleft z_{6}=z_{3}$} \quad
\color{dc8}{$z_{6}\triangleleft z_{3}\triangleleft z_{4}\triangleleft z_{4}\triangleleft z_{6}\triangleleft z_{2}=z_{6}$} \quad
\color{dc8}{$z_{6}\triangleleft z_{2}\triangleleft z_{4}\triangleleft z_{4}\triangleleft z_{3}\triangleleft z_{6}=z_{6}$} \quad
\color{dc9}{$z_{5}\triangleleft z_{5}\triangleleft z_{6}\triangleleft z_{2}\triangleleft z_{3}\triangleleft z_{3}=z_{5}$} \quad
\color{dc9}{$z_{5}\triangleleft z_{2}\triangleleft z_{3}\triangleleft z_{5}\triangleleft z_{6}\triangleleft z_{3}=z_{5}$} \quad
\color{dc9}{$z_{6}\triangleleft z_{2}\triangleleft z_{3}\triangleleft z_{5}\triangleleft z_{3}\triangleleft z_{5}=z_{6}$} \quad
\color{dc9}{$z_{5}\triangleleft z_{2}\triangleleft z_{3}\triangleleft z_{3}\triangleleft z_{5}\triangleleft z_{6}=z_{5}$} \quad
\color{dc9}{$z_{3}\triangleleft z_{5}\triangleleft z_{5}\triangleleft z_{2}\triangleleft z_{3}\triangleleft z_{6}=z_{3}$} \quad
\color{dc9}{$z_{5}\triangleleft z_{6}\triangleleft z_{3}\triangleleft z_{2}\triangleleft z_{3}\triangleleft z_{5}=z_{5}$} \quad
\color{dc9}{$z_{2}\triangleleft z_{5}\triangleleft z_{3}\triangleleft z_{3}\triangleleft z_{6}\triangleleft z_{5}=z_{2}$} \quad
\color{dc9}{$z_{6}\triangleleft z_{2}\triangleleft z_{3}\triangleleft z_{3}\triangleleft z_{5}\triangleleft z_{5}=z_{6}$} \quad
\color{dc9}{$z_{3}\triangleleft z_{5}\triangleleft z_{3}\triangleleft z_{6}\triangleleft z_{5}\triangleleft z_{2}=z_{3}$} \quad
\color{dc9}{$z_{6}\triangleleft z_{3}\triangleleft z_{2}\triangleleft z_{3}\triangleleft z_{5}\triangleleft z_{5}=z_{6}$} \quad
\color{dc10}{$z_{5}\triangleleft z_{5}\triangleleft z_{5}\triangleleft z_{3}\triangleleft z_{5}\triangleleft z_{2}=z_{5}$} \quad
\color{dc10}{$z_{5}\triangleleft z_{5}\triangleleft z_{2}\triangleleft z_{5}\triangleleft z_{3}\triangleleft z_{5}=z_{5}$} \quad
\color{dc10}{$z_{3}\triangleleft z_{5}\triangleleft z_{5}\triangleleft z_{2}\triangleleft z_{5}\triangleleft z_{5}=z_{3}$} \quad
\color{dc10}{$z_{5}\triangleleft z_{2}\triangleleft z_{5}\triangleleft z_{5}\triangleleft z_{3}\triangleleft z_{5}=z_{5}$} \quad
\color{dc10}{$z_{5}\triangleleft z_{3}\triangleleft z_{2}\triangleleft z_{5}\triangleleft z_{5}\triangleleft z_{5}=z_{5}$} \quad
\color{dc10}{$z_{3}\triangleleft z_{5}\triangleleft z_{5}\triangleleft z_{5}\triangleleft z_{2}\triangleleft z_{5}=z_{3}$} \quad
\color{dc10}{$z_{5}\triangleleft z_{5}\triangleleft z_{3}\triangleleft z_{5}\triangleleft z_{5}\triangleleft z_{2}=z_{5}$} \quad
\color{dc10}{$z_{5}\triangleleft z_{5}\triangleleft z_{5}\triangleleft z_{3}\triangleleft z_{2}\triangleleft z_{5}=z_{5}$} \quad
\color{dc10}{$z_{5}\triangleleft z_{2}\triangleleft z_{3}\triangleleft z_{5}\triangleleft z_{5}\triangleleft z_{5}=z_{5}$} \quad
\color{dc10}{$z_{2}\triangleleft z_{5}\triangleleft z_{3}\triangleleft z_{5}\triangleleft z_{5}\triangleleft z_{5}=z_{2}$} \quad
\color{dc11}{$z_{6}\triangleleft z_{6}\triangleleft z_{2}\triangleleft z_{5}\triangleleft z_{5}\triangleleft z_{3}=z_{6}$} \quad
\color{dc11}{$z_{6}\triangleleft z_{6}\triangleleft z_{5}\triangleleft z_{2}\triangleleft z_{5}\triangleleft z_{3}=z_{6}$} \quad
\color{dc11}{$z_{5}\triangleleft z_{6}\triangleleft z_{3}\triangleleft z_{2}\triangleleft z_{5}\triangleleft z_{6}=z_{5}$} \quad
\color{dc11}{$z_{5}\triangleleft z_{3}\triangleleft z_{2}\triangleleft z_{6}\triangleleft z_{6}\triangleleft z_{5}=z_{5}$} \quad
\color{dc1}{$z_{6}\triangleleft z_{5}\triangleleft z_{4}\triangleleft z_{1}\triangleleft z_{2}\triangleleft z_{4}=z_{6}$} \quad
\color{dc2}{$z_{6}\triangleleft z_{2}\triangleleft z_{3}\triangleleft z_{5}\triangleleft z_{6}\triangleleft z_{1}=z_{6}$} \quad
\color{dc3}{$z_{4}\triangleleft z_{6}\triangleleft z_{2}\triangleleft z_{2}\triangleleft z_{4}\triangleleft z_{5}=z_{4}$} \quad
\color{dc4}{$z_{2}\triangleleft z_{3}\triangleleft z_{6}\triangleleft z_{1}\triangleleft z_{5}\triangleleft z_{5}=z_{2}$} \quad
\color{dc5}{$z_{5}\triangleleft z_{4}\triangleleft z_{5}\triangleleft z_{5}\triangleleft z_{4}\triangleleft z_{2}=z_{5}$} \quad
\color{dc6}{$z_{4}\triangleleft z_{1}\triangleleft z_{1}\triangleleft z_{4}\triangleleft z_{5}\triangleleft z_{1}=z_{4}$} \quad
\color{dc0}{$z_{0}\triangleleft z_{4}\triangleleft z_{3}\triangleleft z_{5}\triangleleft z_{3}\triangleleft z_{1}=z_{0}$} \quad
\color{dc0}{$z_{4}\triangleleft z_{0}\triangleleft z_{3}\triangleleft z_{5}\triangleleft z_{3}\triangleleft z_{1}=z_{4}$} \quad
\color{dc0}{$z_{4}\triangleleft z_{3}\triangleleft z_{0}\triangleleft z_{5}\triangleleft z_{3}\triangleleft z_{1}=z_{4}$} \quad
\color{dc0}{$z_{4}\triangleleft z_{3}\triangleleft z_{5}\triangleleft z_{0}\triangleleft z_{3}\triangleleft z_{1}=z_{4}$} \quad
\color{dc0}{$z_{4}\triangleleft z_{3}\triangleleft z_{5}\triangleleft z_{3}\triangleleft z_{0}\triangleleft z_{1}=z_{4}$} \quad
\color{dc0}{$z_{4}\triangleleft z_{3}\triangleleft z_{5}\triangleleft z_{3}\triangleleft z_{1}\triangleleft z_{0}=z_{4}$} \quad
\color{dc0}{$z_{4}\triangleleft z_{3}\triangleleft z_{5}\triangleleft z_{3}\triangleleft z_{1}=z_{4}~$} \quad
\color{dc1}{$z_{0}\triangleleft z_{1}\triangleleft z_{5}\triangleleft z_{6}\triangleleft z_{6}\triangleleft z_{1}=z_{0}$} \quad
\color{dc1}{$z_{1}\triangleleft z_{0}\triangleleft z_{5}\triangleleft z_{6}\triangleleft z_{6}\triangleleft z_{1}=z_{1}$} \quad
\color{dc1}{$z_{1}\triangleleft z_{5}\triangleleft z_{0}\triangleleft z_{6}\triangleleft z_{6}\triangleleft z_{1}=z_{1}$} \quad
\color{dc1}{$z_{1}\triangleleft z_{5}\triangleleft z_{6}\triangleleft z_{0}\triangleleft z_{6}\triangleleft z_{1}=z_{1}$} \quad
\color{dc1}{$z_{1}\triangleleft z_{5}\triangleleft z_{6}\triangleleft z_{6}\triangleleft z_{0}\triangleleft z_{1}=z_{1}$} \quad
\color{dc1}{$z_{1}\triangleleft z_{5}\triangleleft z_{6}\triangleleft z_{6}\triangleleft z_{1}\triangleleft z_{0}=z_{1}$} \quad
\color{dc1}{$z_{1}\triangleleft z_{5}\triangleleft z_{6}\triangleleft z_{6}\triangleleft z_{1}=z_{1}~$} \quad
\color{dc2}{$z_{0}\triangleleft z_{5}\triangleleft z_{2}\triangleleft z_{5}\triangleleft z_{2}\triangleleft z_{3}=z_{0}$} \quad
\color{dc2}{$z_{5}\triangleleft z_{0}\triangleleft z_{2}\triangleleft z_{5}\triangleleft z_{2}\triangleleft z_{3}=z_{5}$} \quad
\color{dc2}{$z_{5}\triangleleft z_{2}\triangleleft z_{0}\triangleleft z_{5}\triangleleft z_{2}\triangleleft z_{3}=z_{5}$} \quad
\color{dc2}{$z_{5}\triangleleft z_{2}\triangleleft z_{5}\triangleleft z_{0}\triangleleft z_{2}\triangleleft z_{3}=z_{5}$} \quad
\color{dc2}{$z_{5}\triangleleft z_{2}\triangleleft z_{5}\triangleleft z_{2}\triangleleft z_{0}\triangleleft z_{3}=z_{5}$} \quad
\color{dc2}{$z_{5}\triangleleft z_{2}\triangleleft z_{5}\triangleleft z_{2}\triangleleft z_{3}\triangleleft z_{0}=z_{5}$} \quad
\color{dc2}{$z_{5}\triangleleft z_{2}\triangleleft z_{5}\triangleleft z_{2}\triangleleft z_{3}=z_{5}~$} \quad
\color{dc3}{$z_{0}\triangleleft z_{3}\triangleleft z_{1}\triangleleft z_{2}\triangleleft z_{2}\triangleleft z_{2}=z_{0}$} \quad
\color{dc3}{$z_{3}\triangleleft z_{0}\triangleleft z_{1}\triangleleft z_{2}\triangleleft z_{2}\triangleleft z_{2}=z_{3}$} \quad
\color{dc3}{$z_{3}\triangleleft z_{1}\triangleleft z_{0}\triangleleft z_{2}\triangleleft z_{2}\triangleleft z_{2}=z_{3}$} \quad
\color{dc3}{$z_{3}\triangleleft z_{1}\triangleleft z_{2}\triangleleft z_{0}\triangleleft z_{2}\triangleleft z_{2}=z_{3}$} \quad
\color{dc3}{$z_{3}\triangleleft z_{1}\triangleleft z_{2}\triangleleft z_{2}\triangleleft z_{0}\triangleleft z_{2}=z_{3}$} \quad
\color{dc3}{$z_{3}\triangleleft z_{1}\triangleleft z_{2}\triangleleft z_{2}\triangleleft z_{2}\triangleleft z_{0}=z_{3}$} \quad
\color{dc3}{$z_{3}\triangleleft z_{1}\triangleleft z_{2}\triangleleft z_{2}\triangleleft z_{2}=z_{3}~$} \quad
\color{dc4}{$z_{0}\triangleleft z_{2}\triangleleft z_{4}\triangleleft z_{4}\triangleleft z_{3}\triangleleft z_{4}=z_{0}$} \quad
\color{dc4}{$z_{2}\triangleleft z_{0}\triangleleft z_{4}\triangleleft z_{4}\triangleleft z_{3}\triangleleft z_{4}=z_{2}$} \quad
\color{dc4}{$z_{2}\triangleleft z_{4}\triangleleft z_{0}\triangleleft z_{4}\triangleleft z_{3}\triangleleft z_{4}=z_{2}$} \quad
\color{dc4}{$z_{2}\triangleleft z_{4}\triangleleft z_{4}\triangleleft z_{0}\triangleleft z_{3}\triangleleft z_{4}=z_{2}$} \quad
\color{dc4}{$z_{2}\triangleleft z_{4}\triangleleft z_{4}\triangleleft z_{3}\triangleleft z_{0}\triangleleft z_{4}=z_{2}$} \quad
\color{dc4}{$z_{2}\triangleleft z_{4}\triangleleft z_{4}\triangleleft z_{3}\triangleleft z_{4}\triangleleft z_{0}=z_{2}$} \quad
\color{dc4}{$z_{2}\triangleleft z_{4}\triangleleft z_{4}\triangleleft z_{3}\triangleleft z_{4}=z_{2}~$} \quad
\color{dc5}{$z_{0}\triangleleft z_{1}\triangleleft z_{4}\triangleleft z_{6}\triangleleft z_{5}\triangleleft z_{2}=z_{0}$} \quad
\color{dc5}{$z_{1}\triangleleft z_{0}\triangleleft z_{4}\triangleleft z_{6}\triangleleft z_{5}\triangleleft z_{2}=z_{1}$} \quad
\color{dc5}{$z_{1}\triangleleft z_{4}\triangleleft z_{0}\triangleleft z_{6}\triangleleft z_{5}\triangleleft z_{2}=z_{1}$} \quad
\color{dc5}{$z_{1}\triangleleft z_{4}\triangleleft z_{6}\triangleleft z_{0}\triangleleft z_{5}\triangleleft z_{2}=z_{1}$} \quad
\color{dc5}{$z_{1}\triangleleft z_{4}\triangleleft z_{6}\triangleleft z_{5}\triangleleft z_{0}\triangleleft z_{2}=z_{1}$} \quad
\color{dc5}{$z_{1}\triangleleft z_{4}\triangleleft z_{6}\triangleleft z_{5}\triangleleft z_{2}=z_{1}~$} \quad
\color{dc11}{$z_{5}\triangleleft z_{6}\triangleleft z_{1}\triangleleft z_{4}\triangleleft z_{4}=z_{5}~$} \quad
\color{dc12}{$z_{4}\triangleleft z_{2}\triangleleft z_{4}\triangleleft z_{6}\triangleleft z_{3}=z_{4}~$} \quad
\color{dc13}{$z_{5}\triangleleft z_{1}\triangleleft z_{1}\triangleleft z_{5}\triangleleft z_{3}=z_{5}~$} \quad
\color{dc14}{$z_{2}\triangleleft z_{1}\triangleleft z_{2}\triangleleft z_{6}\triangleleft z_{2}=z_{2}~$} \quad
\color{dc15}{$z_{6}\triangleleft z_{1}\triangleleft z_{2}\triangleleft z_{5}\triangleleft z_{4}=z_{6}~$} \quad
\color{dc16}{$z_{1}\triangleleft z_{4}\triangleleft z_{6}\triangleleft z_{1}\triangleleft z_{3}=z_{1}~$} \quad
\color{dc17}{$z_{4}\triangleleft z_{1}\triangleleft z_{2}\triangleleft z_{4}\triangleleft z_{6}=z_{4}~$} \quad
\color{dc18}{$z_{4}\triangleleft z_{4}\triangleleft z_{1}\triangleleft z_{2}\triangleleft z_{2}=z_{4}~$} \quad
\color{dc19}{$z_{2}\triangleleft z_{1}\triangleleft z_{3}\triangleleft z_{5}\triangleleft z_{4}=z_{2}~$} \quad
}

\paragraph{\color{black}{Testing, for Operator $\triangleleft$}}\mbox{}\\
{\footnotesize \color{dc1}{$z_{4}\triangleleft z_{4}\triangleleft z_{5}\triangleleft z_{6}\triangleleft z_{2}\triangleleft z_{1}=z_{4}$} \quad
\color{dc1}{$z_{4}\triangleleft z_{2}\triangleleft z_{6}\triangleleft z_{5}\triangleleft z_{4}\triangleleft z_{1}=z_{4}$} \quad
\color{dc1}{$z_{4}\triangleleft z_{4}\triangleleft z_{2}\triangleleft z_{1}\triangleleft z_{6}\triangleleft z_{5}=z_{4}$} \quad
\color{dc1}{$z_{6}\triangleleft z_{2}\triangleleft z_{1}\triangleleft z_{5}\triangleleft z_{4}\triangleleft z_{4}=z_{6}$} \quad
\color{dc1}{$z_{6}\triangleleft z_{4}\triangleleft z_{1}\triangleleft z_{4}\triangleleft z_{2}\triangleleft z_{5}=z_{6}$} \quad
\color{dc1}{$z_{5}\triangleleft z_{2}\triangleleft z_{4}\triangleleft z_{4}\triangleleft z_{6}\triangleleft z_{1}=z_{5}$} \quad
\color{dc1}{$z_{4}\triangleleft z_{5}\triangleleft z_{4}\triangleleft z_{6}\triangleleft z_{1}\triangleleft z_{2}=z_{4}$} \quad
\color{dc1}{$z_{6}\triangleleft z_{5}\triangleleft z_{1}\triangleleft z_{4}\triangleleft z_{2}\triangleleft z_{4}=z_{6}$} \quad
\color{dc1}{$z_{1}\triangleleft z_{2}\triangleleft z_{4}\triangleleft z_{6}\triangleleft z_{5}\triangleleft z_{4}=z_{1}$} \quad
\color{dc2}{$z_{6}\triangleleft z_{1}\triangleleft z_{6}\triangleleft z_{2}\triangleleft z_{3}\triangleleft z_{5}=z_{6}$} \quad
\color{dc2}{$z_{2}\triangleleft z_{5}\triangleleft z_{3}\triangleleft z_{1}\triangleleft z_{6}\triangleleft z_{6}=z_{2}$} \quad
\color{dc2}{$z_{3}\triangleleft z_{2}\triangleleft z_{6}\triangleleft z_{1}\triangleleft z_{5}\triangleleft z_{6}=z_{3}$} \quad
\color{dc2}{$z_{6}\triangleleft z_{2}\triangleleft z_{5}\triangleleft z_{1}\triangleleft z_{3}\triangleleft z_{6}=z_{6}$} \quad
\color{dc2}{$z_{6}\triangleleft z_{6}\triangleleft z_{3}\triangleleft z_{5}\triangleleft z_{2}\triangleleft z_{1}=z_{6}$} \quad
\color{dc2}{$z_{1}\triangleleft z_{6}\triangleleft z_{5}\triangleleft z_{6}\triangleleft z_{3}\triangleleft z_{2}=z_{1}$} \quad
\color{dc2}{$z_{5}\triangleleft z_{1}\triangleleft z_{6}\triangleleft z_{3}\triangleleft z_{6}\triangleleft z_{2}=z_{5}$} \quad
\color{dc2}{$z_{5}\triangleleft z_{6}\triangleleft z_{6}\triangleleft z_{3}\triangleleft z_{1}\triangleleft z_{2}=z_{5}$} \quad
\color{dc2}{$z_{5}\triangleleft z_{3}\triangleleft z_{6}\triangleleft z_{2}\triangleleft z_{6}\triangleleft z_{1}=z_{5}$} \quad
\color{dc3}{$z_{2}\triangleleft z_{4}\triangleleft z_{5}\triangleleft z_{2}\triangleleft z_{4}\triangleleft z_{6}=z_{2}$} \quad
\color{dc3}{$z_{4}\triangleleft z_{5}\triangleleft z_{4}\triangleleft z_{2}\triangleleft z_{2}\triangleleft z_{6}=z_{4}$} \quad
\color{dc3}{$z_{2}\triangleleft z_{5}\triangleleft z_{2}\triangleleft z_{4}\triangleleft z_{4}\triangleleft z_{6}=z_{2}$} \quad
\color{dc3}{$z_{5}\triangleleft z_{2}\triangleleft z_{6}\triangleleft z_{4}\triangleleft z_{2}\triangleleft z_{4}=z_{5}$} \quad
\color{dc3}{$z_{5}\triangleleft z_{4}\triangleleft z_{6}\triangleleft z_{2}\triangleleft z_{4}\triangleleft z_{2}=z_{5}$} \quad
\color{dc3}{$z_{2}\triangleleft z_{4}\triangleleft z_{2}\triangleleft z_{6}\triangleleft z_{4}\triangleleft z_{5}=z_{2}$} \quad
\color{dc3}{$z_{4}\triangleleft z_{2}\triangleleft z_{4}\triangleleft z_{5}\triangleleft z_{6}\triangleleft z_{2}=z_{4}$} \quad
\color{dc3}{$z_{2}\triangleleft z_{4}\triangleleft z_{5}\triangleleft z_{6}\triangleleft z_{4}\triangleleft z_{2}=z_{2}$} \quad
\color{dc3}{$z_{4}\triangleleft z_{5}\triangleleft z_{2}\triangleleft z_{2}\triangleleft z_{4}\triangleleft z_{6}=z_{4}$} \quad
\color{dc4}{$z_{2}\triangleleft z_{1}\triangleleft z_{6}\triangleleft z_{5}\triangleleft z_{5}\triangleleft z_{3}=z_{2}$} \quad
\color{dc4}{$z_{6}\triangleleft z_{5}\triangleleft z_{3}\triangleleft z_{1}\triangleleft z_{5}\triangleleft z_{2}=z_{6}$} \quad
\color{dc4}{$z_{6}\triangleleft z_{2}\triangleleft z_{5}\triangleleft z_{5}\triangleleft z_{1}\triangleleft z_{3}=z_{6}$} \quad
\color{dc4}{$z_{3}\triangleleft z_{6}\triangleleft z_{1}\triangleleft z_{5}\triangleleft z_{2}\triangleleft z_{5}=z_{3}$} \quad
\color{dc4}{$z_{5}\triangleleft z_{2}\triangleleft z_{5}\triangleleft z_{1}\triangleleft z_{6}\triangleleft z_{3}=z_{5}$} \quad
\color{dc4}{$z_{3}\triangleleft z_{1}\triangleleft z_{6}\triangleleft z_{2}\triangleleft z_{5}\triangleleft z_{5}=z_{3}$} \quad
\color{dc4}{$z_{3}\triangleleft z_{6}\triangleleft z_{5}\triangleleft z_{2}\triangleleft z_{5}\triangleleft z_{1}=z_{3}$} \quad
\color{dc4}{$z_{6}\triangleleft z_{5}\triangleleft z_{1}\triangleleft z_{3}\triangleleft z_{2}\triangleleft z_{5}=z_{6}$} \quad
\color{dc4}{$z_{1}\triangleleft z_{5}\triangleleft z_{5}\triangleleft z_{2}\triangleleft z_{6}\triangleleft z_{3}=z_{1}$} \quad
\color{dc5}{$z_{2}\triangleleft z_{5}\triangleleft z_{5}\triangleleft z_{5}\triangleleft z_{4}\triangleleft z_{4}=z_{2}$} \quad
\color{dc5}{$z_{4}\triangleleft z_{5}\triangleleft z_{5}\triangleleft z_{2}\triangleleft z_{5}\triangleleft z_{4}=z_{4}$} \quad
\color{dc5}{$z_{2}\triangleleft z_{5}\triangleleft z_{5}\triangleleft z_{4}\triangleleft z_{5}\triangleleft z_{4}=z_{2}$} \quad
\color{dc5}{$z_{2}\triangleleft z_{5}\triangleleft z_{4}\triangleleft z_{5}\triangleleft z_{5}\triangleleft z_{4}=z_{2}$} \quad
\color{dc5}{$z_{5}\triangleleft z_{4}\triangleleft z_{5}\triangleleft z_{5}\triangleleft z_{2}\triangleleft z_{4}=z_{5}$} \quad
\color{dc5}{$z_{2}\triangleleft z_{4}\triangleleft z_{5}\triangleleft z_{4}\triangleleft z_{5}\triangleleft z_{5}=z_{2}$} \quad
\color{dc5}{$z_{5}\triangleleft z_{4}\triangleleft z_{2}\triangleleft z_{5}\triangleleft z_{5}\triangleleft z_{4}=z_{5}$} \quad
\color{dc5}{$z_{5}\triangleleft z_{4}\triangleleft z_{5}\triangleleft z_{2}\triangleleft z_{5}\triangleleft z_{4}=z_{5}$} \quad
\color{dc5}{$z_{5}\triangleleft z_{2}\triangleleft z_{4}\triangleleft z_{5}\triangleleft z_{5}\triangleleft z_{4}=z_{5}$} \quad
\color{dc6}{$z_{4}\triangleleft z_{1}\triangleleft z_{4}\triangleleft z_{5}\triangleleft z_{1}\triangleleft z_{1}=z_{4}$} \quad
\color{dc6}{$z_{1}\triangleleft z_{5}\triangleleft z_{4}\triangleleft z_{1}\triangleleft z_{1}\triangleleft z_{4}=z_{1}$} \quad
\color{dc6}{$z_{1}\triangleleft z_{1}\triangleleft z_{4}\triangleleft z_{1}\triangleleft z_{5}\triangleleft z_{4}=z_{1}$} \quad
\color{dc6}{$z_{4}\triangleleft z_{1}\triangleleft z_{1}\triangleleft z_{5}\triangleleft z_{4}\triangleleft z_{1}=z_{4}$} \quad
\color{dc6}{$z_{1}\triangleleft z_{5}\triangleleft z_{1}\triangleleft z_{4}\triangleleft z_{1}\triangleleft z_{4}=z_{1}$} \quad
\color{dc11}{$z_{0}\triangleleft z_{5}\triangleleft z_{6}\triangleleft z_{1}\triangleleft z_{4}\triangleleft z_{4}=z_{0}$} \quad
\color{dc11}{$z_{5}\triangleleft z_{0}\triangleleft z_{6}\triangleleft z_{1}\triangleleft z_{4}\triangleleft z_{4}=z_{5}$} \quad
\color{dc11}{$z_{5}\triangleleft z_{6}\triangleleft z_{0}\triangleleft z_{1}\triangleleft z_{4}\triangleleft z_{4}=z_{5}$} \quad
\color{dc11}{$z_{5}\triangleleft z_{6}\triangleleft z_{1}\triangleleft z_{0}\triangleleft z_{4}\triangleleft z_{4}=z_{5}$} \quad
\color{dc11}{$z_{5}\triangleleft z_{6}\triangleleft z_{1}\triangleleft z_{4}\triangleleft z_{0}\triangleleft z_{4}=z_{5}$} \quad
\color{dc11}{$z_{5}\triangleleft z_{6}\triangleleft z_{1}\triangleleft z_{4}\triangleleft z_{4}\triangleleft z_{0}=z_{5}$} \quad
\color{dc12}{$z_{0}\triangleleft z_{4}\triangleleft z_{2}\triangleleft z_{4}\triangleleft z_{6}\triangleleft z_{3}=z_{0}$} \quad
\color{dc12}{$z_{4}\triangleleft z_{0}\triangleleft z_{2}\triangleleft z_{4}\triangleleft z_{6}\triangleleft z_{3}=z_{4}$} \quad
\color{dc12}{$z_{4}\triangleleft z_{2}\triangleleft z_{0}\triangleleft z_{4}\triangleleft z_{6}\triangleleft z_{3}=z_{4}$} \quad
\color{dc12}{$z_{4}\triangleleft z_{2}\triangleleft z_{4}\triangleleft z_{0}\triangleleft z_{6}\triangleleft z_{3}=z_{4}$} \quad
\color{dc12}{$z_{4}\triangleleft z_{2}\triangleleft z_{4}\triangleleft z_{6}\triangleleft z_{0}\triangleleft z_{3}=z_{4}$} \quad
\color{dc12}{$z_{4}\triangleleft z_{2}\triangleleft z_{4}\triangleleft z_{6}\triangleleft z_{3}\triangleleft z_{0}=z_{4}$} \quad
\color{dc13}{$z_{0}\triangleleft z_{5}\triangleleft z_{1}\triangleleft z_{1}\triangleleft z_{5}\triangleleft z_{3}=z_{0}$} \quad
\color{dc13}{$z_{5}\triangleleft z_{0}\triangleleft z_{1}\triangleleft z_{1}\triangleleft z_{5}\triangleleft z_{3}=z_{5}$} \quad
\color{dc13}{$z_{5}\triangleleft z_{1}\triangleleft z_{0}\triangleleft z_{1}\triangleleft z_{5}\triangleleft z_{3}=z_{5}$} \quad
\color{dc13}{$z_{5}\triangleleft z_{1}\triangleleft z_{1}\triangleleft z_{0}\triangleleft z_{5}\triangleleft z_{3}=z_{5}$} \quad
\color{dc13}{$z_{5}\triangleleft z_{1}\triangleleft z_{1}\triangleleft z_{5}\triangleleft z_{0}\triangleleft z_{3}=z_{5}$} \quad
\color{dc13}{$z_{5}\triangleleft z_{1}\triangleleft z_{1}\triangleleft z_{5}\triangleleft z_{3}\triangleleft z_{0}=z_{5}$} \quad
\color{dc14}{$z_{0}\triangleleft z_{2}\triangleleft z_{1}\triangleleft z_{2}\triangleleft z_{6}\triangleleft z_{2}=z_{0}$} \quad
\color{dc14}{$z_{2}\triangleleft z_{0}\triangleleft z_{1}\triangleleft z_{2}\triangleleft z_{6}\triangleleft z_{2}=z_{2}$} \quad
\color{dc14}{$z_{2}\triangleleft z_{1}\triangleleft z_{0}\triangleleft z_{2}\triangleleft z_{6}\triangleleft z_{2}=z_{2}$} \quad
\color{dc14}{$z_{2}\triangleleft z_{1}\triangleleft z_{2}\triangleleft z_{0}\triangleleft z_{6}\triangleleft z_{2}=z_{2}$} \quad
\color{dc14}{$z_{2}\triangleleft z_{1}\triangleleft z_{2}\triangleleft z_{6}\triangleleft z_{0}\triangleleft z_{2}=z_{2}$} \quad
\color{dc14}{$z_{2}\triangleleft z_{1}\triangleleft z_{2}\triangleleft z_{6}\triangleleft z_{2}\triangleleft z_{0}=z_{2}$} \quad
\color{dc15}{$z_{0}\triangleleft z_{6}\triangleleft z_{1}\triangleleft z_{2}\triangleleft z_{5}\triangleleft z_{4}=z_{0}$} \quad
\color{dc15}{$z_{6}\triangleleft z_{0}\triangleleft z_{1}\triangleleft z_{2}\triangleleft z_{5}\triangleleft z_{4}=z_{6}$} \quad
\color{dc15}{$z_{6}\triangleleft z_{1}\triangleleft z_{0}\triangleleft z_{2}\triangleleft z_{5}\triangleleft z_{4}=z_{6}$} \quad
\color{dc15}{$z_{6}\triangleleft z_{1}\triangleleft z_{2}\triangleleft z_{0}\triangleleft z_{5}\triangleleft z_{4}=z_{6}$} \quad
\color{dc15}{$z_{6}\triangleleft z_{1}\triangleleft z_{2}\triangleleft z_{5}\triangleleft z_{0}\triangleleft z_{4}=z_{6}$} \quad
\color{dc15}{$z_{6}\triangleleft z_{1}\triangleleft z_{2}\triangleleft z_{5}\triangleleft z_{4}\triangleleft z_{0}=z_{6}$} \quad
\color{dc16}{$z_{0}\triangleleft z_{1}\triangleleft z_{4}\triangleleft z_{6}\triangleleft z_{1}\triangleleft z_{3}=z_{0}$} \quad
\color{dc16}{$z_{1}\triangleleft z_{0}\triangleleft z_{4}\triangleleft z_{6}\triangleleft z_{1}\triangleleft z_{3}=z_{1}$} \quad
\color{dc16}{$z_{1}\triangleleft z_{4}\triangleleft z_{0}\triangleleft z_{6}\triangleleft z_{1}\triangleleft z_{3}=z_{1}$} \quad
\color{dc16}{$z_{1}\triangleleft z_{4}\triangleleft z_{6}\triangleleft z_{0}\triangleleft z_{1}\triangleleft z_{3}=z_{1}$} \quad
\color{dc16}{$z_{1}\triangleleft z_{4}\triangleleft z_{6}\triangleleft z_{1}\triangleleft z_{0}\triangleleft z_{3}=z_{1}$} \quad
\color{dc16}{$z_{1}\triangleleft z_{4}\triangleleft z_{6}\triangleleft z_{1}\triangleleft z_{3}\triangleleft z_{0}=z_{1}$} \quad
\color{dc17}{$z_{0}\triangleleft z_{4}\triangleleft z_{1}\triangleleft z_{2}\triangleleft z_{4}\triangleleft z_{6}=z_{0}$} \quad
\color{dc17}{$z_{4}\triangleleft z_{0}\triangleleft z_{1}\triangleleft z_{2}\triangleleft z_{4}\triangleleft z_{6}=z_{4}$} \quad
\color{dc17}{$z_{4}\triangleleft z_{1}\triangleleft z_{0}\triangleleft z_{2}\triangleleft z_{4}\triangleleft z_{6}=z_{4}$} \quad
\color{dc17}{$z_{4}\triangleleft z_{1}\triangleleft z_{2}\triangleleft z_{0}\triangleleft z_{4}\triangleleft z_{6}=z_{4}$} \quad
\color{dc17}{$z_{4}\triangleleft z_{1}\triangleleft z_{2}\triangleleft z_{4}\triangleleft z_{0}\triangleleft z_{6}=z_{4}$} \quad
\color{dc17}{$z_{4}\triangleleft z_{1}\triangleleft z_{2}\triangleleft z_{4}\triangleleft z_{6}\triangleleft z_{0}=z_{4}$} \quad
\color{dc18}{$z_{0}\triangleleft z_{4}\triangleleft z_{4}\triangleleft z_{1}\triangleleft z_{2}\triangleleft z_{2}=z_{0}$} \quad
\color{dc18}{$z_{4}\triangleleft z_{0}\triangleleft z_{4}\triangleleft z_{1}\triangleleft z_{2}\triangleleft z_{2}=z_{4}$} \quad
\color{dc18}{$z_{4}\triangleleft z_{4}\triangleleft z_{0}\triangleleft z_{1}\triangleleft z_{2}\triangleleft z_{2}=z_{4}$} \quad
\color{dc18}{$z_{4}\triangleleft z_{4}\triangleleft z_{1}\triangleleft z_{0}\triangleleft z_{2}\triangleleft z_{2}=z_{4}$} \quad
\color{dc18}{$z_{4}\triangleleft z_{4}\triangleleft z_{1}\triangleleft z_{2}\triangleleft z_{0}\triangleleft z_{2}=z_{4}$} \quad
\color{dc18}{$z_{4}\triangleleft z_{4}\triangleleft z_{1}\triangleleft z_{2}\triangleleft z_{2}\triangleleft z_{0}=z_{4}$} \quad
\color{dc19}{$z_{0}\triangleleft z_{2}\triangleleft z_{1}\triangleleft z_{3}\triangleleft z_{5}\triangleleft z_{4}=z_{0}$} \quad
\color{dc19}{$z_{2}\triangleleft z_{0}\triangleleft z_{1}\triangleleft z_{3}\triangleleft z_{5}\triangleleft z_{4}=z_{2}$} \quad

\paragraph{\color{black}{Training, for Operator $\triangleright$}}\mbox{}\\
{\footnotesize \color{dc7}{$z_{2}\triangleright z_{2}\triangleright z_{4}\triangleright z_{3}\triangleright z_{6}\triangleright z_{4}=z_{4}$} \quad
\color{dc7}{$z_{4}\triangleright z_{3}\triangleright z_{6}\triangleright z_{4}\triangleright z_{2}\triangleright z_{2}=z_{2}$} \quad
\color{dc7}{$z_{3}\triangleright z_{6}\triangleright z_{4}\triangleright z_{4}\triangleright z_{2}\triangleright z_{2}=z_{2}$} \quad
\color{dc7}{$z_{6}\triangleright z_{2}\triangleright z_{3}\triangleright z_{2}\triangleright z_{4}\triangleright z_{4}=z_{4}$} \quad
\color{dc7}{$z_{2}\triangleright z_{3}\triangleright z_{4}\triangleright z_{2}\triangleright z_{6}\triangleright z_{4}=z_{4}$} \quad
\color{dc7}{$z_{2}\triangleright z_{6}\triangleright z_{4}\triangleright z_{4}\triangleright z_{3}\triangleright z_{2}=z_{2}$} \quad
\color{dc7}{$z_{6}\triangleright z_{2}\triangleright z_{4}\triangleright z_{2}\triangleright z_{4}\triangleright z_{3}=z_{3}$} \quad
\color{dc7}{$z_{4}\triangleright z_{2}\triangleright z_{6}\triangleright z_{2}\triangleright z_{3}\triangleright z_{4}=z_{4}$} \quad
\color{dc7}{$z_{6}\triangleright z_{2}\triangleright z_{2}\triangleright z_{3}\triangleright z_{4}\triangleright z_{4}=z_{4}$} \quad
\color{dc7}{$z_{6}\triangleright z_{4}\triangleright z_{2}\triangleright z_{2}\triangleright z_{3}\triangleright z_{4}=z_{4}$} \quad
\color{dc8}{$z_{6}\triangleright z_{6}\triangleright z_{4}\triangleright z_{4}\triangleright z_{2}\triangleright z_{3}=z_{3}$} \quad
\color{dc8}{$z_{6}\triangleright z_{4}\triangleright z_{2}\triangleright z_{6}\triangleright z_{4}\triangleright z_{3}=z_{3}$} \quad
\color{dc8}{$z_{4}\triangleright z_{6}\triangleright z_{2}\triangleright z_{3}\triangleright z_{6}\triangleright z_{4}=z_{4}$} \quad
\color{dc8}{$z_{4}\triangleright z_{2}\triangleright z_{6}\triangleright z_{4}\triangleright z_{6}\triangleright z_{3}=z_{3}$} \quad
\color{dc8}{$z_{6}\triangleright z_{4}\triangleright z_{6}\triangleright z_{3}\triangleright z_{2}\triangleright z_{4}=z_{4}$} \quad
\color{dc8}{$z_{4}\triangleright z_{4}\triangleright z_{6}\triangleright z_{2}\triangleright z_{3}\triangleright z_{6}=z_{6}$} \quad
\color{dc8}{$z_{6}\triangleright z_{3}\triangleright z_{4}\triangleright z_{6}\triangleright z_{4}\triangleright z_{2}=z_{2}$} \quad
\color{dc8}{$z_{3}\triangleright z_{6}\triangleright z_{2}\triangleright z_{4}\triangleright z_{4}\triangleright z_{6}=z_{6}$} \quad
\color{dc8}{$z_{6}\triangleright z_{3}\triangleright z_{4}\triangleright z_{4}\triangleright z_{6}\triangleright z_{2}=z_{2}$} \quad
\color{dc8}{$z_{6}\triangleright z_{2}\triangleright z_{4}\triangleright z_{4}\triangleright z_{3}\triangleright z_{6}=z_{6}$} \quad
\color{dc9}{$z_{5}\triangleright z_{5}\triangleright z_{6}\triangleright z_{2}\triangleright z_{3}\triangleright z_{3}=z_{3}$} \quad
\color{dc9}{$z_{5}\triangleright z_{2}\triangleright z_{3}\triangleright z_{5}\triangleright z_{6}\triangleright z_{3}=z_{3}$} \quad
\color{dc9}{$z_{6}\triangleright z_{2}\triangleright z_{3}\triangleright z_{5}\triangleright z_{3}\triangleright z_{5}=z_{5}$} \quad
\color{dc9}{$z_{5}\triangleright z_{2}\triangleright z_{3}\triangleright z_{3}\triangleright z_{5}\triangleright z_{6}=z_{6}$} \quad
\color{dc9}{$z_{3}\triangleright z_{5}\triangleright z_{5}\triangleright z_{2}\triangleright z_{3}\triangleright z_{6}=z_{6}$} \quad
\color{dc9}{$z_{5}\triangleright z_{6}\triangleright z_{3}\triangleright z_{2}\triangleright z_{3}\triangleright z_{5}=z_{5}$} \quad
\color{dc9}{$z_{2}\triangleright z_{5}\triangleright z_{3}\triangleright z_{3}\triangleright z_{6}\triangleright z_{5}=z_{5}$} \quad
\color{dc9}{$z_{6}\triangleright z_{2}\triangleright z_{3}\triangleright z_{3}\triangleright z_{5}\triangleright z_{5}=z_{5}$} \quad
\color{dc9}{$z_{3}\triangleright z_{5}\triangleright z_{3}\triangleright z_{6}\triangleright z_{5}\triangleright z_{2}=z_{2}$} \quad
\color{dc9}{$z_{6}\triangleright z_{3}\triangleright z_{2}\triangleright z_{3}\triangleright z_{5}\triangleright z_{5}=z_{5}$} \quad
\color{dc10}{$z_{5}\triangleright z_{5}\triangleright z_{5}\triangleright z_{3}\triangleright z_{5}\triangleright z_{2}=z_{2}$} \quad
\color{dc10}{$z_{5}\triangleright z_{5}\triangleright z_{2}\triangleright z_{5}\triangleright z_{3}\triangleright z_{5}=z_{5}$} \quad
\color{dc10}{$z_{3}\triangleright z_{5}\triangleright z_{5}\triangleright z_{2}\triangleright z_{5}\triangleright z_{5}=z_{5}$} \quad
\color{dc10}{$z_{5}\triangleright z_{2}\triangleright z_{5}\triangleright z_{5}\triangleright z_{3}\triangleright z_{5}=z_{5}$} \quad
\color{dc10}{$z_{5}\triangleright z_{3}\triangleright z_{2}\triangleright z_{5}\triangleright z_{5}\triangleright z_{5}=z_{5}$} \quad
\color{dc10}{$z_{3}\triangleright z_{5}\triangleright z_{5}\triangleright z_{5}\triangleright z_{2}\triangleright z_{5}=z_{5}$} \quad
\color{dc10}{$z_{5}\triangleright z_{5}\triangleright z_{3}\triangleright z_{5}\triangleright z_{5}\triangleright z_{2}=z_{2}$} \quad
\color{dc10}{$z_{5}\triangleright z_{5}\triangleright z_{5}\triangleright z_{3}\triangleright z_{2}\triangleright z_{5}=z_{5}$} \quad
\color{dc10}{$z_{5}\triangleright z_{2}\triangleright z_{3}\triangleright z_{5}\triangleright z_{5}\triangleright z_{5}=z_{5}$} \quad
\color{dc10}{$z_{2}\triangleright z_{5}\triangleright z_{3}\triangleright z_{5}\triangleright z_{5}\triangleright z_{5}=z_{5}$} \quad
\color{dc11}{$z_{6}\triangleright z_{6}\triangleright z_{2}\triangleright z_{5}\triangleright z_{5}\triangleright z_{3}=z_{3}$} \quad
\color{dc11}{$z_{6}\triangleright z_{6}\triangleright z_{5}\triangleright z_{2}\triangleright z_{5}\triangleright z_{3}=z_{3}$} \quad
\color{dc11}{$z_{5}\triangleright z_{6}\triangleright z_{3}\triangleright z_{2}\triangleright z_{5}\triangleright z_{6}=z_{6}$} \quad
\color{dc11}{$z_{5}\triangleright z_{3}\triangleright z_{2}\triangleright z_{6}\triangleright z_{6}\triangleright z_{5}=z_{5}$} \quad
\color{dc1}{$z_{6}\triangleright z_{5}\triangleright z_{4}\triangleright z_{1}\triangleright z_{2}\triangleright z_{4}=z_{4}$} \quad
\color{dc2}{$z_{6}\triangleright z_{2}\triangleright z_{3}\triangleright z_{5}\triangleright z_{6}\triangleright z_{1}=z_{1}$} \quad
\color{dc3}{$z_{4}\triangleright z_{6}\triangleright z_{2}\triangleright z_{2}\triangleright z_{4}\triangleright z_{5}=z_{5}$} \quad
\color{dc4}{$z_{2}\triangleright z_{3}\triangleright z_{6}\triangleright z_{1}\triangleright z_{5}\triangleright z_{5}=z_{5}$} \quad
\color{dc5}{$z_{5}\triangleright z_{4}\triangleright z_{5}\triangleright z_{5}\triangleright z_{4}\triangleright z_{2}=z_{2}$} \quad
\color{dc6}{$z_{4}\triangleright z_{1}\triangleright z_{1}\triangleright z_{4}\triangleright z_{5}\triangleright z_{1}=z_{1}$} \quad
\color{dc0}{$z_{0}\triangleright z_{4}\triangleright z_{3}\triangleright z_{5}\triangleright z_{3}\triangleright z_{1}=z_{1}$} \quad
\color{dc0}{$z_{4}\triangleright z_{0}\triangleright z_{3}\triangleright z_{5}\triangleright z_{3}\triangleright z_{1}=z_{1}$} \quad
\color{dc0}{$z_{4}\triangleright z_{3}\triangleright z_{0}\triangleright z_{5}\triangleright z_{3}\triangleright z_{1}=z_{1}$} \quad
\color{dc0}{$z_{4}\triangleright z_{3}\triangleright z_{5}\triangleright z_{0}\triangleright z_{3}\triangleright z_{1}=z_{1}$} \quad
\color{dc0}{$z_{4}\triangleright z_{3}\triangleright z_{5}\triangleright z_{3}\triangleright z_{0}\triangleright z_{1}=z_{1}$} \quad
\color{dc0}{$z_{4}\triangleright z_{3}\triangleright z_{5}\triangleright z_{3}\triangleright z_{1}\triangleright z_{0}=z_{0}$} \quad
\color{dc0}{$z_{4}\triangleright z_{3}\triangleright z_{5}\triangleright z_{3}\triangleright z_{1}=z_{1}~$} \quad
\color{dc1}{$z_{0}\triangleright z_{1}\triangleright z_{5}\triangleright z_{6}\triangleright z_{6}\triangleright z_{1}=z_{1}$} \quad
\color{dc1}{$z_{1}\triangleright z_{0}\triangleright z_{5}\triangleright z_{6}\triangleright z_{6}\triangleright z_{1}=z_{1}$} \quad
\color{dc1}{$z_{1}\triangleright z_{5}\triangleright z_{0}\triangleright z_{6}\triangleright z_{6}\triangleright z_{1}=z_{1}$} \quad
\color{dc1}{$z_{1}\triangleright z_{5}\triangleright z_{6}\triangleright z_{0}\triangleright z_{6}\triangleright z_{1}=z_{1}$} \quad
\color{dc1}{$z_{1}\triangleright z_{5}\triangleright z_{6}\triangleright z_{6}\triangleright z_{0}\triangleright z_{1}=z_{1}$} \quad
\color{dc1}{$z_{1}\triangleright z_{5}\triangleright z_{6}\triangleright z_{6}\triangleright z_{1}\triangleright z_{0}=z_{0}$} \quad
\color{dc1}{$z_{1}\triangleright z_{5}\triangleright z_{6}\triangleright z_{6}\triangleright z_{1}=z_{1}~$} \quad
\color{dc2}{$z_{0}\triangleright z_{5}\triangleright z_{2}\triangleright z_{5}\triangleright z_{2}\triangleright z_{3}=z_{3}$} \quad
\color{dc2}{$z_{5}\triangleright z_{0}\triangleright z_{2}\triangleright z_{5}\triangleright z_{2}\triangleright z_{3}=z_{3}$} \quad
\color{dc2}{$z_{5}\triangleright z_{2}\triangleright z_{0}\triangleright z_{5}\triangleright z_{2}\triangleright z_{3}=z_{3}$} \quad
\color{dc2}{$z_{5}\triangleright z_{2}\triangleright z_{5}\triangleright z_{0}\triangleright z_{2}\triangleright z_{3}=z_{3}$} \quad
\color{dc2}{$z_{5}\triangleright z_{2}\triangleright z_{5}\triangleright z_{2}\triangleright z_{0}\triangleright z_{3}=z_{3}$} \quad
\color{dc2}{$z_{5}\triangleright z_{2}\triangleright z_{5}\triangleright z_{2}\triangleright z_{3}\triangleright z_{0}=z_{0}$} \quad
\color{dc2}{$z_{5}\triangleright z_{2}\triangleright z_{5}\triangleright z_{2}\triangleright z_{3}=z_{3}~$} \quad
\color{dc3}{$z_{0}\triangleright z_{3}\triangleright z_{1}\triangleright z_{2}\triangleright z_{2}\triangleright z_{2}=z_{2}$} \quad
\color{dc3}{$z_{3}\triangleright z_{0}\triangleright z_{1}\triangleright z_{2}\triangleright z_{2}\triangleright z_{2}=z_{2}$} \quad
\color{dc3}{$z_{3}\triangleright z_{1}\triangleright z_{0}\triangleright z_{2}\triangleright z_{2}\triangleright z_{2}=z_{2}$} \quad
\color{dc3}{$z_{3}\triangleright z_{1}\triangleright z_{2}\triangleright z_{0}\triangleright z_{2}\triangleright z_{2}=z_{2}$} \quad
\color{dc3}{$z_{3}\triangleright z_{1}\triangleright z_{2}\triangleright z_{2}\triangleright z_{0}\triangleright z_{2}=z_{2}$} \quad
\color{dc3}{$z_{3}\triangleright z_{1}\triangleright z_{2}\triangleright z_{2}\triangleright z_{2}\triangleright z_{0}=z_{0}$} \quad
\color{dc3}{$z_{3}\triangleright z_{1}\triangleright z_{2}\triangleright z_{2}\triangleright z_{2}=z_{2}~$} \quad
\color{dc4}{$z_{0}\triangleright z_{2}\triangleright z_{4}\triangleright z_{4}\triangleright z_{3}\triangleright z_{4}=z_{4}$} \quad
\color{dc4}{$z_{2}\triangleright z_{0}\triangleright z_{4}\triangleright z_{4}\triangleright z_{3}\triangleright z_{4}=z_{4}$} \quad
\color{dc4}{$z_{2}\triangleright z_{4}\triangleright z_{0}\triangleright z_{4}\triangleright z_{3}\triangleright z_{4}=z_{4}$} \quad
\color{dc4}{$z_{2}\triangleright z_{4}\triangleright z_{4}\triangleright z_{0}\triangleright z_{3}\triangleright z_{4}=z_{4}$} \quad
\color{dc4}{$z_{2}\triangleright z_{4}\triangleright z_{4}\triangleright z_{3}\triangleright z_{0}\triangleright z_{4}=z_{4}$} \quad
\color{dc4}{$z_{2}\triangleright z_{4}\triangleright z_{4}\triangleright z_{3}\triangleright z_{4}\triangleright z_{0}=z_{0}$} \quad
\color{dc4}{$z_{2}\triangleright z_{4}\triangleright z_{4}\triangleright z_{3}\triangleright z_{4}=z_{4}~$} \quad
\color{dc5}{$z_{0}\triangleright z_{1}\triangleright z_{4}\triangleright z_{6}\triangleright z_{5}\triangleright z_{2}=z_{2}$} \quad
\color{dc5}{$z_{1}\triangleright z_{0}\triangleright z_{4}\triangleright z_{6}\triangleright z_{5}\triangleright z_{2}=z_{2}$} \quad
\color{dc5}{$z_{1}\triangleright z_{4}\triangleright z_{0}\triangleright z_{6}\triangleright z_{5}\triangleright z_{2}=z_{2}$} \quad
\color{dc5}{$z_{1}\triangleright z_{4}\triangleright z_{6}\triangleright z_{0}\triangleright z_{5}\triangleright z_{2}=z_{2}$} \quad
\color{dc5}{$z_{1}\triangleright z_{4}\triangleright z_{6}\triangleright z_{5}\triangleright z_{0}\triangleright z_{2}=z_{2}$} \quad
\color{dc5}{$z_{1}\triangleright z_{4}\triangleright z_{6}\triangleright z_{5}\triangleright z_{2}=z_{2}~$} \quad
\color{dc11}{$z_{5}\triangleright z_{6}\triangleright z_{1}\triangleright z_{4}\triangleright z_{4}=z_{4}~$} \quad
\color{dc12}{$z_{4}\triangleright z_{2}\triangleright z_{4}\triangleright z_{6}\triangleright z_{3}=z_{3}~$} \quad
\color{dc13}{$z_{5}\triangleright z_{1}\triangleright z_{1}\triangleright z_{5}\triangleright z_{3}=z_{3}~$} \quad
\color{dc14}{$z_{2}\triangleright z_{1}\triangleright z_{2}\triangleright z_{6}\triangleright z_{2}=z_{2}~$} \quad
\color{dc15}{$z_{6}\triangleright z_{1}\triangleright z_{2}\triangleright z_{5}\triangleright z_{4}=z_{4}~$} \quad
\color{dc16}{$z_{1}\triangleright z_{4}\triangleright z_{6}\triangleright z_{1}\triangleright z_{3}=z_{3}~$} \quad
\color{dc17}{$z_{4}\triangleright z_{1}\triangleright z_{2}\triangleright z_{4}\triangleright z_{6}=z_{6}~$} \quad
\color{dc18}{$z_{4}\triangleright z_{4}\triangleright z_{1}\triangleright z_{2}\triangleright z_{2}=z_{2}~$} \quad
\color{dc19}{$z_{2}\triangleright z_{1}\triangleright z_{3}\triangleright z_{5}\triangleright z_{4}=z_{4}~$} \quad
}

\paragraph{\color{black}{Testing, for Operator $\triangleright$}}\mbox{}\\
{\footnotesize \color{dc1}{$z_{4}\triangleright z_{4}\triangleright z_{5}\triangleright z_{6}\triangleright z_{2}\triangleright z_{1}=z_{1}$} \quad
\color{dc1}{$z_{4}\triangleright z_{2}\triangleright z_{6}\triangleright z_{5}\triangleright z_{4}\triangleright z_{1}=z_{1}$} \quad
\color{dc1}{$z_{4}\triangleright z_{4}\triangleright z_{2}\triangleright z_{1}\triangleright z_{6}\triangleright z_{5}=z_{5}$} \quad
\color{dc1}{$z_{6}\triangleright z_{2}\triangleright z_{1}\triangleright z_{5}\triangleright z_{4}\triangleright z_{4}=z_{4}$} \quad
\color{dc1}{$z_{6}\triangleright z_{4}\triangleright z_{1}\triangleright z_{4}\triangleright z_{2}\triangleright z_{5}=z_{5}$} \quad
\color{dc1}{$z_{5}\triangleright z_{2}\triangleright z_{4}\triangleright z_{4}\triangleright z_{6}\triangleright z_{1}=z_{1}$} \quad
\color{dc1}{$z_{4}\triangleright z_{5}\triangleright z_{4}\triangleright z_{6}\triangleright z_{1}\triangleright z_{2}=z_{2}$} \quad
\color{dc1}{$z_{6}\triangleright z_{5}\triangleright z_{1}\triangleright z_{4}\triangleright z_{2}\triangleright z_{4}=z_{4}$} \quad
\color{dc1}{$z_{1}\triangleright z_{2}\triangleright z_{4}\triangleright z_{6}\triangleright z_{5}\triangleright z_{4}=z_{4}$} \quad
\color{dc2}{$z_{6}\triangleright z_{1}\triangleright z_{6}\triangleright z_{2}\triangleright z_{3}\triangleright z_{5}=z_{5}$} \quad
\color{dc2}{$z_{2}\triangleright z_{5}\triangleright z_{3}\triangleright z_{1}\triangleright z_{6}\triangleright z_{6}=z_{6}$} \quad
\color{dc2}{$z_{3}\triangleright z_{2}\triangleright z_{6}\triangleright z_{1}\triangleright z_{5}\triangleright z_{6}=z_{6}$} \quad
\color{dc2}{$z_{6}\triangleright z_{2}\triangleright z_{5}\triangleright z_{1}\triangleright z_{3}\triangleright z_{6}=z_{6}$} \quad
\color{dc2}{$z_{6}\triangleright z_{6}\triangleright z_{3}\triangleright z_{5}\triangleright z_{2}\triangleright z_{1}=z_{1}$} \quad
\color{dc2}{$z_{1}\triangleright z_{6}\triangleright z_{5}\triangleright z_{6}\triangleright z_{3}\triangleright z_{2}=z_{2}$} \quad
\color{dc2}{$z_{5}\triangleright z_{1}\triangleright z_{6}\triangleright z_{3}\triangleright z_{6}\triangleright z_{2}=z_{2}$} \quad
\color{dc2}{$z_{5}\triangleright z_{6}\triangleright z_{6}\triangleright z_{3}\triangleright z_{1}\triangleright z_{2}=z_{2}$} \quad
\color{dc2}{$z_{5}\triangleright z_{3}\triangleright z_{6}\triangleright z_{2}\triangleright z_{6}\triangleright z_{1}=z_{1}$} \quad
\color{dc3}{$z_{2}\triangleright z_{4}\triangleright z_{5}\triangleright z_{2}\triangleright z_{4}\triangleright z_{6}=z_{6}$} \quad
\color{dc3}{$z_{4}\triangleright z_{5}\triangleright z_{4}\triangleright z_{2}\triangleright z_{2}\triangleright z_{6}=z_{6}$} \quad
\color{dc3}{$z_{2}\triangleright z_{5}\triangleright z_{2}\triangleright z_{4}\triangleright z_{4}\triangleright z_{6}=z_{6}$} \quad
\color{dc3}{$z_{5}\triangleright z_{2}\triangleright z_{6}\triangleright z_{4}\triangleright z_{2}\triangleright z_{4}=z_{4}$} \quad
\color{dc3}{$z_{5}\triangleright z_{4}\triangleright z_{6}\triangleright z_{2}\triangleright z_{4}\triangleright z_{2}=z_{2}$} \quad
\color{dc3}{$z_{2}\triangleright z_{4}\triangleright z_{2}\triangleright z_{6}\triangleright z_{4}\triangleright z_{5}=z_{5}$} \quad
\color{dc3}{$z_{4}\triangleright z_{2}\triangleright z_{4}\triangleright z_{5}\triangleright z_{6}\triangleright z_{2}=z_{2}$} \quad
\color{dc3}{$z_{2}\triangleright z_{4}\triangleright z_{5}\triangleright z_{6}\triangleright z_{4}\triangleright z_{2}=z_{2}$} \quad
\color{dc3}{$z_{4}\triangleright z_{5}\triangleright z_{2}\triangleright z_{2}\triangleright z_{4}\triangleright z_{6}=z_{6}$} \quad
\color{dc4}{$z_{2}\triangleright z_{1}\triangleright z_{6}\triangleright z_{5}\triangleright z_{5}\triangleright z_{3}=z_{3}$} \quad
\color{dc4}{$z_{6}\triangleright z_{5}\triangleright z_{3}\triangleright z_{1}\triangleright z_{5}\triangleright z_{2}=z_{2}$} \quad
\color{dc4}{$z_{6}\triangleright z_{2}\triangleright z_{5}\triangleright z_{5}\triangleright z_{1}\triangleright z_{3}=z_{3}$} \quad
\color{dc4}{$z_{3}\triangleright z_{6}\triangleright z_{1}\triangleright z_{5}\triangleright z_{2}\triangleright z_{5}=z_{5}$} \quad
\color{dc4}{$z_{5}\triangleright z_{2}\triangleright z_{5}\triangleright z_{1}\triangleright z_{6}\triangleright z_{3}=z_{3}$} \quad
\color{dc4}{$z_{3}\triangleright z_{1}\triangleright z_{6}\triangleright z_{2}\triangleright z_{5}\triangleright z_{5}=z_{5}$} \quad
\color{dc4}{$z_{3}\triangleright z_{6}\triangleright z_{5}\triangleright z_{2}\triangleright z_{5}\triangleright z_{1}=z_{1}$} \quad
\color{dc4}{$z_{6}\triangleright z_{5}\triangleright z_{1}\triangleright z_{3}\triangleright z_{2}\triangleright z_{5}=z_{5}$} \quad
\color{dc4}{$z_{1}\triangleright z_{5}\triangleright z_{5}\triangleright z_{2}\triangleright z_{6}\triangleright z_{3}=z_{3}$} \quad
\color{dc5}{$z_{2}\triangleright z_{5}\triangleright z_{5}\triangleright z_{5}\triangleright z_{4}\triangleright z_{4}=z_{4}$} \quad
\color{dc5}{$z_{4}\triangleright z_{5}\triangleright z_{5}\triangleright z_{2}\triangleright z_{5}\triangleright z_{4}=z_{4}$} \quad
\color{dc5}{$z_{2}\triangleright z_{5}\triangleright z_{5}\triangleright z_{4}\triangleright z_{5}\triangleright z_{4}=z_{4}$} \quad
\color{dc5}{$z_{2}\triangleright z_{5}\triangleright z_{4}\triangleright z_{5}\triangleright z_{5}\triangleright z_{4}=z_{4}$} \quad
\color{dc5}{$z_{5}\triangleright z_{4}\triangleright z_{5}\triangleright z_{5}\triangleright z_{2}\triangleright z_{4}=z_{4}$} \quad
\color{dc5}{$z_{2}\triangleright z_{4}\triangleright z_{5}\triangleright z_{4}\triangleright z_{5}\triangleright z_{5}=z_{5}$} \quad
\color{dc5}{$z_{5}\triangleright z_{4}\triangleright z_{2}\triangleright z_{5}\triangleright z_{5}\triangleright z_{4}=z_{4}$} \quad
\color{dc5}{$z_{5}\triangleright z_{4}\triangleright z_{5}\triangleright z_{2}\triangleright z_{5}\triangleright z_{4}=z_{4}$} \quad
\color{dc5}{$z_{5}\triangleright z_{2}\triangleright z_{4}\triangleright z_{5}\triangleright z_{5}\triangleright z_{4}=z_{4}$} \quad
\color{dc6}{$z_{4}\triangleright z_{1}\triangleright z_{4}\triangleright z_{5}\triangleright z_{1}\triangleright z_{1}=z_{1}$} \quad
\color{dc6}{$z_{1}\triangleright z_{5}\triangleright z_{4}\triangleright z_{1}\triangleright z_{1}\triangleright z_{4}=z_{4}$} \quad
\color{dc6}{$z_{1}\triangleright z_{1}\triangleright z_{4}\triangleright z_{1}\triangleright z_{5}\triangleright z_{4}=z_{4}$} \quad
\color{dc6}{$z_{4}\triangleright z_{1}\triangleright z_{1}\triangleright z_{5}\triangleright z_{4}\triangleright z_{1}=z_{1}$} \quad
\color{dc6}{$z_{1}\triangleright z_{5}\triangleright z_{1}\triangleright z_{4}\triangleright z_{1}\triangleright z_{4}=z_{4}$} \quad
\color{dc11}{$z_{0}\triangleright z_{5}\triangleright z_{6}\triangleright z_{1}\triangleright z_{4}\triangleright z_{4}=z_{4}$} \quad
\color{dc11}{$z_{5}\triangleright z_{0}\triangleright z_{6}\triangleright z_{1}\triangleright z_{4}\triangleright z_{4}=z_{4}$} \quad
\color{dc11}{$z_{5}\triangleright z_{6}\triangleright z_{0}\triangleright z_{1}\triangleright z_{4}\triangleright z_{4}=z_{4}$} \quad
\color{dc11}{$z_{5}\triangleright z_{6}\triangleright z_{1}\triangleright z_{0}\triangleright z_{4}\triangleright z_{4}=z_{4}$} \quad
\color{dc11}{$z_{5}\triangleright z_{6}\triangleright z_{1}\triangleright z_{4}\triangleright z_{0}\triangleright z_{4}=z_{4}$} \quad
\color{dc11}{$z_{5}\triangleright z_{6}\triangleright z_{1}\triangleright z_{4}\triangleright z_{4}\triangleright z_{0}=z_{0}$} \quad
\color{dc12}{$z_{0}\triangleright z_{4}\triangleright z_{2}\triangleright z_{4}\triangleright z_{6}\triangleright z_{3}=z_{3}$} \quad
\color{dc12}{$z_{4}\triangleright z_{0}\triangleright z_{2}\triangleright z_{4}\triangleright z_{6}\triangleright z_{3}=z_{3}$} \quad
\color{dc12}{$z_{4}\triangleright z_{2}\triangleright z_{0}\triangleright z_{4}\triangleright z_{6}\triangleright z_{3}=z_{3}$} \quad
\color{dc12}{$z_{4}\triangleright z_{2}\triangleright z_{4}\triangleright z_{0}\triangleright z_{6}\triangleright z_{3}=z_{3}$} \quad
\color{dc12}{$z_{4}\triangleright z_{2}\triangleright z_{4}\triangleright z_{6}\triangleright z_{0}\triangleright z_{3}=z_{3}$} \quad
\color{dc12}{$z_{4}\triangleright z_{2}\triangleright z_{4}\triangleright z_{6}\triangleright z_{3}\triangleright z_{0}=z_{0}$} \quad
\color{dc13}{$z_{0}\triangleright z_{5}\triangleright z_{1}\triangleright z_{1}\triangleright z_{5}\triangleright z_{3}=z_{3}$} \quad
\color{dc13}{$z_{5}\triangleright z_{0}\triangleright z_{1}\triangleright z_{1}\triangleright z_{5}\triangleright z_{3}=z_{3}$} \quad
\color{dc13}{$z_{5}\triangleright z_{1}\triangleright z_{0}\triangleright z_{1}\triangleright z_{5}\triangleright z_{3}=z_{3}$} \quad
\color{dc13}{$z_{5}\triangleright z_{1}\triangleright z_{1}\triangleright z_{0}\triangleright z_{5}\triangleright z_{3}=z_{3}$} \quad
\color{dc13}{$z_{5}\triangleright z_{1}\triangleright z_{1}\triangleright z_{5}\triangleright z_{0}\triangleright z_{3}=z_{3}$} \quad
\color{dc13}{$z_{5}\triangleright z_{1}\triangleright z_{1}\triangleright z_{5}\triangleright z_{3}\triangleright z_{0}=z_{0}$} \quad
\color{dc14}{$z_{0}\triangleright z_{2}\triangleright z_{1}\triangleright z_{2}\triangleright z_{6}\triangleright z_{2}=z_{2}$} \quad
\color{dc14}{$z_{2}\triangleright z_{0}\triangleright z_{1}\triangleright z_{2}\triangleright z_{6}\triangleright z_{2}=z_{2}$} \quad
\color{dc14}{$z_{2}\triangleright z_{1}\triangleright z_{0}\triangleright z_{2}\triangleright z_{6}\triangleright z_{2}=z_{2}$} \quad
\color{dc14}{$z_{2}\triangleright z_{1}\triangleright z_{2}\triangleright z_{0}\triangleright z_{6}\triangleright z_{2}=z_{2}$} \quad
\color{dc14}{$z_{2}\triangleright z_{1}\triangleright z_{2}\triangleright z_{6}\triangleright z_{0}\triangleright z_{2}=z_{2}$} \quad
\color{dc14}{$z_{2}\triangleright z_{1}\triangleright z_{2}\triangleright z_{6}\triangleright z_{2}\triangleright z_{0}=z_{0}$} \quad
\color{dc15}{$z_{0}\triangleright z_{6}\triangleright z_{1}\triangleright z_{2}\triangleright z_{5}\triangleright z_{4}=z_{4}$} \quad
\color{dc15}{$z_{6}\triangleright z_{0}\triangleright z_{1}\triangleright z_{2}\triangleright z_{5}\triangleright z_{4}=z_{4}$} \quad
\color{dc15}{$z_{6}\triangleright z_{1}\triangleright z_{0}\triangleright z_{2}\triangleright z_{5}\triangleright z_{4}=z_{4}$} \quad
\color{dc15}{$z_{6}\triangleright z_{1}\triangleright z_{2}\triangleright z_{0}\triangleright z_{5}\triangleright z_{4}=z_{4}$} \quad
\color{dc15}{$z_{6}\triangleright z_{1}\triangleright z_{2}\triangleright z_{5}\triangleright z_{0}\triangleright z_{4}=z_{4}$} \quad
\color{dc15}{$z_{6}\triangleright z_{1}\triangleright z_{2}\triangleright z_{5}\triangleright z_{4}\triangleright z_{0}=z_{0}$} \quad
\color{dc16}{$z_{0}\triangleright z_{1}\triangleright z_{4}\triangleright z_{6}\triangleright z_{1}\triangleright z_{3}=z_{3}$} \quad
\color{dc16}{$z_{1}\triangleright z_{0}\triangleright z_{4}\triangleright z_{6}\triangleright z_{1}\triangleright z_{3}=z_{3}$} \quad
\color{dc16}{$z_{1}\triangleright z_{4}\triangleright z_{0}\triangleright z_{6}\triangleright z_{1}\triangleright z_{3}=z_{3}$} \quad
\color{dc16}{$z_{1}\triangleright z_{4}\triangleright z_{6}\triangleright z_{0}\triangleright z_{1}\triangleright z_{3}=z_{3}$} \quad
\color{dc16}{$z_{1}\triangleright z_{4}\triangleright z_{6}\triangleright z_{1}\triangleright z_{0}\triangleright z_{3}=z_{3}$} \quad
\color{dc16}{$z_{1}\triangleright z_{4}\triangleright z_{6}\triangleright z_{1}\triangleright z_{3}\triangleright z_{0}=z_{0}$} \quad
\color{dc17}{$z_{0}\triangleright z_{4}\triangleright z_{1}\triangleright z_{2}\triangleright z_{4}\triangleright z_{6}=z_{6}$} \quad
\color{dc17}{$z_{4}\triangleright z_{0}\triangleright z_{1}\triangleright z_{2}\triangleright z_{4}\triangleright z_{6}=z_{6}$} \quad
\color{dc17}{$z_{4}\triangleright z_{1}\triangleright z_{0}\triangleright z_{2}\triangleright z_{4}\triangleright z_{6}=z_{6}$} \quad
\color{dc17}{$z_{4}\triangleright z_{1}\triangleright z_{2}\triangleright z_{0}\triangleright z_{4}\triangleright z_{6}=z_{6}$} \quad
\color{dc17}{$z_{4}\triangleright z_{1}\triangleright z_{2}\triangleright z_{4}\triangleright z_{0}\triangleright z_{6}=z_{6}$} \quad
\color{dc17}{$z_{4}\triangleright z_{1}\triangleright z_{2}\triangleright z_{4}\triangleright z_{6}\triangleright z_{0}=z_{0}$} \quad
\color{dc18}{$z_{0}\triangleright z_{4}\triangleright z_{4}\triangleright z_{1}\triangleright z_{2}\triangleright z_{2}=z_{2}$} \quad
\color{dc18}{$z_{4}\triangleright z_{0}\triangleright z_{4}\triangleright z_{1}\triangleright z_{2}\triangleright z_{2}=z_{2}$} \quad
\color{dc18}{$z_{4}\triangleright z_{4}\triangleright z_{0}\triangleright z_{1}\triangleright z_{2}\triangleright z_{2}=z_{2}$} \quad
\color{dc18}{$z_{4}\triangleright z_{4}\triangleright z_{1}\triangleright z_{0}\triangleright z_{2}\triangleright z_{2}=z_{2}$} \quad
\color{dc18}{$z_{4}\triangleright z_{4}\triangleright z_{1}\triangleright z_{2}\triangleright z_{0}\triangleright z_{2}=z_{2}$} \quad
\color{dc18}{$z_{4}\triangleright z_{4}\triangleright z_{1}\triangleright z_{2}\triangleright z_{2}\triangleright z_{0}=z_{0}$} \quad
\color{dc19}{$z_{0}\triangleright z_{2}\triangleright z_{1}\triangleright z_{3}\triangleright z_{5}\triangleright z_{4}=z_{4}$} \quad
\color{dc19}{$z_{2}\triangleright z_{0}\triangleright z_{1}\triangleright z_{3}\triangleright z_{5}\triangleright z_{4}=z_{4}$} \quad
}

\end{document}